\documentclass{article}

\usepackage{amsmath,amssymb}
\usepackage{amsfonts}

\usepackage{multirow}
\usepackage{color}
\usepackage{diagbox}
\usepackage{rotating}
\usepackage{array}
\usepackage{multirow}
\usepackage[super]{nth}
\usepackage{adjustbox}
\usepackage{url}
\usepackage{cellspace}
\usepackage[caption=false]{subfig}

\usepackage[a4paper, total={6in, 9in}]{geometry}

\usepackage{hyperref}
\hypersetup{
    colorlinks=true,
    linkcolor=red,
    filecolor=blue,      
    urlcolor=blue,
    pdftitle={Overleaf Example},
    pdfpagemode=FullScreen,
    }

\usepackage{url}
\usepackage{cellspace}
\newcolumntype{A}{>{\centering\arraybackslash}m{1.2cm}}
\newcolumntype{B}{>{\centering\arraybackslash}m{2.4cm}}
\newcolumntype{C}{>{\centering\arraybackslash}m{1.8cm}}
\newcolumntype{K}{>{\centering\arraybackslash}m{0.3cm}}
\newcolumntype{L}{>{\centering\arraybackslash}m{2cm}}
\newcolumntype{T}{>{\centering\arraybackslash}m{2.4cm}}
\newcolumntype{X}{>{\centering\arraybackslash}m{0.5cm}}
\newcolumntype{J}{>{\centering\arraybackslash}m{0.3cm}}
\newcolumntype{Q}{>{\centering\arraybackslash}m{0.2cm}}
\newcolumntype{Y}{>{\centering\arraybackslash}m{2.04cm}}
\newcolumntype{H}{>{\centering\arraybackslash}m{2.0cm}}
\newcolumntype{P}{>{\centering\arraybackslash}m{3.4cm}}
\newcolumntype{R}{>{\centering\arraybackslash}m{2.2cm}}

\newcommand{\capstyle}[1]{\begin{scriptsize}\textrm{#1}\end{scriptsize}}

\begin{document}


\title{PhIT-Net: Photo-consistent Image Transform for Robust Illumination Invariant Matching} 
\author{Damian Kaliroff \qquad Guy Gilboa\\ 
\hspace{1.2cm}{\tt\small dkaliroff@technion.ac.il} \quad  {\tt\small     guy.gilboa@ee.technion.ac.il}
\\
\\
Technion - Israel Institute of Technology\\
Haifa, Israel
}
\date{}
\maketitle

\begin{abstract}
We propose a new and completely data-driven approach for
generating a photo-consistent image transform.
We show that simple classical algorithms which operate in the transform domain become extremely resilient to illumination changes.
This considerably improves matching accuracy, outperforming 
the use of state-of-the-art invariant representations as well as new matching methods based on deep features. 
The transform is obtained by training a neural network with a specialized triplet loss, designed to emphasize actual scene changes while attenuating illumination changes. The transform yields an illumination invariant representation, structured as an image map, which is highly flexible and can be easily used for various tasks.  
\end{abstract}


\section{Introduction}
\label{sec:introduction}
Image processing and computer vision (CV) tasks often benefit from representations which are invariant to certain image changes.
Photo-consistency is a highly desired property, essential for tasks  based on color and contrast cues, such as matching, registration and recognition.
Traditionally, illumination invariant representations were designed in a model-based manner. Lately, with the rise of deep learning, new data-driven algorithms are proposed to solve the problem. However, the physical assumptions and the models used for the data-driven approaches appear to limit their performance. We thus seek a very general, unconstrained, learning approach.

\begin{figure}[t]
\centering
\includegraphics[trim=0 0.5cm 0 0.3cm,clip,width=0.98\linewidth]{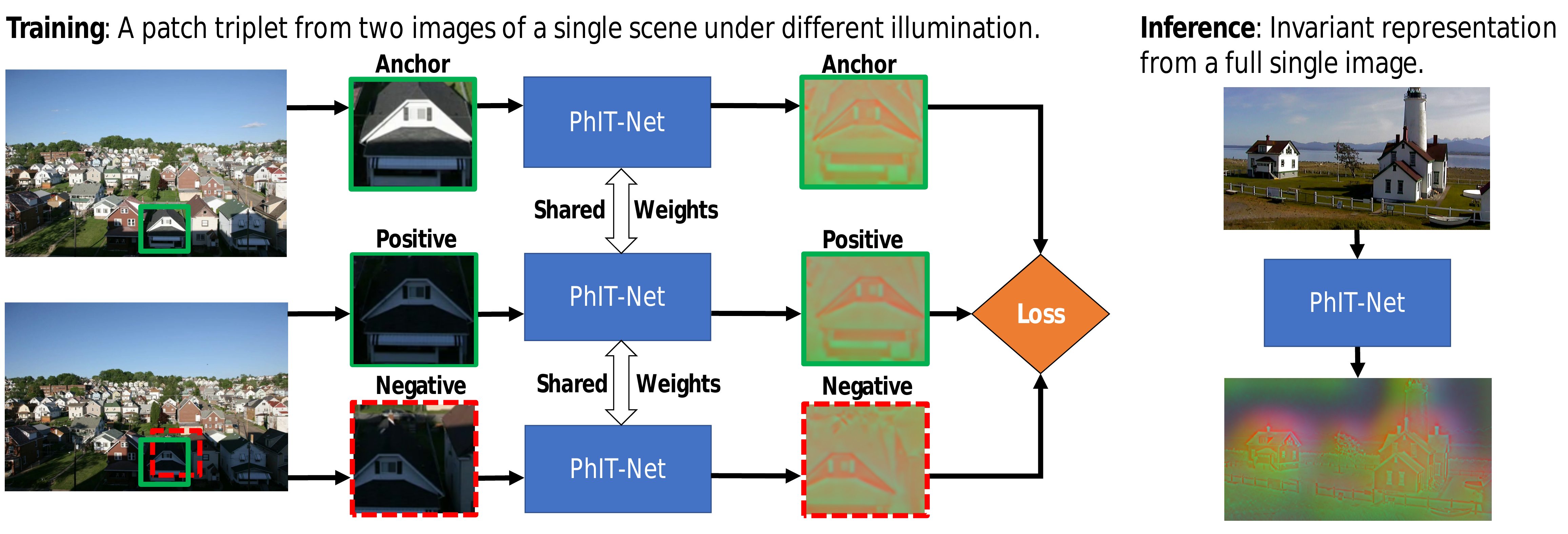}
\\vspace{2em}
\includegraphics[trim=0 0.22cm 0 0.4cm,clip,width=0.98\linewidth]{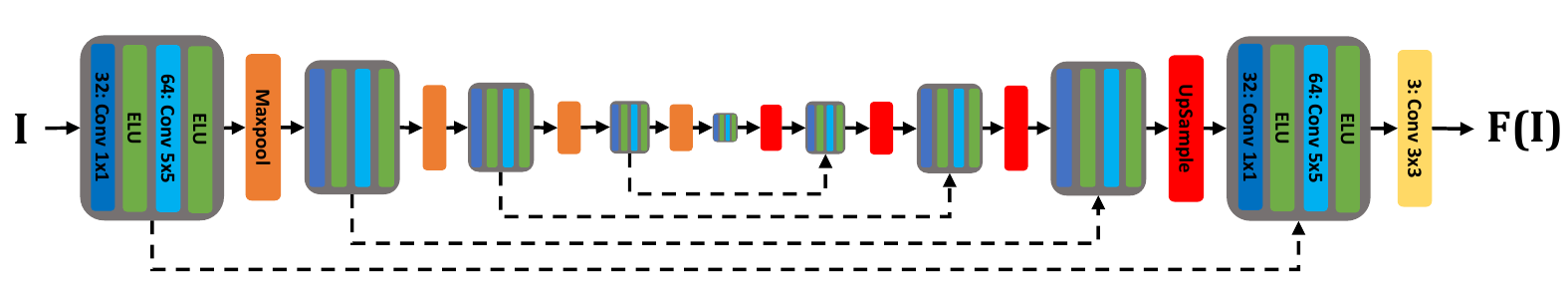}
\caption{PhIT-Net learns a transformation using patch triplets taken from image pairs under different illumination conditions. In inference time the transformation creates an invariant representation from a single input image. The architecture is based on U-Net.}
\label{fig:my_arch}
\end{figure}

In this paper we propose a new paradigm for generating an illumination invariant image map. It is an unconstrained representation, generated in a self-supervised manner, completely data-driven, without using limiting inaccurate assumptions, such as the Lambertian model. We impose mild scale-consistency and geometrical constraints. The surprising representation derived by the training process provides new insights on invariant representations for matching. It can be used as a pre-processing stage for a wide variety of classical and learning-based algorithms, making them considerably more robust to lighting conditions.

To accomplish this, we design a deep neural network, referred to as \emph{PhIT-Net} (\textbf{Ph}oto-consistent \textbf{I}mage \textbf{T}ransform \textbf{Net}work). It is trained in a self-supervised manner, using multiple sets of images of the same scene under different illuminations. This is illustrated in Fig. \ref{fig:my_arch}.
We validate our proposed transform by various means. First, we show that images of the same scene, illuminated differently, are indeed represented in a very similar manner. This is compared to other representations which seek illumination invariance. 
Next, we investigate the usability of our approach. Quantitative experiments are performed for patch matching and for rigid registration. Results are compared to state-of-the-art photo-consistent representations and to novel algorithms based on deep features. In both cases, we show our approach consistently yields superior results.

\section{Related Work}
\label{sec:related_word}
There are two main branches of photo-consistent representations. The first approach attempts to estimate a physical quantity, the albedo (or reflectance) of objects in the image.   
Since the albedo is not affected by illumination and shading, it is inherently photo-consistent. A second branch is based on photo-consistent transforms, which serve to improve computer vision tasks, such as matching or registration. Our approach belongs to the latter.

\smallskip\noindent\textbf{Seeking the elusive albedo.}
Finding an intrinsic image representation is a long standing problem in computer vision. In \cite{land1971Retinex,barrow1978recovering} the Retinex theory was introduced, followed by numerous algorithms, such as \cite{kimmel2003variational,meylan2006high,elad2005retinex,petro2014multiscale,fu2016weighted}, with the aim of estimating reflectance and shading from a single image.
Following the model by Barrow et al. \cite{barrow1978recovering}, which assumes a Lambertian world, an image $I$ is decomposed by $I = A \cdot S$, where $A$ is albedo and $S$ is shading. When this decomposition is based on a single image, it is referred to as SIID (Single Image Intrinsic Decomposition) \cite{bell2014intrinsic,ma2018single,lettry2018unsupervised}. Obtaining the albedo with SIID techniques is a hard ill-posed problem. 
Recent self-supervised deep learning algorithms attempt to learn this decomposition using extensive image data.
In \cite{lettry2018unsupervised}, Lettry et al. created a synthetic dataset of scenes with images under different illumination and trained a Siamese network \cite{bromley1994signature} to decompose images into albedo and shading. 
In \cite{li2018learning}, Li and Snavely learn an albedo-shading decomposition by using natural photos in a dataset referred to as ``BigTime'' 
of indoor and outdoor scenes, each having several images with different lighting conditions.
Recently, in \cite{liu2020learning} a dataset composed of Google street time-lapses is used to build an intrinsic decomposition approach. This method can work with time-lapses also at test time. They demonstrate their approach for the task of artificial scene relighting.
Both \cite{lettry2018unsupervised} and \cite{li2018learning}  
evaluate their results against ground truth intrinsic datasets, e.g. \cite{grosse2009ground,bell2014intrinsic}. The applicability of their albedo estimation for improving the performance of computer vision tasks is not tested. We use the BigTime dataset to develop our proposed photo-consistent transform.

\smallskip\noindent\textbf{Seeking photo-consistency.}
Computer vision algorithms for matching, registration and optical-flow often assume a degree of object photo-consistency.
In practice, however, illumination changes and shading yield an inconsistent representation in the raw color-space. This can be dealt with by either designing much more complex algorithms or by applying a pre-processing transform to the images, which is specifically targeted to increase photo-consistency. The latter approach has often shown to yield robust results, keeping the main CV algorithm simple and fast. We adopt this approach in our work.

In \cite{hafner2013census,muller2011illumination} the census transform is applied to images, and serves as input for optical-flow computation, improving the robustness to illumination changes. In \cite{goh2008wavelet} and recently in \cite{wang2018illumination}, wavelet based pre-processing methods are used to remove the illumination component in face images to improve face recognition algorithms.
For many recognition tasks, certain locations should be detected under different lighting conditions. 
Following the model of Finlayson et al. \cite{finlayson2004intrinsic,finlayson2005removal} 
the authors of \cite{maddern2014illumination} and \cite{shakeri2016illumination} propose single-channel illumination invariant representations of color images to improve place recognition, visual localization and classification algorithms.

\smallskip

We observed that strong illumination changes do not admit the albedo-shading assumptions of the SIID model. Our experiments indicate that albedo estimations are not very robust. 
We thus direct our efforts to finding an unconstrained photoconsisent map to be used by computer vision algorithms in varying lighting environments.

\section{Unconstrained Invariant Representations}
\label{sec:concept}
In this section we formalize the concept of an invariant representation which is unconstrained by a physical model. 
Let $R_\theta$ be some image transformation with a set of parameters $\theta$. The transformation represents different conditions in which the image was acquired. It can model different attributes, such as illumination changes, noise, fog or atmospheric disturbances. Let $f_{int}$ be an \emph{intrinsic} image representation.
An image instance $f_i$ is obtained by applying the transformation $R_{\theta_i}$, with specific parameters $\theta_i$ to the intrinsic image,
\begin{equation}
\label{eq:fi}
    f_i=R_{\theta_i}(f_{int}).
\end{equation}
For the intrinsic representation problem, the aim is to estimate $f_{int}$, given a single or multiple instances $f_i$ in a blind manner, that is - without knowing $\theta_i$. This is a difficult ill-posed problem. As an example, for the SIID problem, $I=f_i$, $f_{int}=A$, $\theta_i=S$ is some specific shading component and $R_{\theta}(f)=\theta \cdot f$.

\begin{figure}[htb]
\begin{center}
\includegraphics[width=0.98\linewidth]{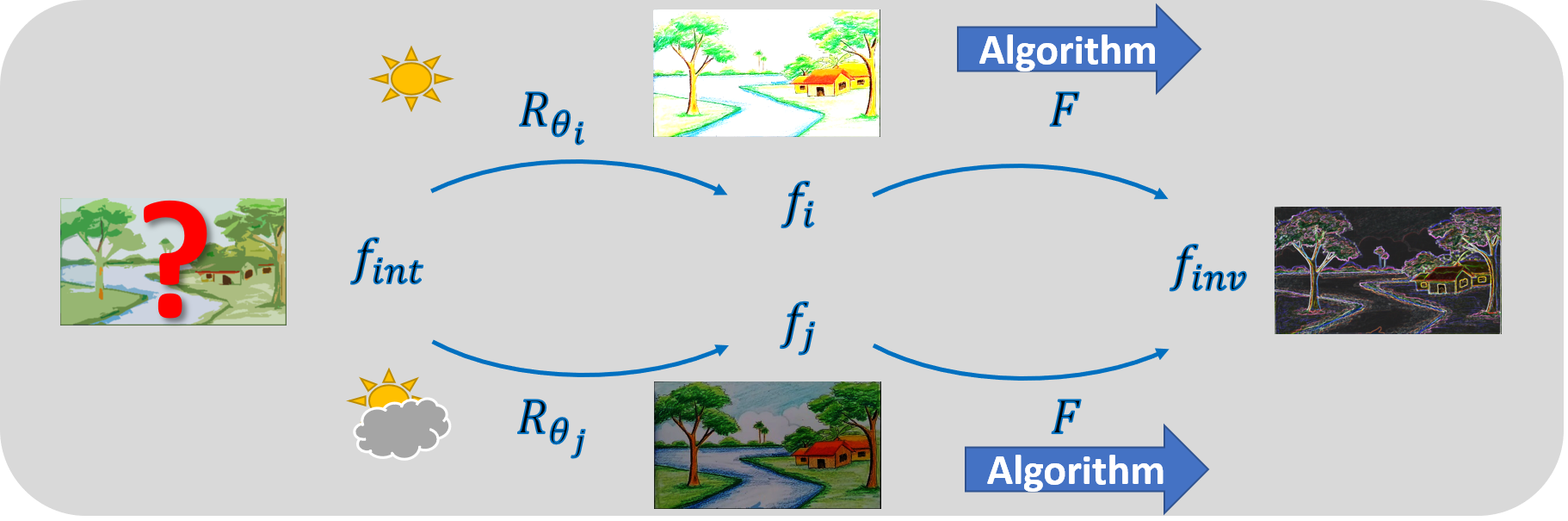}
\end{center}
\caption[Invariant representation ideal concept]{The  invariant representation $f_{inv}$ is generated by applying a transform $F$ to an image $f_k$. Ideally, all images $f_k$, $k=i,j,..$, emerging from the same intrinsic image $f_{int}$, by different transformations $R_{\theta_{k}}$ (image conditions), are mapped to the same invariant representation.}
\label{fig:concept_inv}
\end{figure}

In our approach, we aim only at obtaining an \emph{invariant} representation. Thus, we do not seek to estimate $\theta_i$ and $f_{int}$. Instead, we would like to obtain a transform applied to an image instance, which is invariant to $R_\theta$. This transform is denoted as $F(\cdot)$, and its purpose is depicted in Fig. \ref{fig:concept_inv}. In the ideal case, all image instances $f_{i}$,  created by $R_{\theta_i}$ applied to a specific intrinsic image $f_{int}$, are mapped by  $F(\cdot)$ to the same invariant image, denoted as $f_{inv}$. Moreover, if two intrinsic scenes can be distinguished $d(f_{int},g_{int})>\varepsilon$, where $d(\cdot,\cdot)$ is some metric, so do the respective mappings $d(F(f_i), F(g_i)) > \delta$.

We now formulate the requirements for approximating this concept. 
Let ${\bf f} = \{f_1, .. , f_N\}$ be a set of $N$ instances of the same scene $f_{int}$, taken under different conditions, where $f_i$ is defined by Eq. \eqref{eq:fi}. Let $\bf{g}$ be defined in a similar manner with respect to a different scene $g_{int}$. Then $F$ admits the following properties:
\begin{equation}
\label{eq:cond1}
    D(F(f_i),F(f_j)) \le \epsilon,
\end{equation}
$\forall i,j = 1, ..., N$, where $D(\cdot, \cdot)$ is some distance and $\epsilon$ is a small constant. 
 In addition, 
\begin{equation}
\label{eq:cond2}
    D(F(f_i),F(g_j)) \ge c \cdot D(f_i,g_j),
\end{equation}
$\forall i,j = 1, ..., N$, where $c \gg \epsilon$ is some positive constant. A transform $F$ admitting the above properties yields an unconstrained representation (not limited by formation models), approximately invariant under the transformation $R_\theta$,
\begin{equation}
\label{eq:f_hat}
    {f}_{inv,i} = F(f_i),
\end{equation}
where ${f}_{inv,i} \approx {f}_{inv,j}$, $\forall i,j = 1, ..., N$. 
%
This transform minimizes the difference of images depicting the same scene (created from the same intrinsic image), and emphasizes differences between images from different scenes.

We introduce an additional requirement, which states that the properties above approximately hold for any part of the image. More formally, let us define a cropping operation of the image $crop_{X}$, where $X=(x_1,x_2,y_1,y_2)$ defines the cropping coordinates. We would like
\begin{equation}
\label{eq:crop1}
    crop_{X}(f_{inv,i}) \approx crop_{X}(f_{inv,j}), \,\, \forall i,j = 1, ..., N.
\end{equation}
Moreover, to preserve the geometrical structure, it is desired that the crop operation also approximately commutes with $F$, that is 
\begin{equation}
\label{eq:crop2}
    crop_{X}(F(f)) \approx F(crop_{X}(f)). 
\end{equation}
The combination of both requirements, Eqs. \eqref{eq:crop1} and \eqref{eq:crop2}, can be written as,
\begin{equation}
\label{eq:crop_inv}
    crop_{X}(F(f_i)) \approx F(crop_{X}(f_j)), \,\, \forall i,j = 1, ..., N.
\end{equation}

In order for Eq. \eqref{eq:crop2} to be meaningful, $f$ and $F(f)$ should have the same spatial dimensions. We refer to such a spatial representation as a \emph{map}.
For an input image of $n$ pixels with $k_i$ channels, the output is a map of $n$ pixels with $k_o$ channels, where $k_o$ is a free parameter. We thus have \hbox{$F:\mathbb{R}^{n\times k_i}\to \mathbb{R}^{n\times k_o}$}.
In order to obtain the transform $F$ we do not need to directly model $R_\theta$. We assume to have a training set comprised of $M$ sets ${\bf f^m}$, $m = 1, ..., M$, each comprised of $N$ instances of the same scene $f^m_{int}$ transformed by $R_{\theta_i^m}$. 
We train a network that takes as input an instance $f_i^m$ and produces an output $F(f_i^m)$, using a triplet network model \cite{hoffer2015deep,kumar2016learning}, following Eqs. \eqref{eq:cond1} and \eqref{eq:cond2}. Additional losses are required to obtain a well-behaved, geometrically-consistent, sharp representation with several channels, as detailed above. 
This general methodology can be applied to develop representations invariant to different nuisance attributes. In this work, we develop a photo-consistent transform by applying this approach, obtaining an illumination invariant representation.


\section{Application of Proposed Framework}
\label{sec:method}
We apply the framework presented in Section \ref{sec:concept} for the problem of designing a photo-consistent transform. 
In this context, the transformations $R_\theta$ model different illumination conditions and our aim is to find $F(\cdot)$, such that it admits Eqs. \eqref{eq:cond1}-\eqref{eq:crop2}. 
We use a neural network to compute $F$.
We attempt to decrease the distance in the representation space between two corresponding patches of the same region in the scene, acquired at different lighting conditions.
The training and inference procedures are illustrated schematically in Fig. \ref{fig:my_arch}.
\newline
Our code is publicly available on GitHub \url{https://github.com/dkaliroff/phitnet}.

\newpage
\noindent{\bf CNN Architecture.}
The CNN architecture of PhIT-Net is designed as an encoder-decoder U-net network \cite{ronneberger2015u}. 
The encoder and decoder are constructed using the same convolutional inception-like layers \cite{szegedy2015going}. After the last decoder block, there is a 3x3 convolution layer to generate the final representation. The number of convolutions in this layer is determined by the number of channels in the final representation (three in our model).

\medskip
\noindent{\bf Training Process.}
We train our network using a triplet network training scheme \cite{hoffer2015deep,kumar2016learning}. In this scheme three instances of the same network are trained with shared weights. The input to the model is called a \emph{patch triplet}. Each triplet is extracted from a pair of aligned images $I_1$, $I_2$, of the same scene, under different illumination conditions. For training, we use patches of size $64 \times 64$ pixels. The triplet is composed of an anchor patch (A), a positive patch (P) and a negative patch (N). 
The patches are defined as follows:

\smallskip
\noindent\emph{Anchor Patch (A).} A random patch extracted from $I_1$, with a standard deviation above some threshold $\sigma_{p}$. We chose $\sigma_{p}=25$ for the entire dataset (pixel values are in the range $[0,255]$).

\smallskip
\noindent\emph{Positive Patch (P).} A patch extracted from $I_2$, with the same coordinates as the Anchor, such that
both patches depict the same scene (the RGB difference should stem mainly from illumination differences). In the representation space we would like (A) and (P) to be similar. 

\smallskip
\noindent\emph{Negative Patch (N).} A patch extracted from $I_2$ with shifted coordinates, relative to the positive patch. The shift is of 8 pixels, with respect to the anchor, with randomly chosen direction. The overlap induces a challenging and meaningful learning process. 

\medskip
\noindent{\bf Inference.}
At inference, a full input image is first passed through a single instance of the network, yielding floating point values in an arbitrary range. In order to reach an 8-bit image-format, as in the original images, we normalize the values over all channels linearly such that the minimum is mapped to 0 and the maximum to 255.

\begin{figure}
\centering
\includegraphics[width=0.80\linewidth]{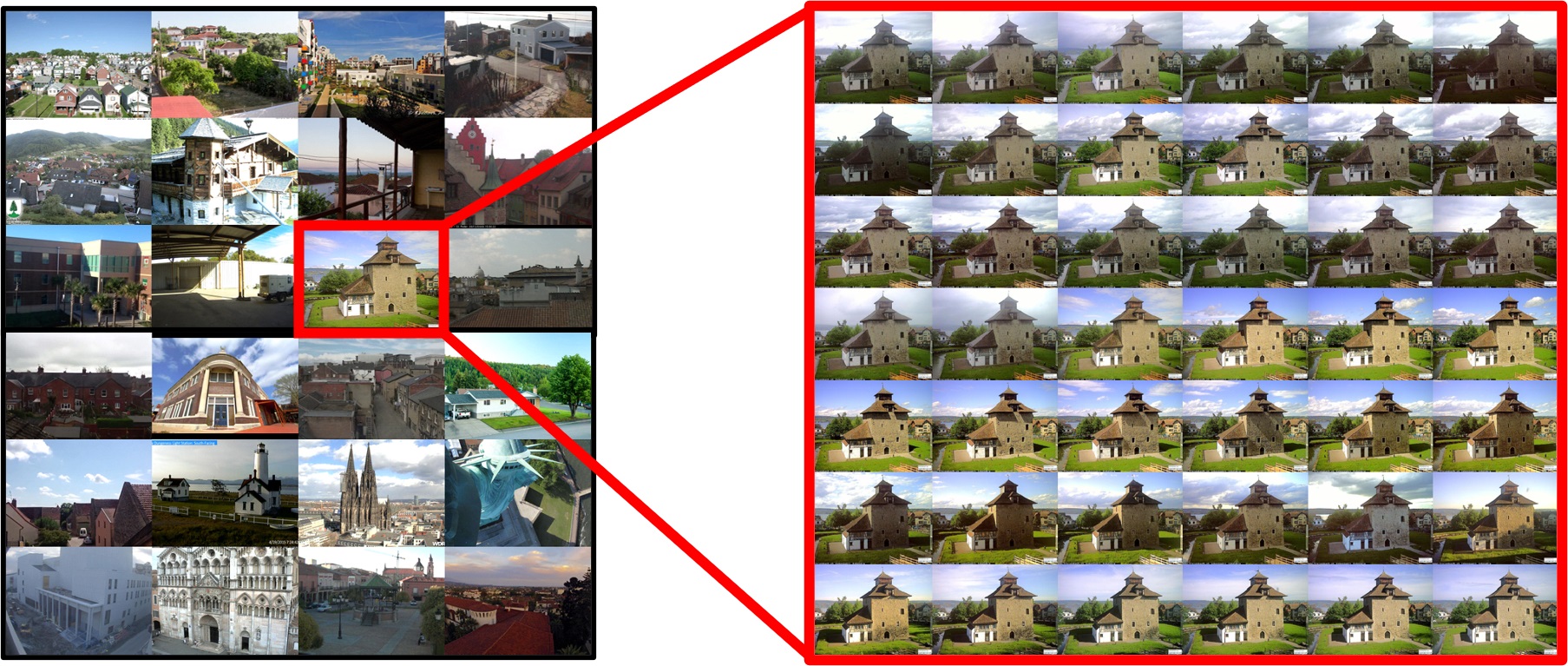}
\caption{BigTime \cite{li2018learning} dataset. Each scene is acquired under several lighting conditions. 
}
\label{fig:BTsample}
\end{figure}

\medskip
\noindent{\bf Training and Test Data.}
We use two different sets for outdoor and indoor settings. The main dataset is the outdoors dataset. It is composed of images from the BigTime dataset (See Fig. \ref{fig:BTsample}). 
We also train and evaluate our model on a set of indoor images. More details about the datasets and evaluation on the additional indoors dataset are provided in the supplementary material.
%
The training is based on square patches of $64 \times 64$ pixels. 
The training set is composed of $240K$ triplets, extracted from $600$ image pairs of 10 outdoor scenes.
The evaluation was done using 100 image pairs selected from 17 additional outdoor scenes not used in training.

\newpage
\noindent{\bf Loss Functions.}
The main loss function for the training process is the triplet loss \cite{weinberger2009distance,hermans2017defense}. It aims at minimizing the distance between (A) and (P), while maximizing (up to a margin) the distance between (A) and (N). In order to reach a meaningful representation additional losses are required. These enable us to achieve some desired properties of the representation, such as scale consistency and channel variability. 
Let $(f_a, f_p, f_n)$ be a triplet of image patches corresponding to (Anchor, Positive, Negative), respectively.
Let $F(\cdot)$ be the output of PhIT-Net.
Let $D_i(\cdot,\cdot)$ be a distance function. 
The total loss function is defined as a weighted sum of the following loss functions:

\smallskip
\noindent\textbf{Triplet Loss (Inter-Loss):} 
\begin{equation}
         L_T(f_a,f_p,f_n)=\max\{0,D_{corr}(F(f_a),F(f_p))\\
         -D_{corr}(F(f_a),F(f_n))+M\},
\label{eq:tripletloss}
\end{equation}
where $M$ is the triplet-loss margin (we use $M=0.1$). Since patch affinity is often defined by correlation, we used the correlation distance function,
\begin{equation}
    D_{corr}(x_1,x_2)=1-\frac{x_1 \cdot x_2}{\|x_1\|_2 \cdot \|x_2 \|_2}.
\end{equation}

\smallskip
\noindent\textbf{Intra-Loss:}
\begin{equation}
        L_I(f_a,f_p)= D_{corr}(f_a,f_p)+ \|f_a - f_p \|_2^2.
\label{eq:intraloss}
\end{equation}
This loss promotes low A-P distance (in addition to the  triplet loss), as suggested by \cite{Cheng_2016_CVPR}. Our experiments verify that adding this loss to the main triplet loss indeed improves performance. It also allows to minimize an additional distance function, not used in the main triplet loss.

\smallskip
\noindent\textbf{Scale Consistency Loss:}
\begin{equation}
\label{eq:scale_const}
    L_{SC}(f_a)=D_{scale}(F(G(f_a,\rho)),G(F(f_a),\rho)),
\end{equation}
where $G$ is "Up-Sample and Crop" and represents a bilinear up-sampling by a random factor $\rho \in \left(1,2\right] $ followed by a crop to the size of the original patch.
The goal of this function, following Eq. \eqref{eq:crop2}, is to make the representation close to commutative with respect to these operations, as real images are. We use $D_{scale}=D_{corr}$.
    
\smallskip
\noindent\textbf{Multi-Channel (MC) Similarity Loss:}
Let $I = F(f)$ be a multi-channel representation of $K$ channels,
$I = (I_1,..\, I_K)$. The multi-channel loss is, 
\begin{equation}
\label{eq:multi_channel}
     L_{MC}(I)=\sum_{i}\sum_{j\neq i} {(1-D_{corr}(I_{i},I_{j}))^2}.
\end{equation}
We want the multi-channel representation to have significant and different information in each channel. Thus, we penalize channel similarity.

\section{Evaluation}
\label{sec:evaluation}
We first examine the nature of the representation and its basic properties. 
As a sanity check, we verify that two images of the same scene under different illuminations are similar in the representation space. 
We then demonstrate the usability of our representation and show it improves the results of common computer vision tasks. Two tasks are examined quantitatively: patch matching and rigid registration.

\subsubsection*{Insights on the New Representation}
\label{sec:qualitative}
\begin{figure}
\begin{center}
\setlength{\tabcolsep}{0.1em} 
\setlength{\belowcaptionskip}{-10pt}
\begin{center}
\begin{tabular}{ARRRRRR}
\capstyle{Original image} &
{\includegraphics[height=2.0cm]{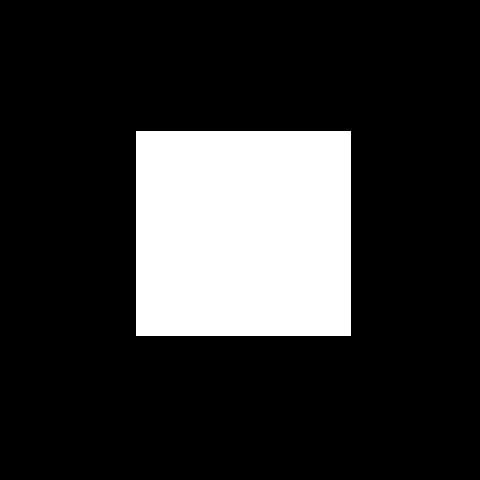}} &
{\includegraphics[height=2.0cm]{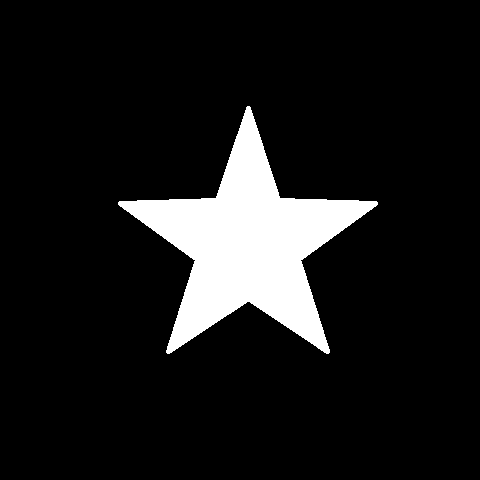}} &
{\includegraphics[height=2.0cm]{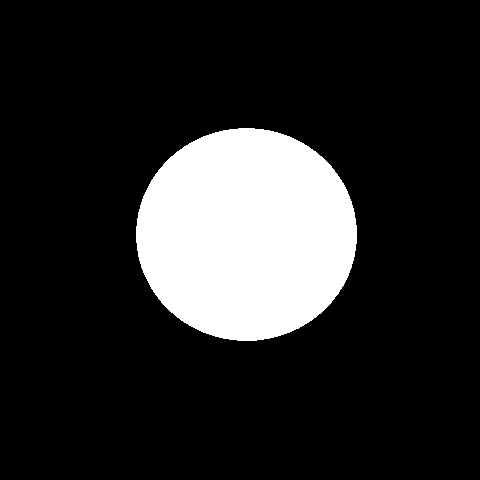}} &
{\includegraphics[height=2.0cm]{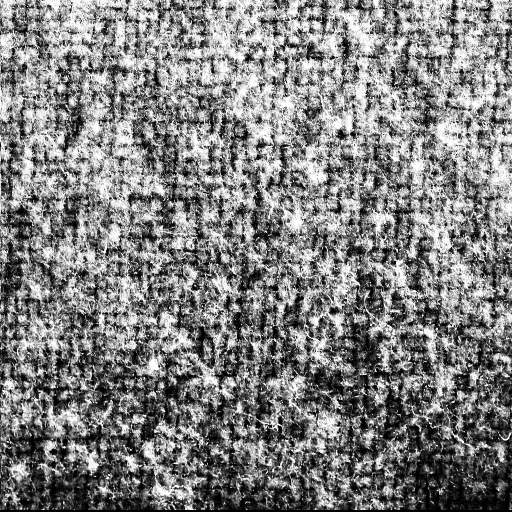}} &
{\includegraphics[height=2.0cm]{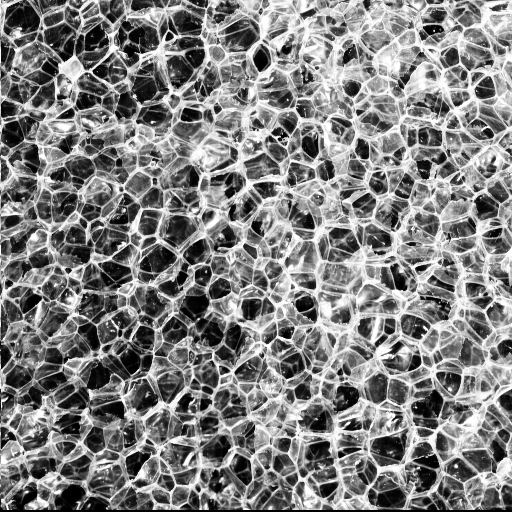}} &
{\includegraphics[height=2.0cm]{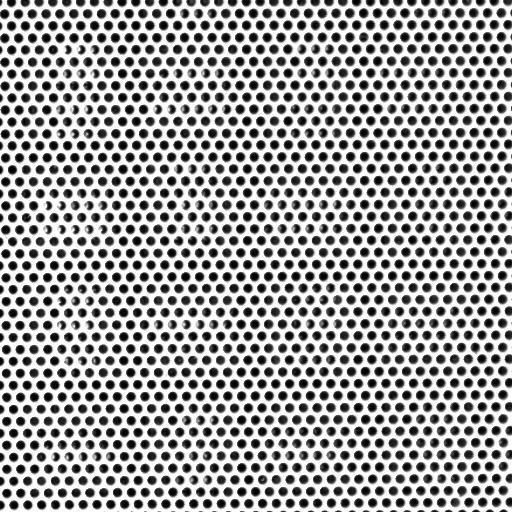}}
\\
\capstyle{PhIT-Net} &
{\includegraphics[height=2.0cm]{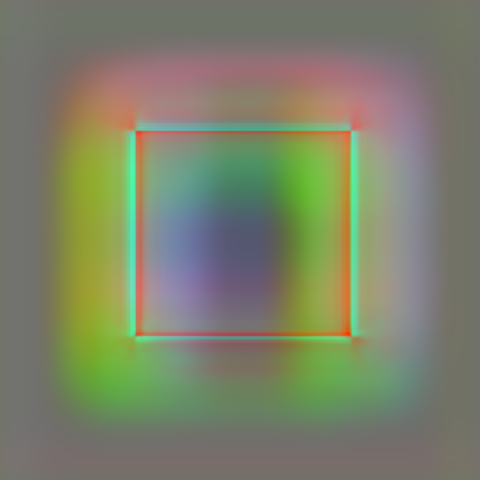}} &
{\includegraphics[height=2.0cm]{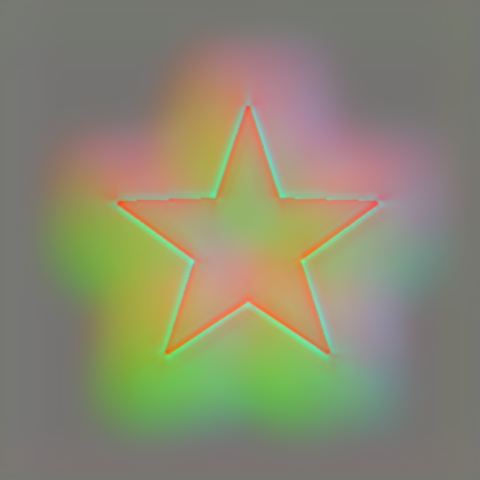}} &
{\includegraphics[height=2.0cm]{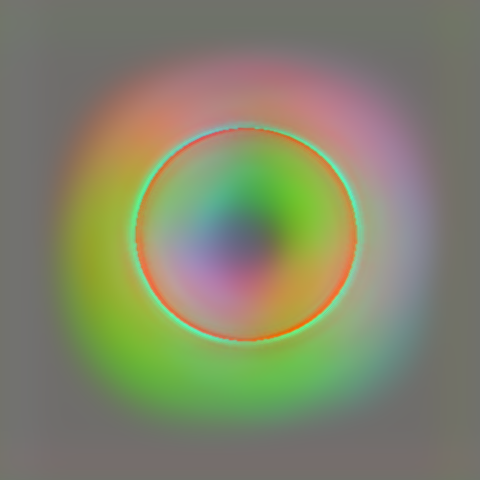}}&
{\includegraphics[height=2.0cm]{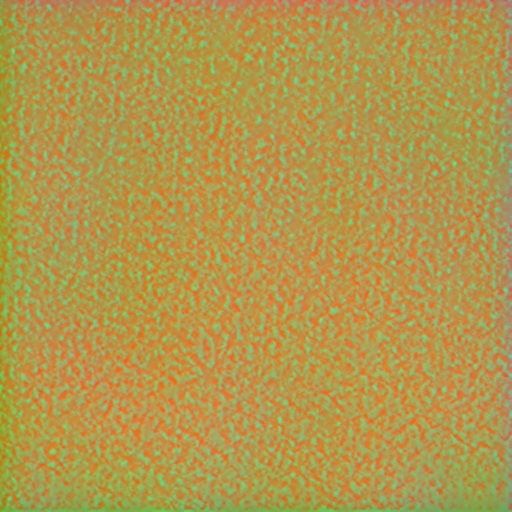}} &
{\includegraphics[height=2.0cm]{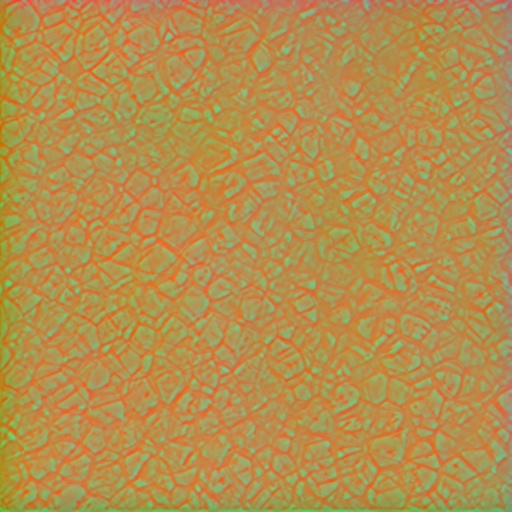}} &
{\includegraphics[height=2.0cm]{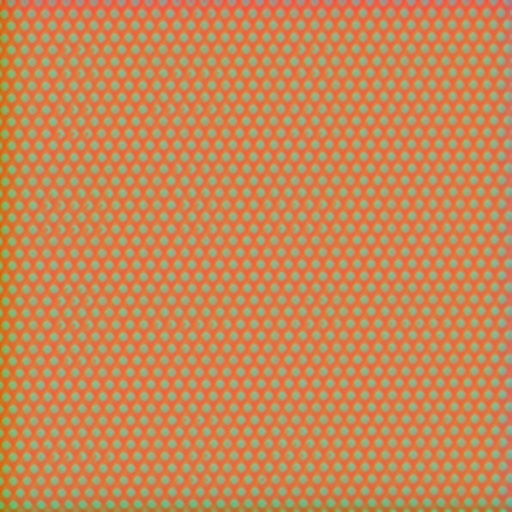}}
\end{tabular}
\end{center}
\caption{
Basic shapes and Brodatz \cite{smith1994transform} textures and their corresponding transform.} 
\label{fig:shapes}
\end{center}
\end{figure}

\smallskip\textbf{Textures, Shapes and Color-Coding.}
In Fig. \ref{fig:shapes} we show the results of the proposed transform applied on  basic shapes and textures. 
First, piecewise-constant shapes are examined. It is evident that edges are clearly defined. We observe that a certain color-coding is created. In flat regions the color provides information on the direction and distance of a nearby dominant edge, in a similar manner to the Chamfer distance transform \cite{barrow1977chamfer}. We interpret this as means to disambiguate better flat regions with little variance. This property can assist matching algorithms. In addition, we examine textures with varying intensity. A highly uniform textural output is obtained in the representation space. Note that such textures do not appear in the training set. This demonstrates the generalization strength of PhIT-Net.

\clearpage

\begin{figure}[htb]
\begin{center}
\setlength{\tabcolsep}{0.1em} 
\begin{tabular}{APPPP}
\centering
\capstyle{Original image} &
{\includegraphics[height=1.800cm]{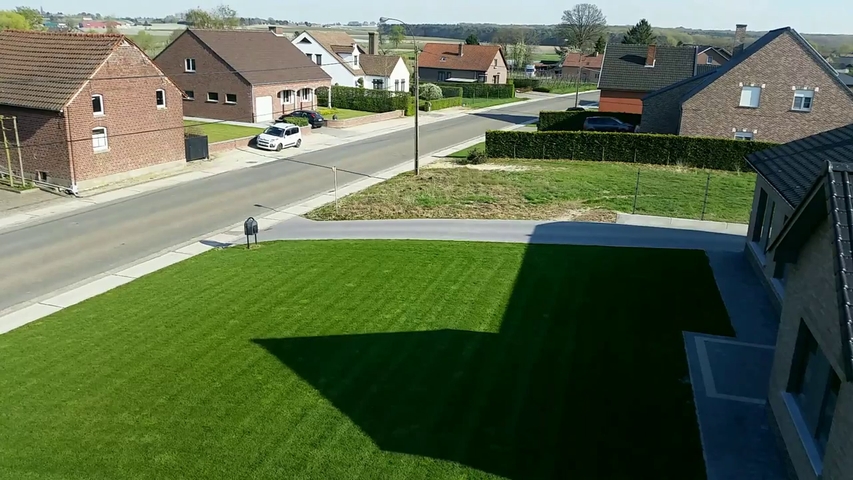}} &
{\includegraphics[height=1.800cm]{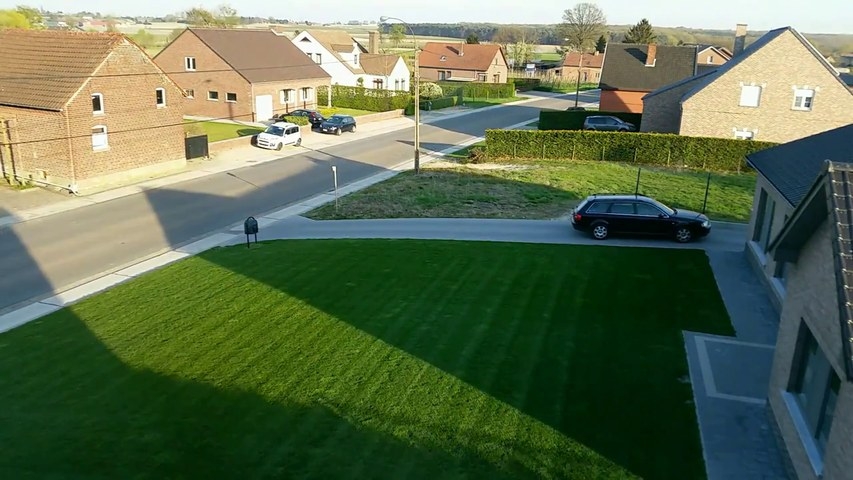}} &
{\includegraphics[height=1.800cm]{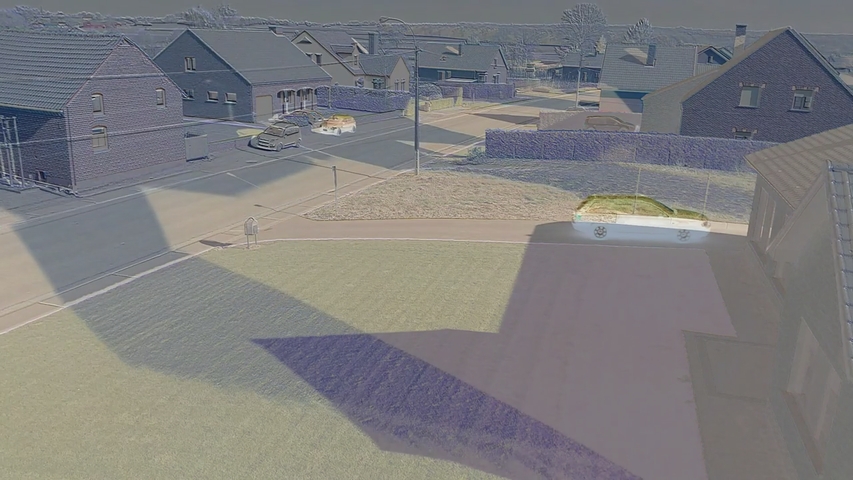}} &
{\includegraphics[height=1.800cm]{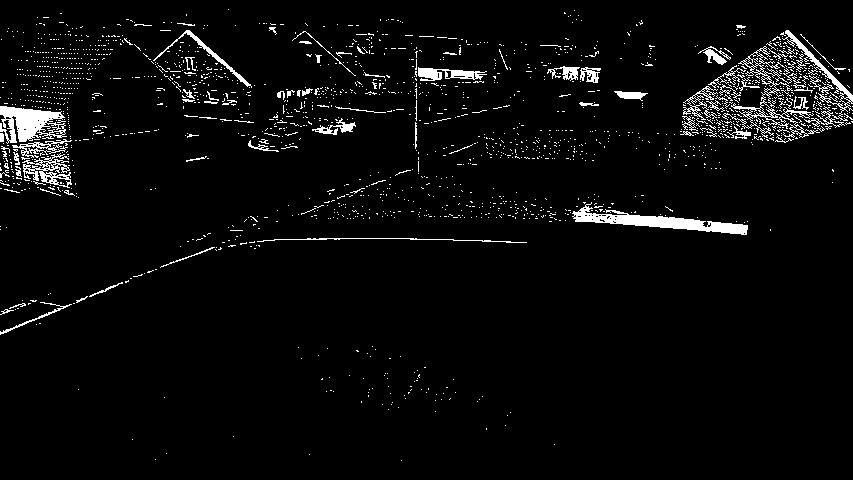}}
\\
\capstyle{Li-Snavely} &
{\includegraphics[height=1.800cm]{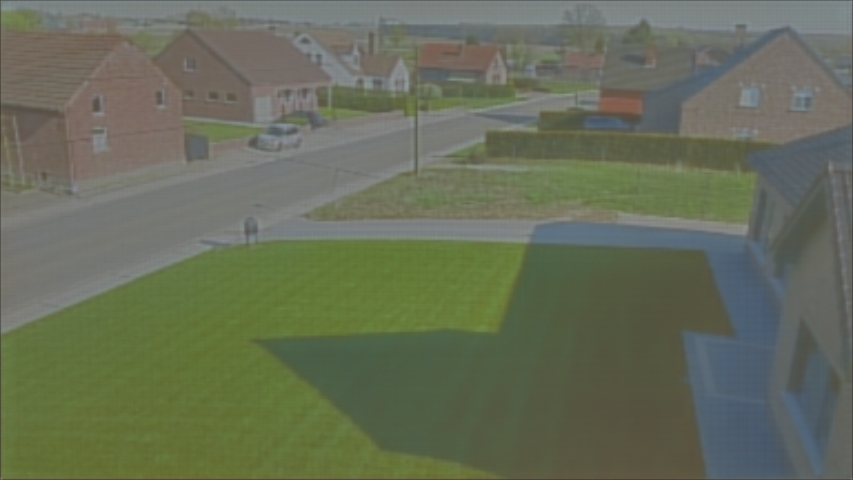}} &
{\includegraphics[height=1.800cm]{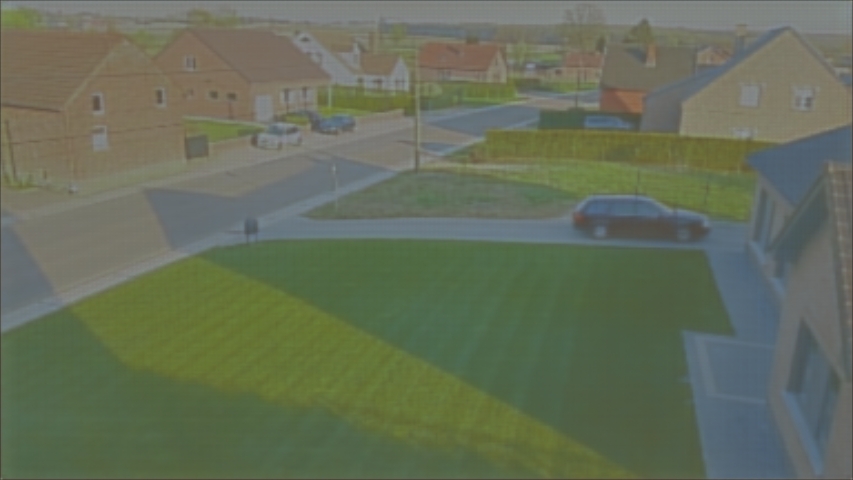}} &
{\includegraphics[height=1.800cm]{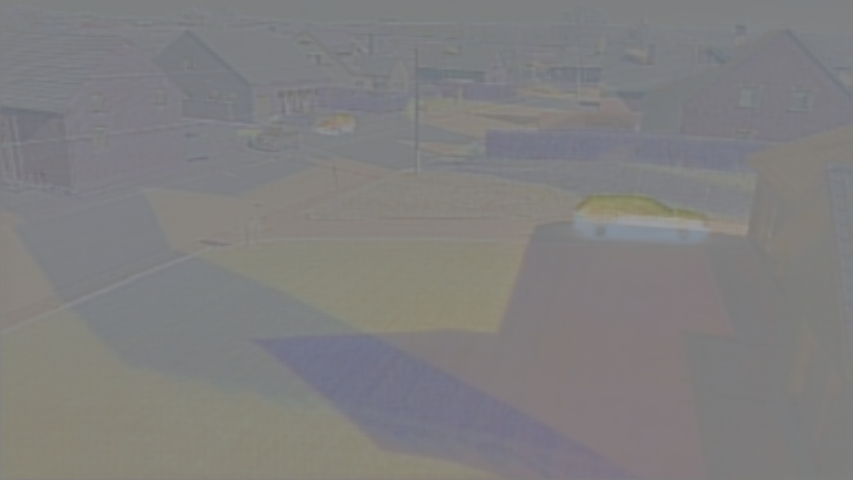}} &
{\includegraphics[height=1.800cm]{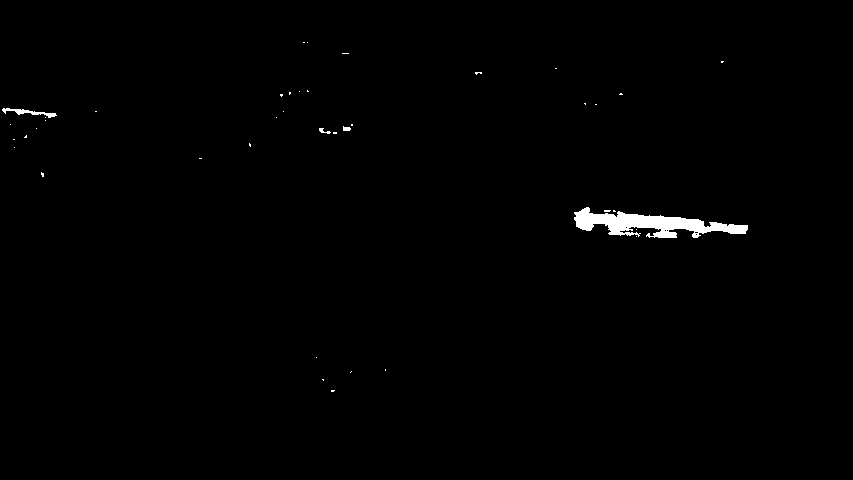}}
\\
\capstyle{Ours} &
{\includegraphics[height=1.800cm]{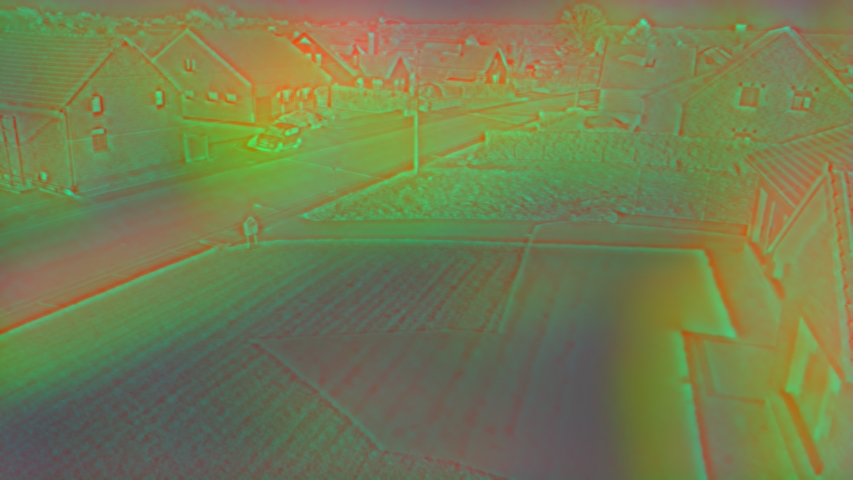}} &
{\includegraphics[height=1.800cm]{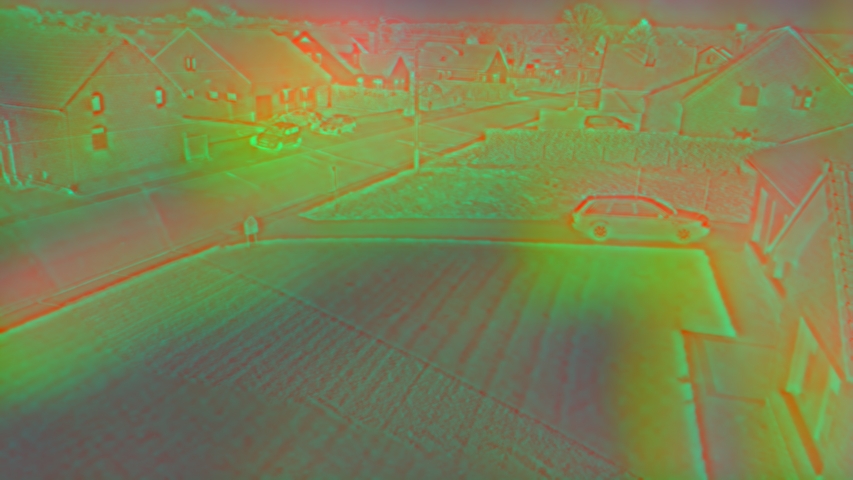}} &
{\includegraphics[height=1.800cm]{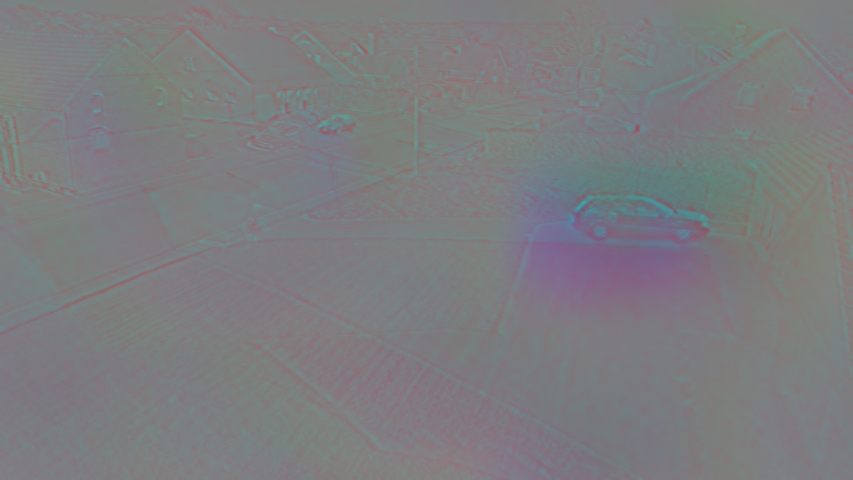}} &
{\includegraphics[height=1.800cm]{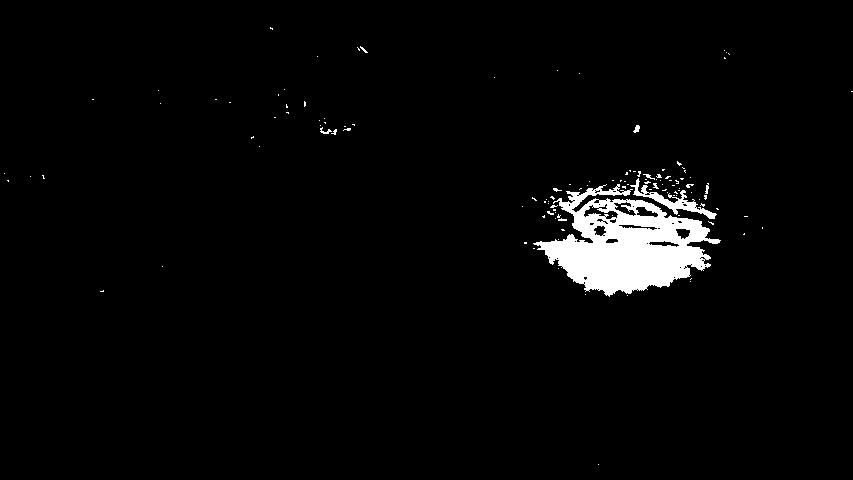}}
\\
&
\capstyle{Image 1} &
\capstyle{Image 2} &
\capstyle{Difference} &
\capstyle{Diff. Thresh.}
\\
\end{tabular}
\caption{
Actual scene differences (the car in this example) can be clearly detected under our transform, while illumination differences are attenuated. 
  scene  differences  (the  car}\label{fig:diff}
\end{center}
\end{figure}

\begin{figure}
\begin{center}
\setlength{\tabcolsep}{0.1em} 
\begin{tabular}{ABBBXCCC}
\capstyle{Original image} &
{\includegraphics[height=1.35cm]{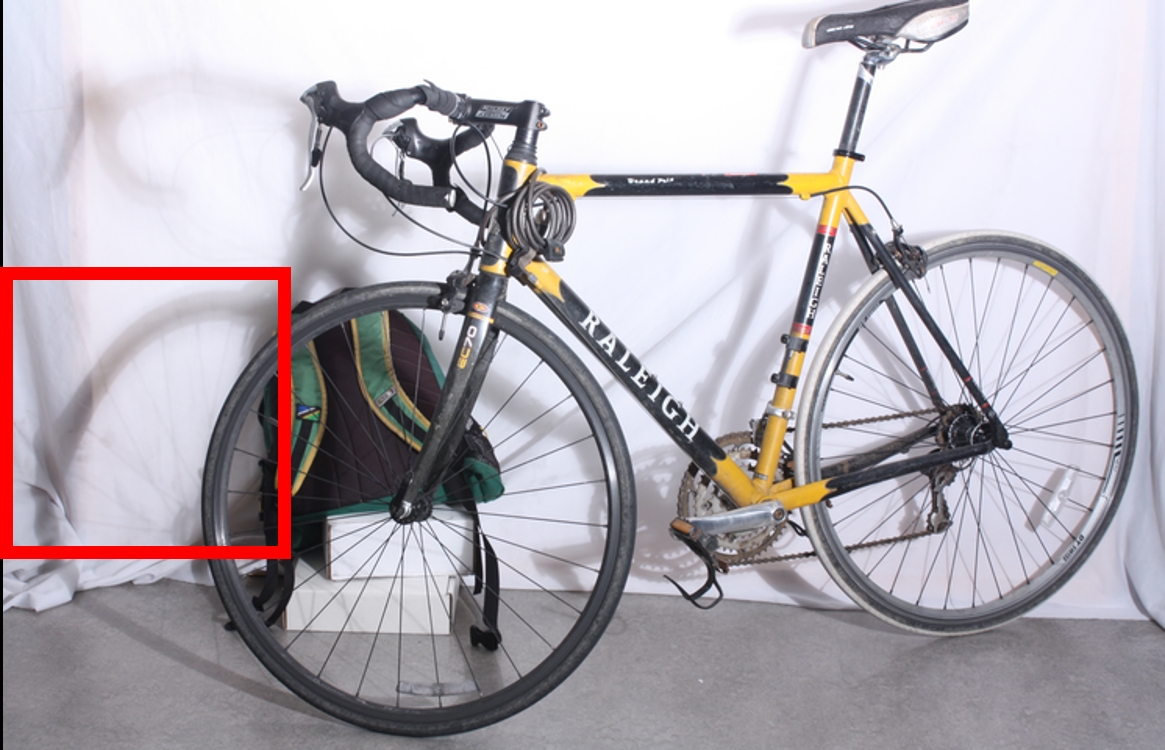}} &
{\includegraphics[height=1.35cm]{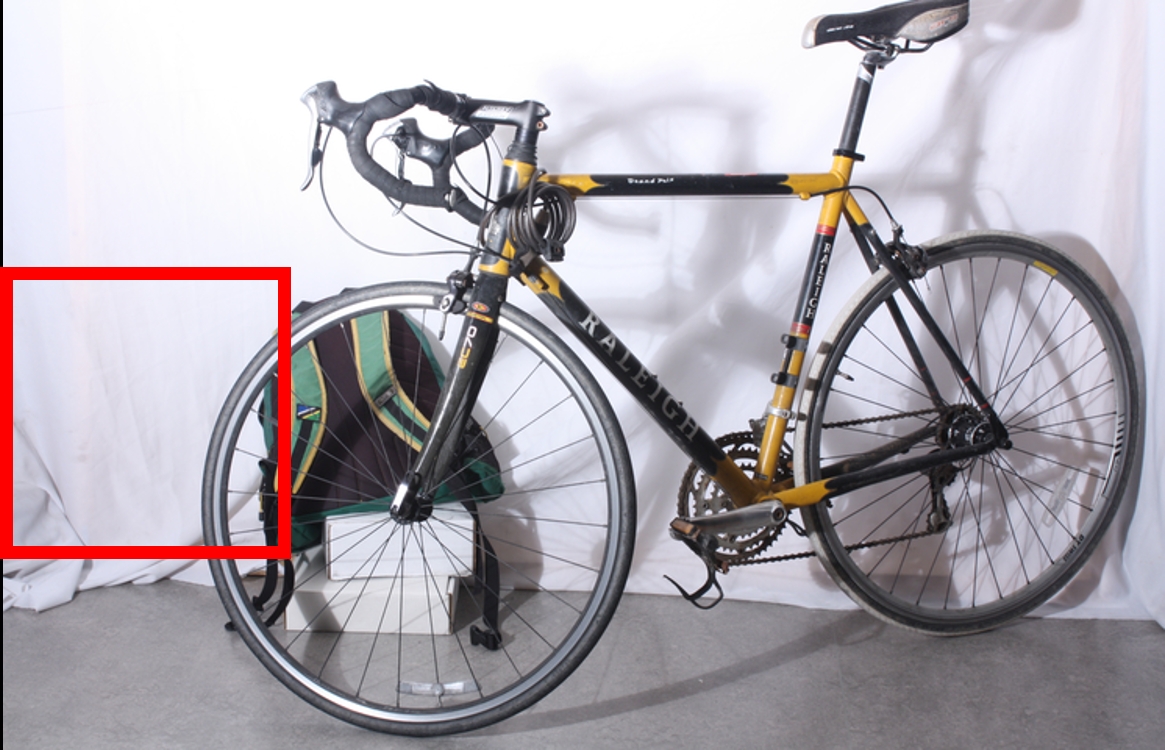}} &
{\includegraphics[height=1.35cm]{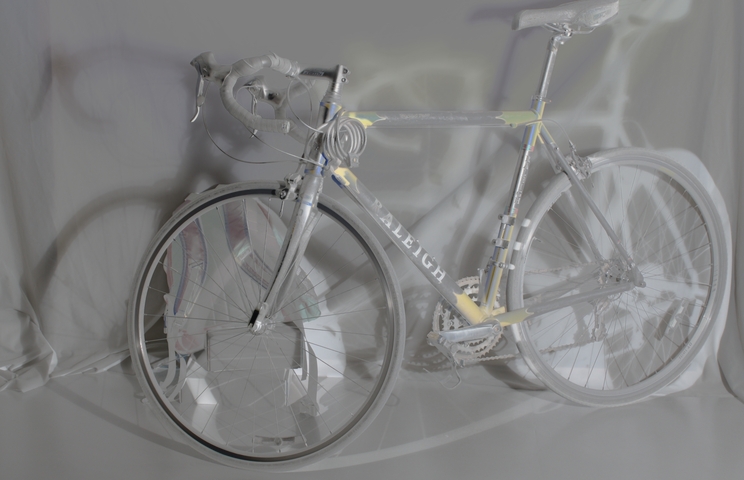}} &
&
{\includegraphics[height=1.35cm]{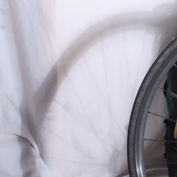}} &
{\includegraphics[height=1.35cm]{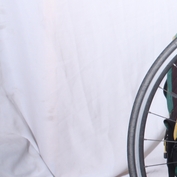}} &
{\includegraphics[height=1.35cm]{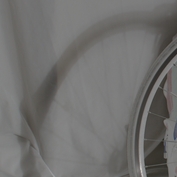}} 
\\
\capstyle{Lettry} &
{\includegraphics[height=1.35cm]{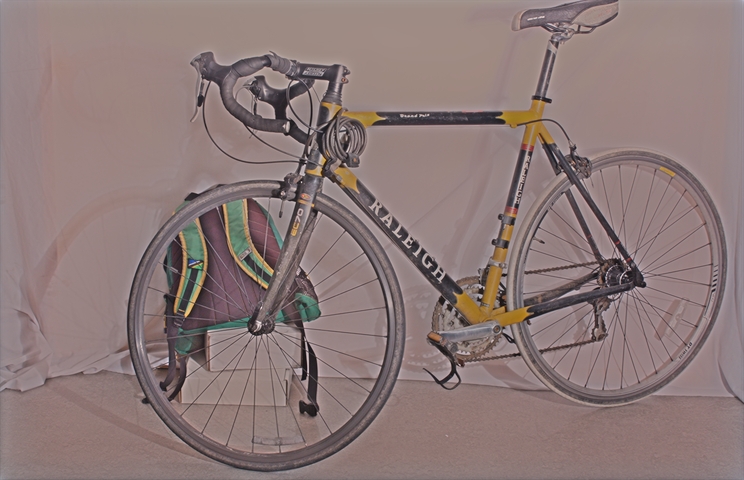}} &
{\includegraphics[height=1.35cm]{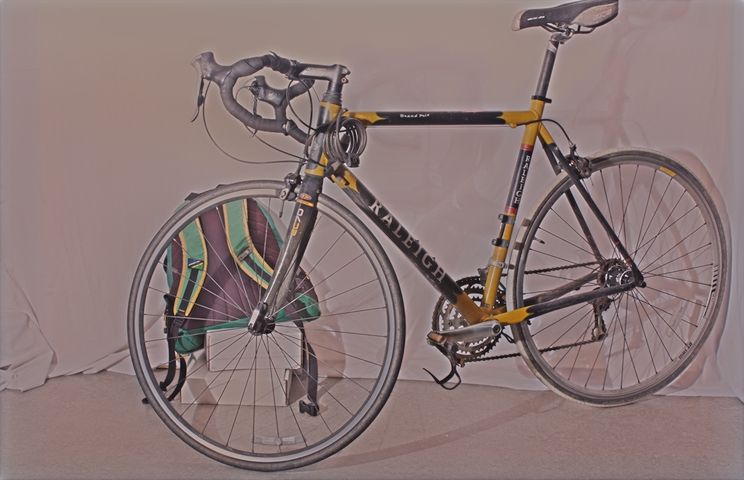}} &
{\includegraphics[height=1.35cm]{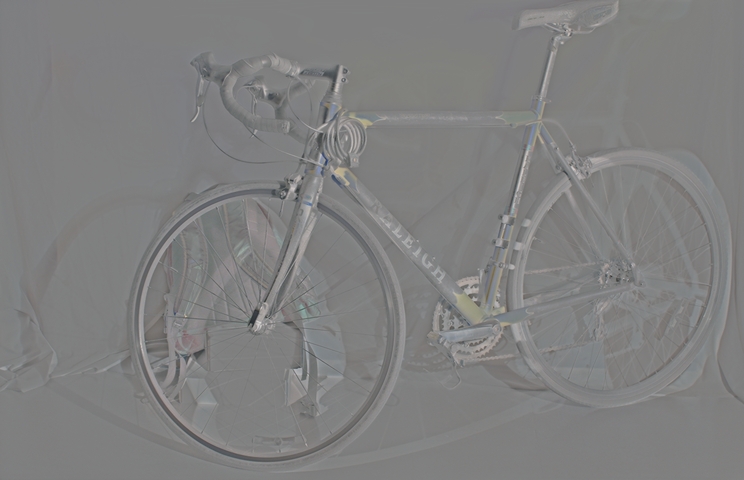}} &
&
{\includegraphics[height=1.35cm]{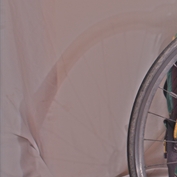}} &
{\includegraphics[height=1.35cm]{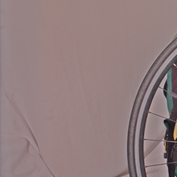}} &
{\includegraphics[height=1.35cm]{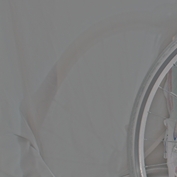}} 
\\
\capstyle{Li-Snavely} &
{\includegraphics[height=1.35cm]{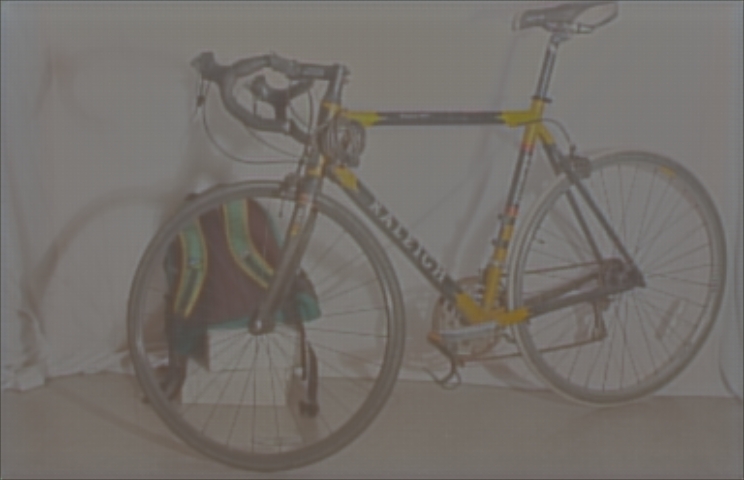}} &
{\includegraphics[height=1.35cm]{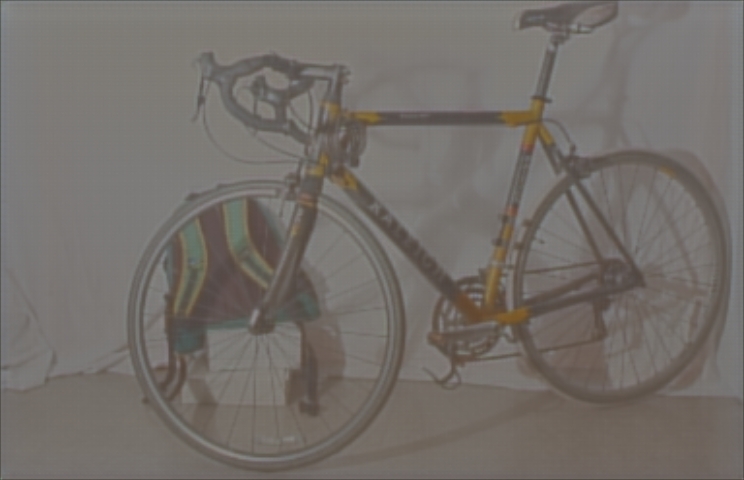}} &
{\includegraphics[height=1.35cm]{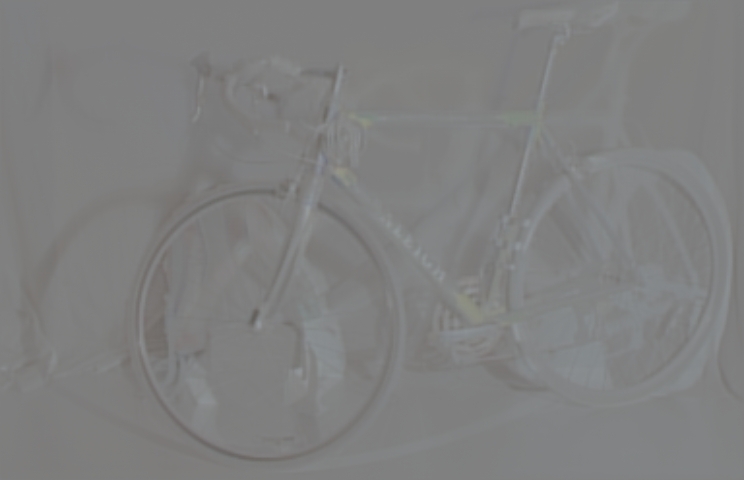}} &
&
{\includegraphics[height=1.35cm]{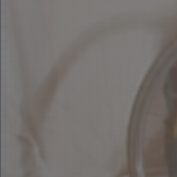}} &
{\includegraphics[height=1.35cm]{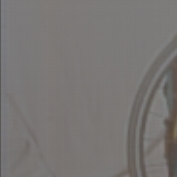}} &
{\includegraphics[height=1.35cm]{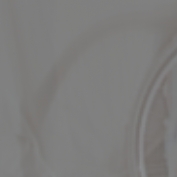}} 
\\
\capstyle{Maddern} &
{\includegraphics[height=1.35cm]{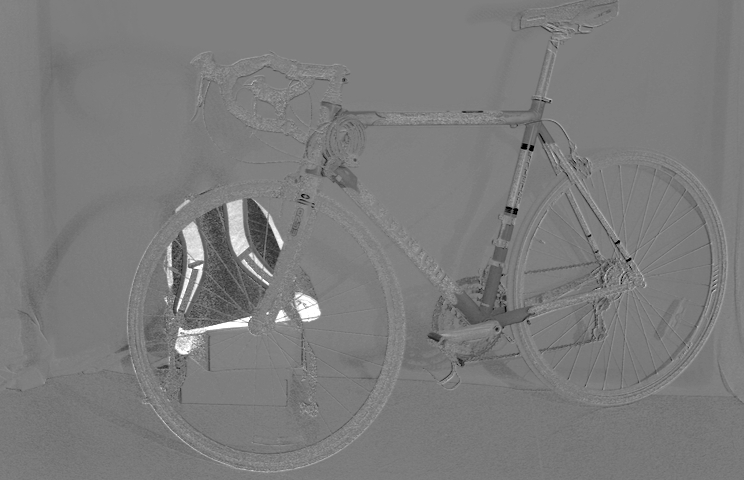}} &
{\includegraphics[height=1.35cm]{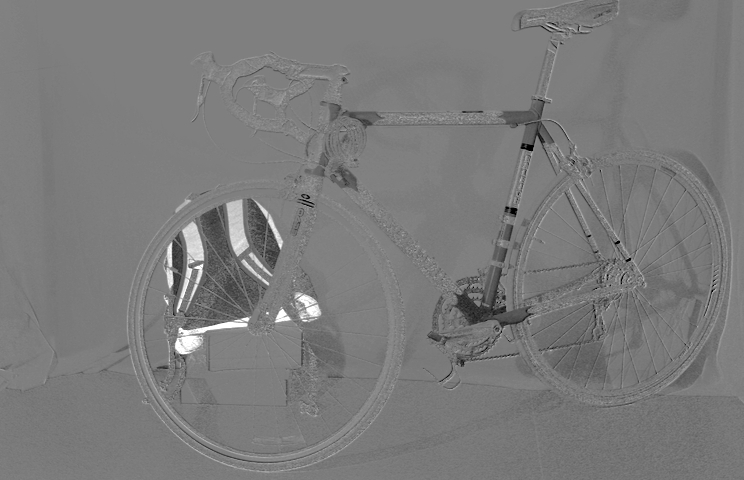}} &
{\includegraphics[height=1.35cm]{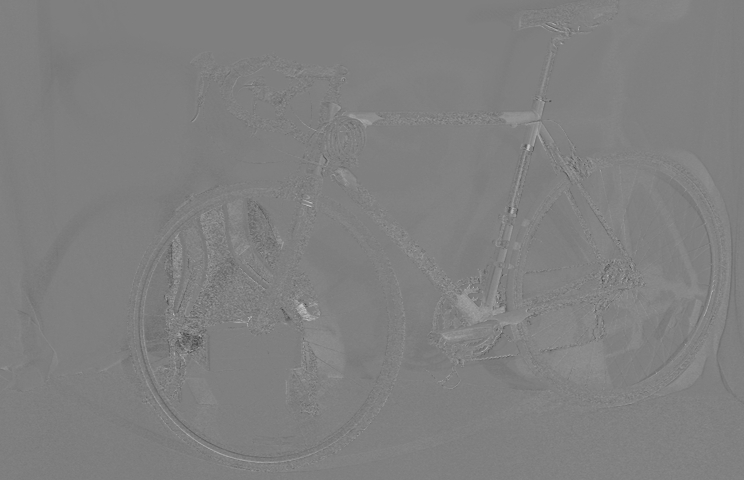}} &
&
{\includegraphics[height=1.35cm]{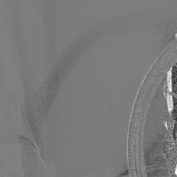}} &
{\includegraphics[height=1.35cm]{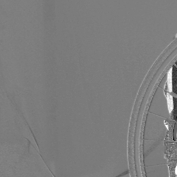}} &
{\includegraphics[height=1.35cm]{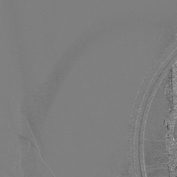}} 
\\
\capstyle{Ours} &
{\includegraphics[height=1.35cm]{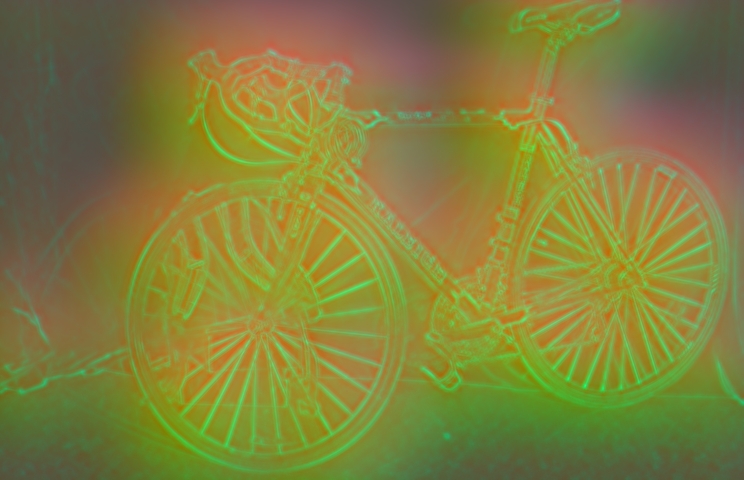}} &
{\includegraphics[height=1.35cm]{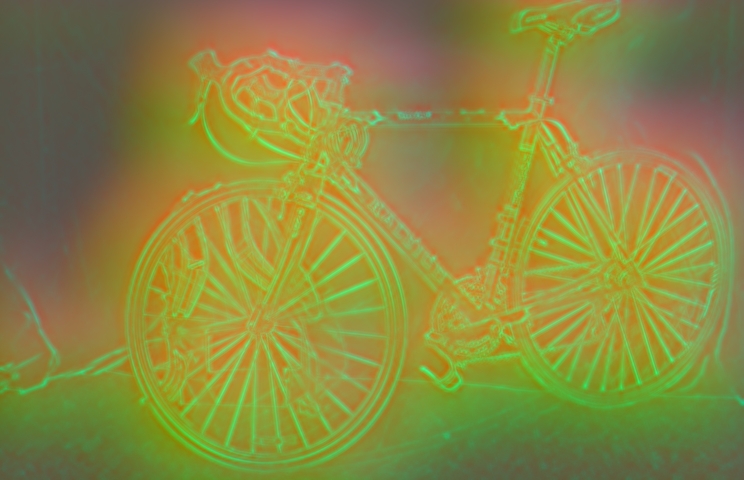}} &
{\includegraphics[height=1.35cm]{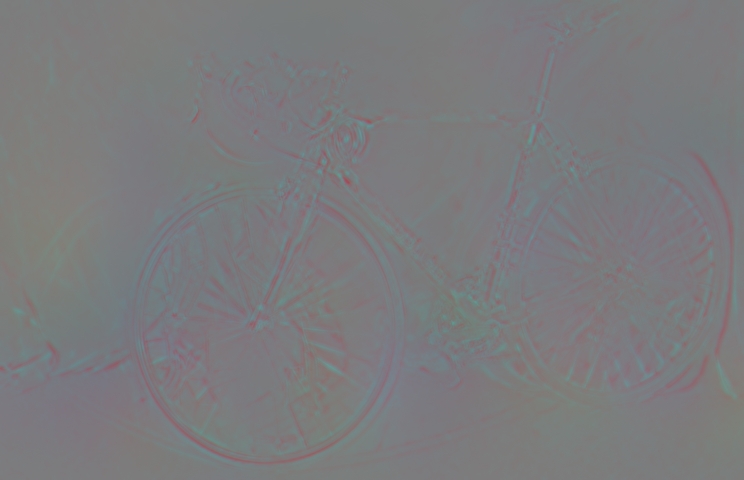}} &
&
{\includegraphics[height=1.35cm]{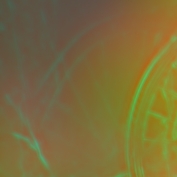}} &
{\includegraphics[height=1.35cm]{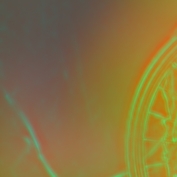}} &
{\includegraphics[height=1.35cm]{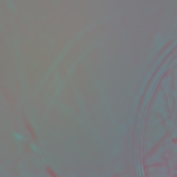}} 
\\
&
\capstyle{Image 1} &
\capstyle{Image 2} &
\capstyle{Difference} &
&
\capstyle{Image 1 Crop} &
\capstyle{Image 2 Crop} &
\capstyle{Difference} \cr
\vspace{0.5cm}
\end{tabular}

\caption{Visual comparison of invariant representation methods. A scene is shown under different illumination conditions. The difference (ideally zero) affirms that our representation is highly stable under illumination
changes (zero is gray). Enlarged crops of the marked red-square are shown on the right.
}
\label{fig:visual_diff_MB}
\end{center}
\end{figure}

\begin{figure}
\setlength{\tabcolsep}{0.1em} 
\begin{center}
\begin{tabular}{ACCCXBBB}
\capstyle{Original image} &
{\includegraphics[height=1.35cm]{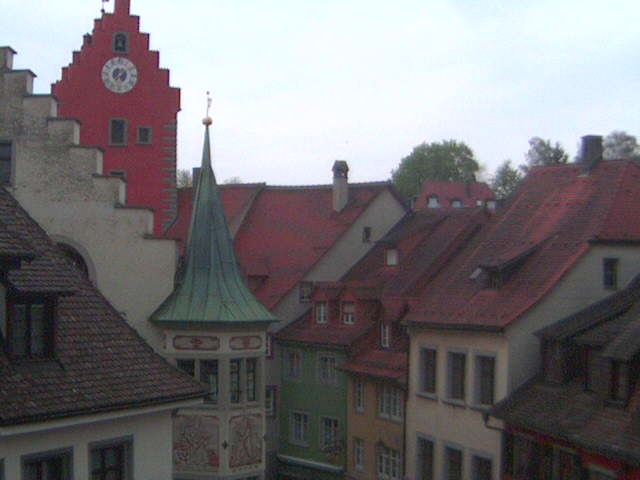}} &
{\includegraphics[height=1.35cm]{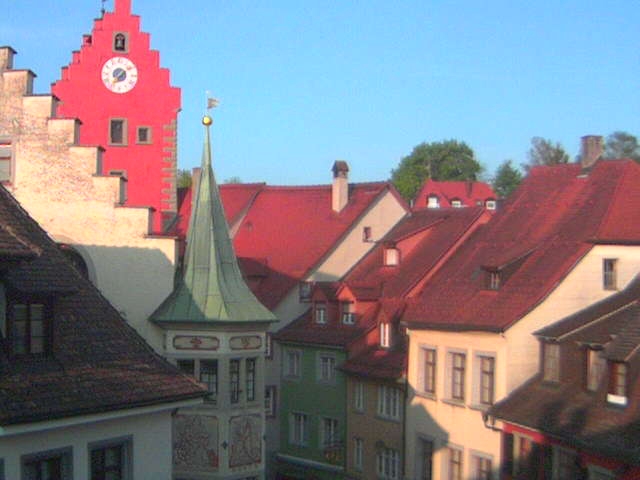}} &
{\includegraphics[height=1.35cm]{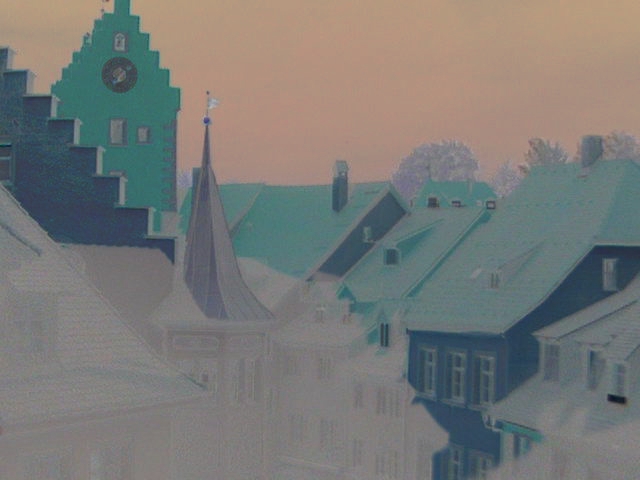}} &
&
{\includegraphics[height=1.35cm]{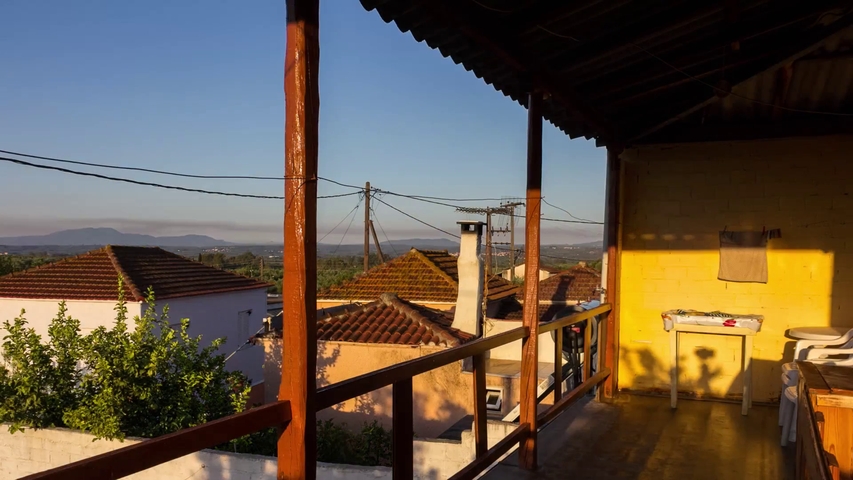}} &
{\includegraphics[height=1.35cm]{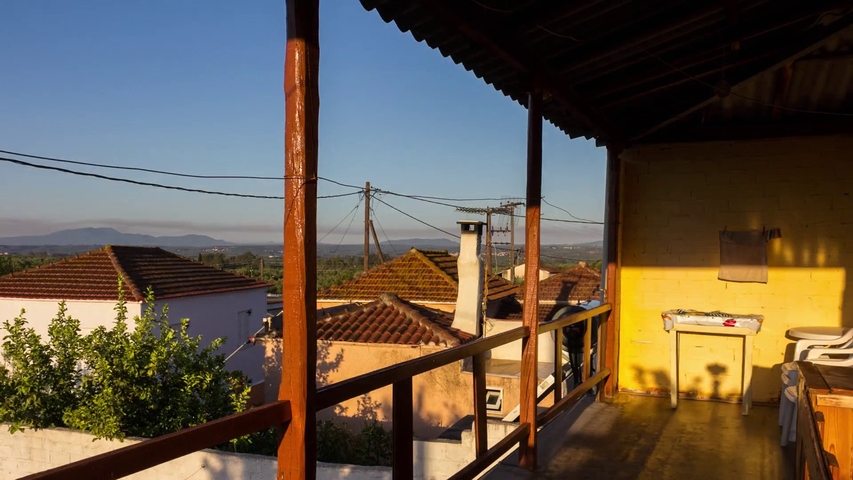}} &
{\includegraphics[height=1.35cm]{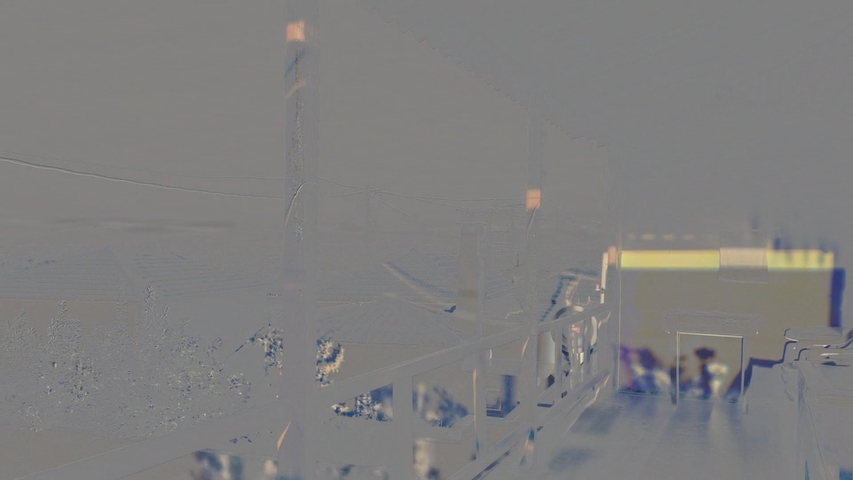}} 
\\
\capstyle{Lettry} &
{\includegraphics[height=1.35cm]{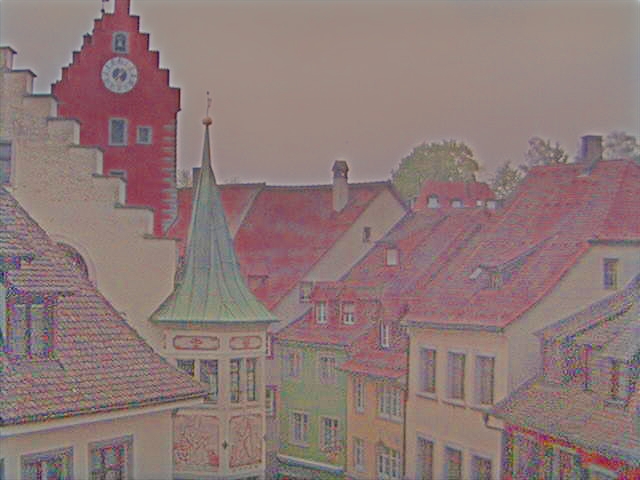}} &
{\includegraphics[height=1.35cm]{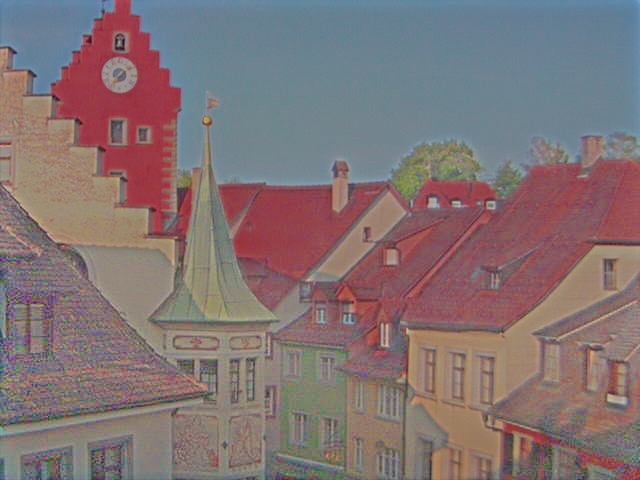}} &
{\includegraphics[height=1.35cm]{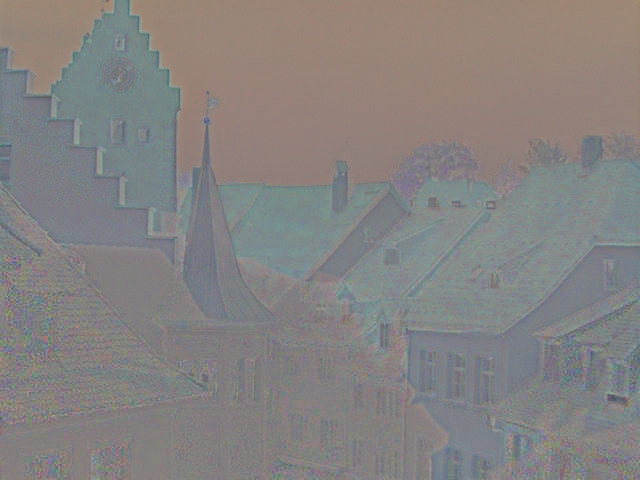}} &
&
{\includegraphics[height=1.35cm]{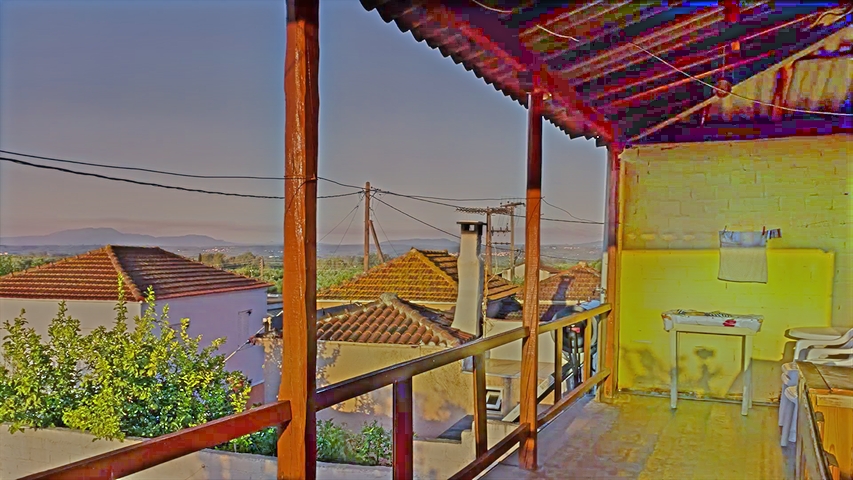}} &
{\includegraphics[height=1.35cm]{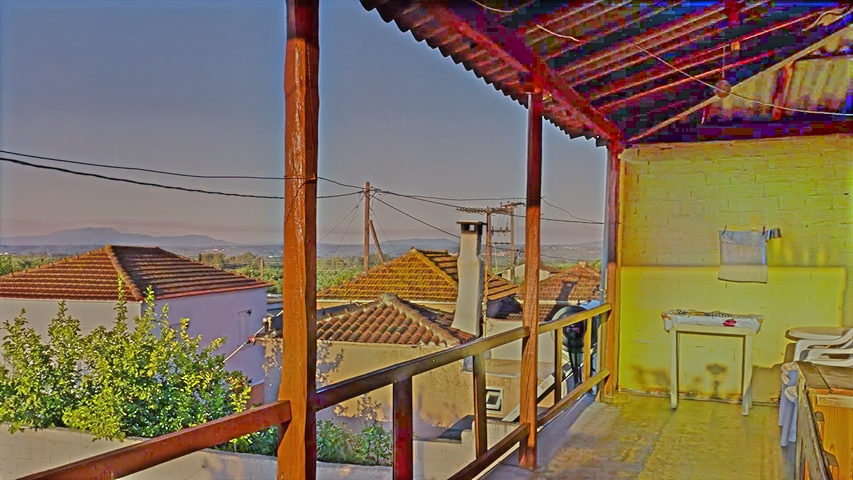}} &
{\includegraphics[height=1.35cm]{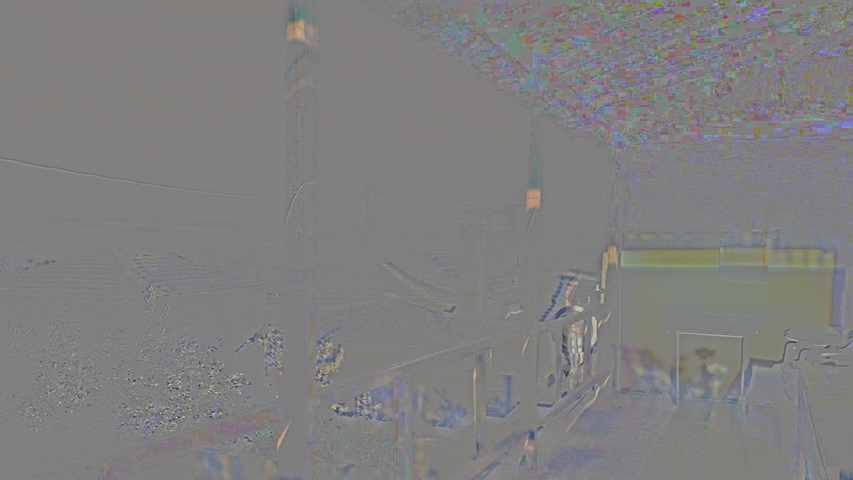}} 
\\
\capstyle{Li-Snavely} &
{\includegraphics[height=1.35cm]{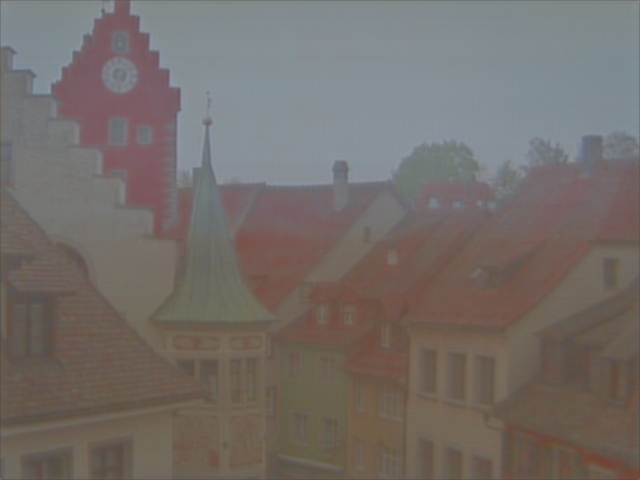}} &
{\includegraphics[height=1.35cm]{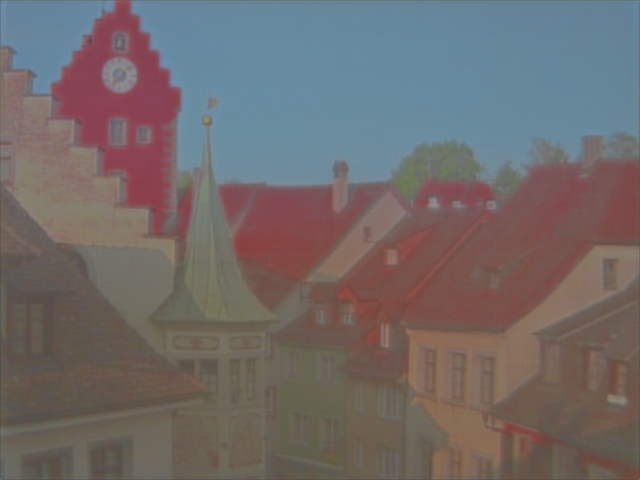}} &
{\includegraphics[height=1.35cm]{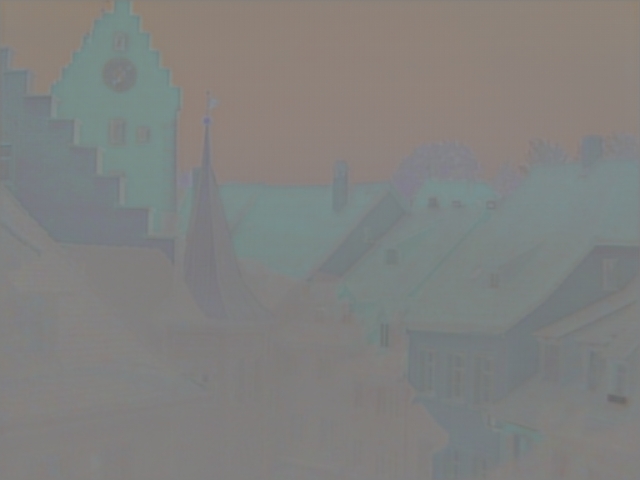}} &
&
{\includegraphics[height=1.35cm]{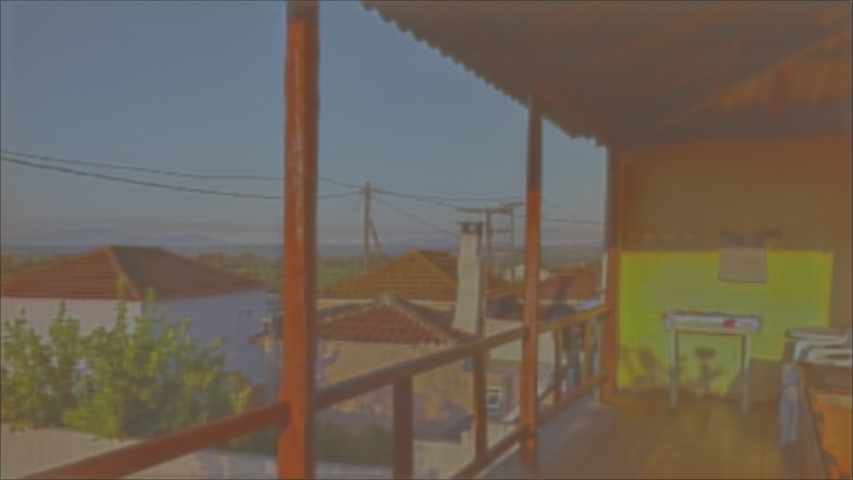}} &
{\includegraphics[height=1.35cm]{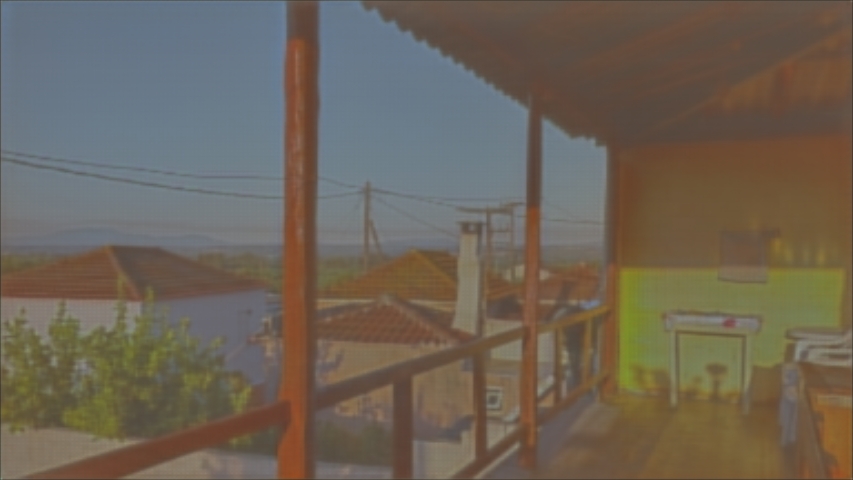}} &
{\includegraphics[height=1.35cm]{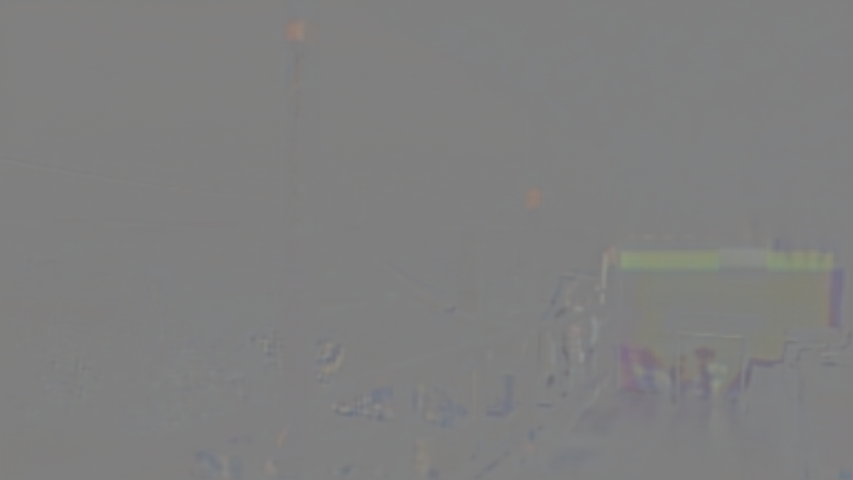}} 
\\
\capstyle{Maddern} &
{\includegraphics[height=1.35cm]{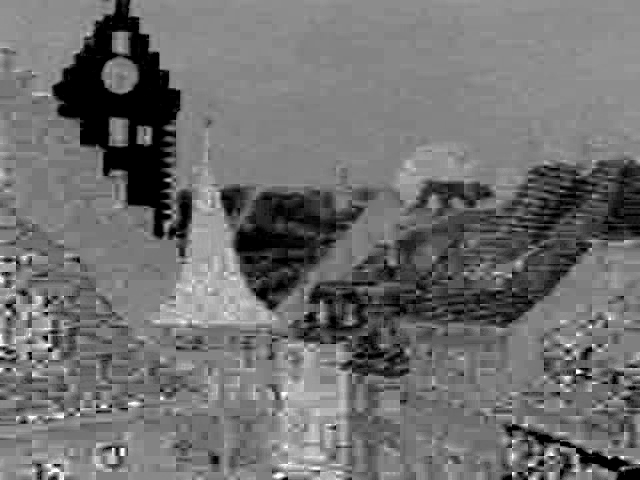}} &
{\includegraphics[height=1.35cm]{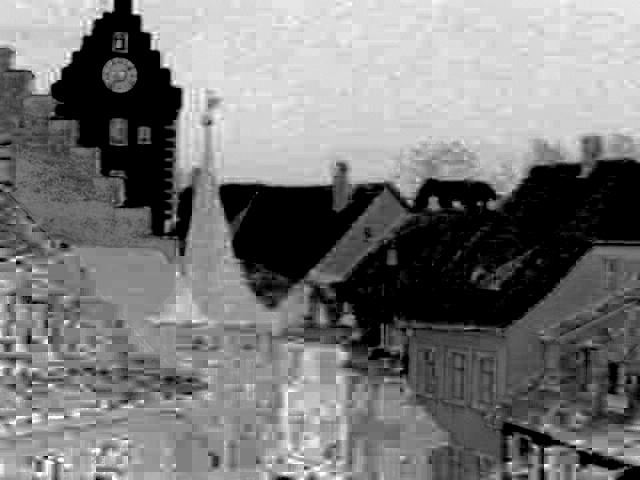}} &
{\includegraphics[height=1.35cm]{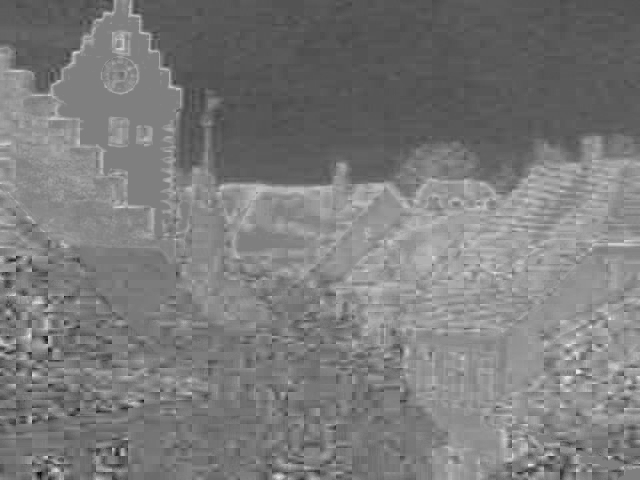}} &
&
{\includegraphics[height=1.35cm]{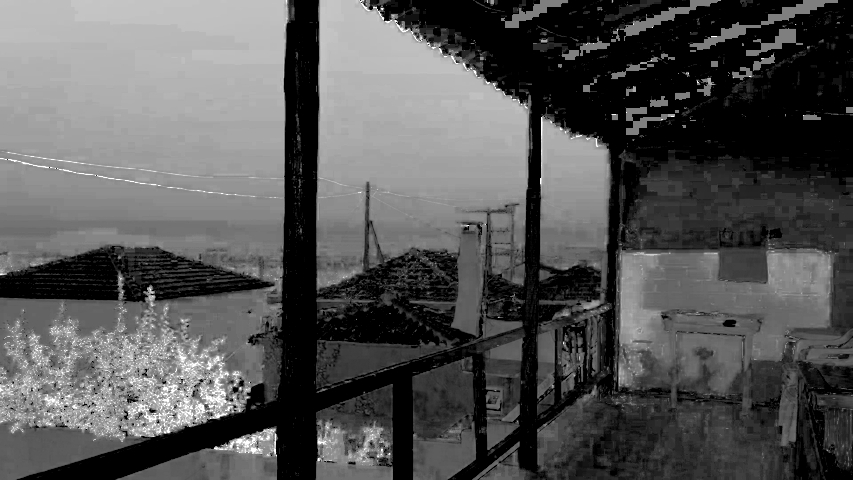}} &
{\includegraphics[height=1.35cm]{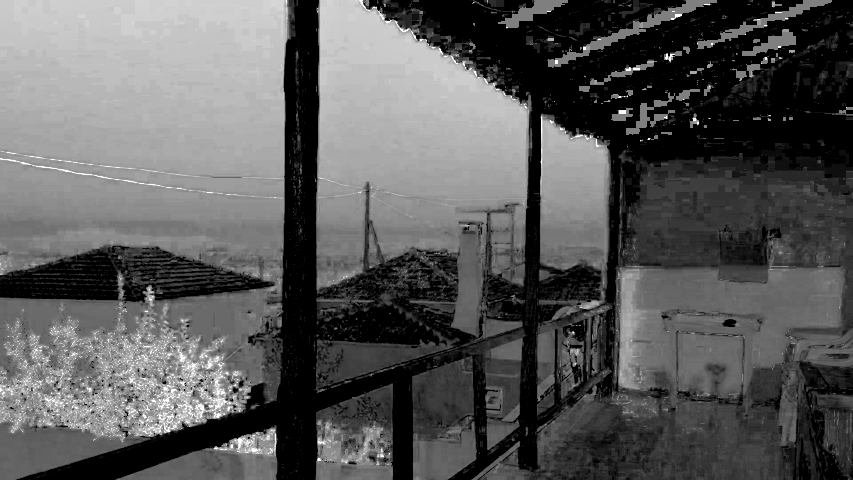}} &
{\includegraphics[height=1.35cm]{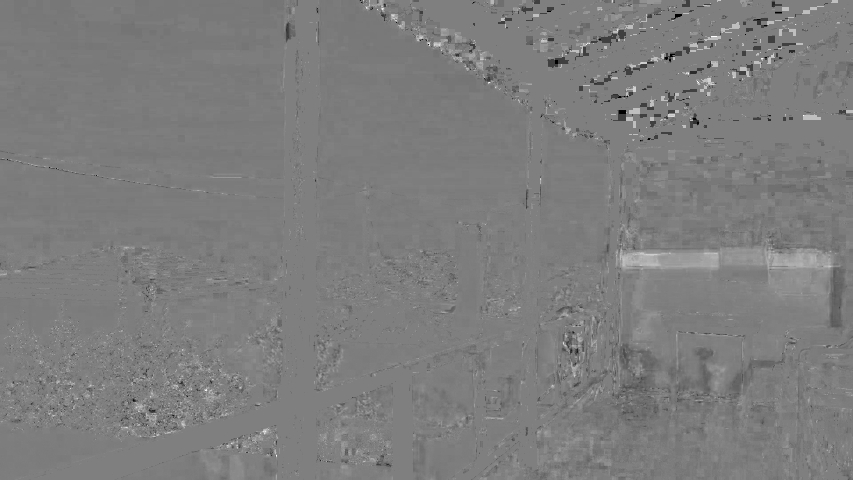}} 
\\
\capstyle{\bf{Ours}} &
{\includegraphics[height=1.35cm]{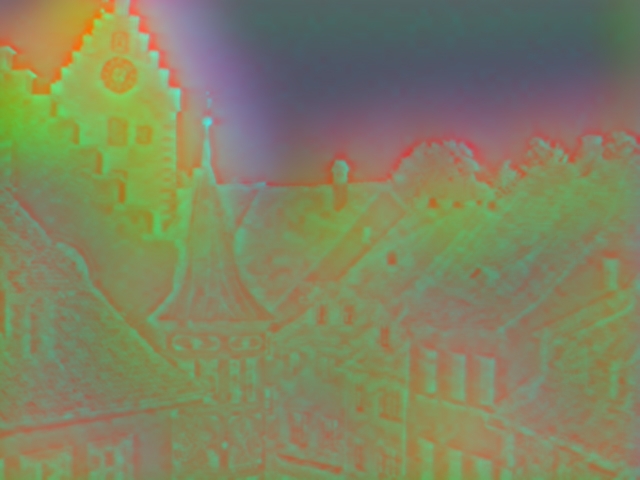}} &
{\includegraphics[height=1.35cm]{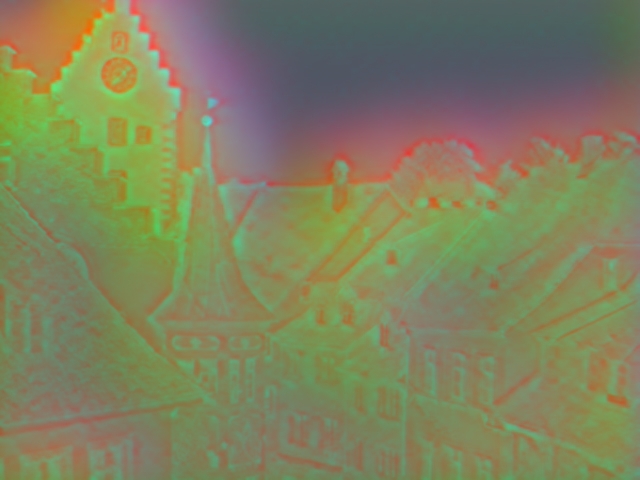}} &
{\includegraphics[height=1.35cm]{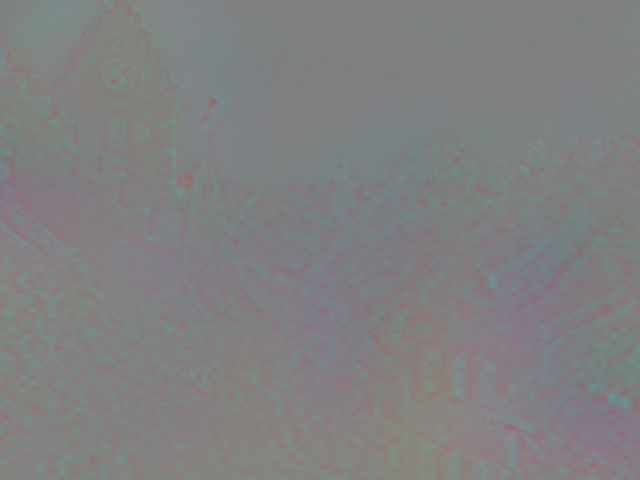}} &
&
{\includegraphics[height=1.35cm]{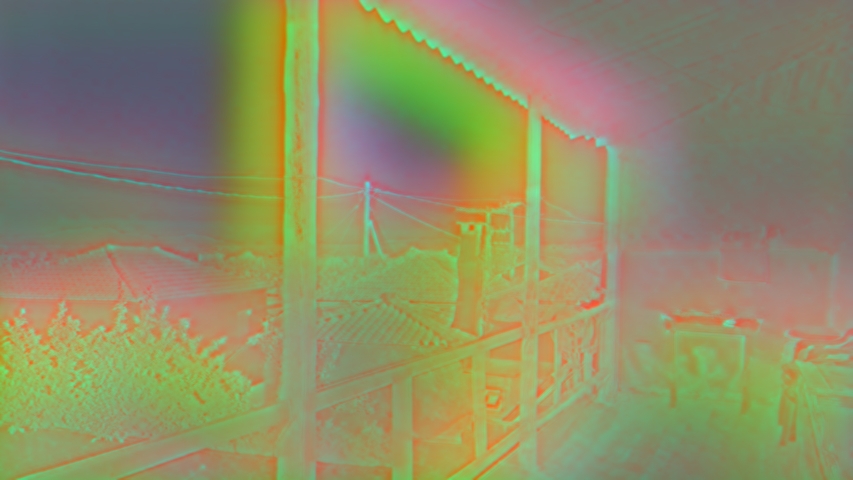}} &
{\includegraphics[height=1.35cm]{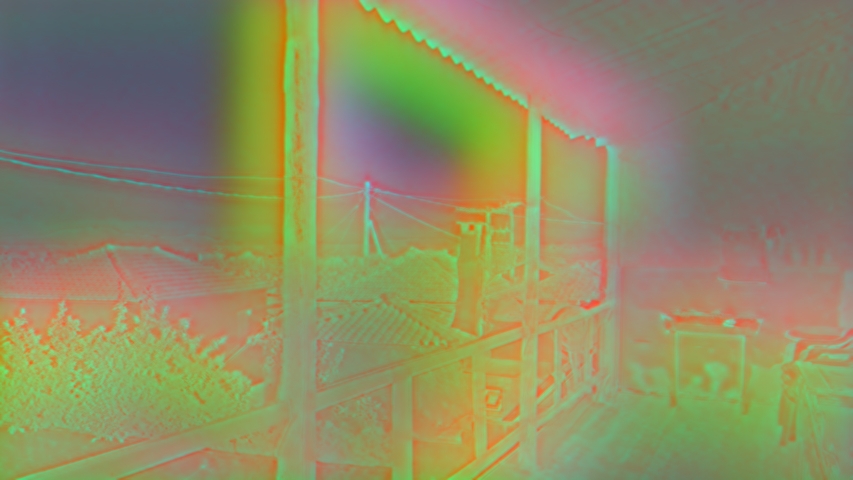}} &
{\includegraphics[height=1.35cm]{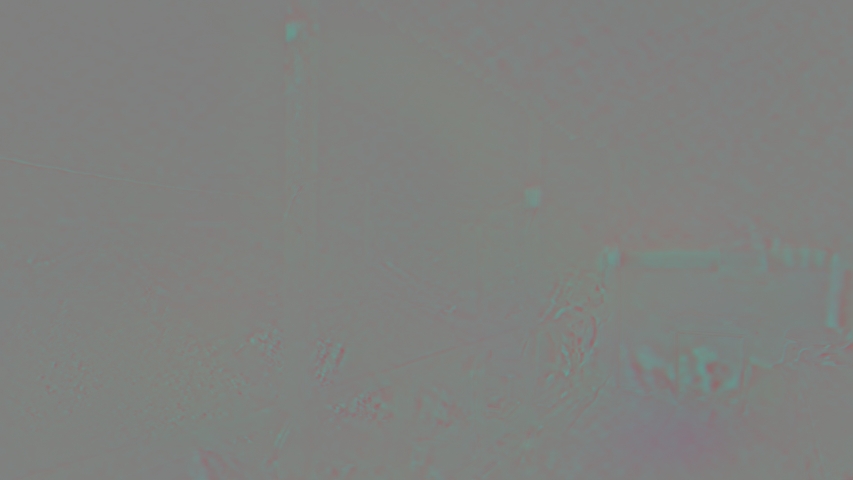}}
\\
&
\capstyle{Image 1} &
\capstyle{Image 2} &
\capstyle{Difference} &
&
\capstyle{Image 1} &
\capstyle{Image 2} &
\capstyle{Difference} 
\end{tabular}
\caption{
Visual comparison of invariant representation methods. For each scene, the representation of two images under different illumination conditions is shown. The difference (ideally zero) affirms that our representation is highly stable under illumination changes.}
\label{fig:visual_diff}
\end{center}
\end{figure}

\smallskip\noindent\textbf{Visual Photo-consistency.}
In Fig. \ref{fig:diff} we show that actual scene changes (the car) can be clearly seen, while illumination differences are attenuated. We compare this change detection experiment to the original images and to the representation of \cite{li2018learning}.
In Fig. \ref{fig:visual_diff} two examples of image pairs from the same scene under different illumination conditions are shown. All image channels in all representations are in the range $[0,255]$.
Our representation has the lowest difference compared to other invariant representations (since the range is similar, no scaling is performed on the difference images).
In order to demonstrate that an unconstrained model can be beneficial compared to the standard intrinsic decomposition (albedo-shading) we show in Fig. \ref{fig:visual_diff_MB} an example of an indoor scene from the Middlebury dataset  \cite{scharstein2014high}.
It can be seen that our representation handles well strong shades on the wall as well as shiny metal, which violates the Lambertian model.

\clearpage
\subsection{Quantitative Evaluation}
\label{sec:quantitive}
We test our representation using two common computer vision tasks, and compare it with two unsupervised data-driven SIID methods, \textit{Li-Snavely} \cite{li2018learning} and \textit{Lettry} \cite{lettry2018unsupervised}, an analytic grayscale representation, \textit{Maddern}, \cite{maddern2014illumination} and also with the \textit{Original image}.
In all cases the different representations are used as a pre-processing stage.

\bigskip
\noindent\textbf{Patch Matching.}
In this task a template patch is selected from a reference image and the aim is to find its location in a target image. Both images are of the same scene but with different illumination conditions.
We used the standard template matching function of OpenCV \cite{opencv_library}, \textit{matchTemplate}, with the normalized cross correlation method.
From each reference image, 10 square patches are randomly selected from significant areas in the image, by setting a minimum standard deviation of 25 (image range is $[0,255]$). 
This is done for three different patch sizes (32, 64 and 128 pixels).
We compare our results to others invariant representation (using the same matching algorithm) and also to state-of-the-art dedicated template matching algorithms: QATM \cite{cheng2019qatm} and DDIS \cite{talmi2017template}. The latter are novel  algorithms, based on deep features.

\begin{figure}[h]
\begin{center}
\setlength{\tabcolsep}{0.22em} 
\begin{tabular}{ccccccc}
\centering
\multirow{4}{*}{\includegraphics[width=0.07\linewidth]{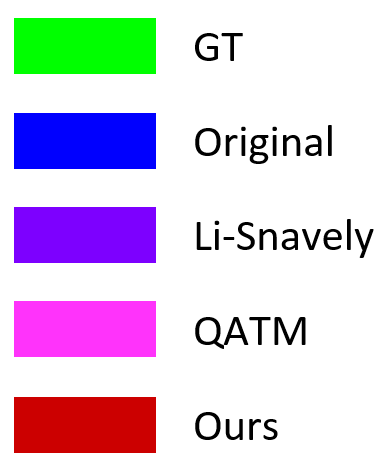}}&
{\includegraphics[width=0.14\linewidth]{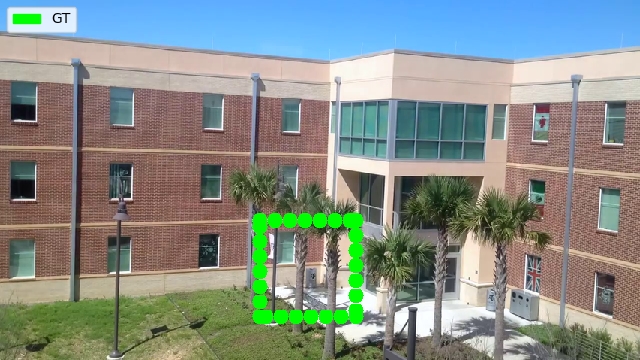}} &
{\includegraphics[width=0.14\linewidth]{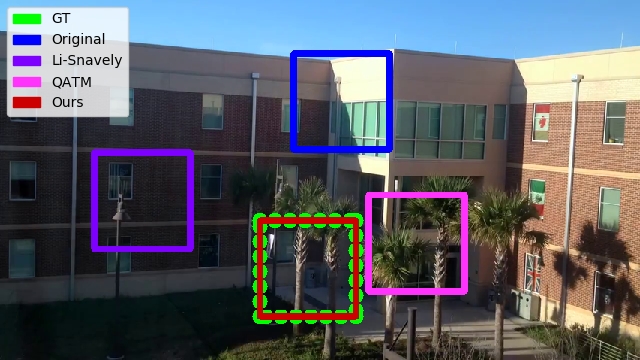}} &
{\includegraphics[width=0.14\linewidth]{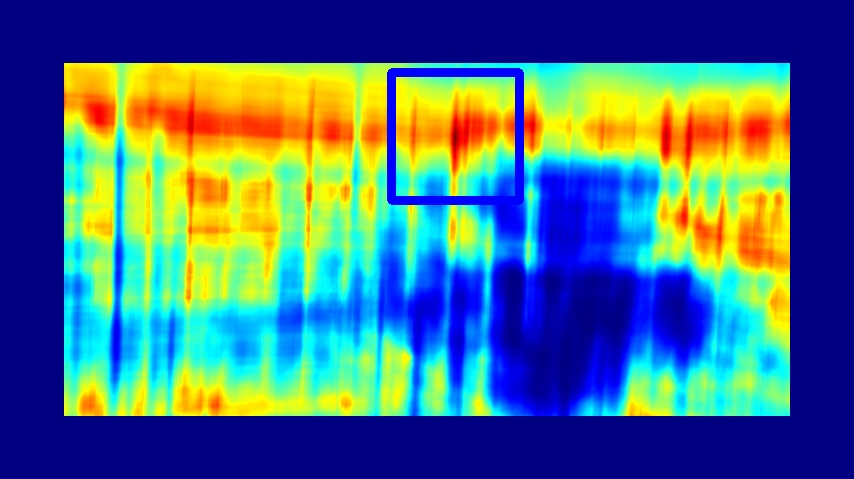}} &
{\includegraphics[width=0.14\linewidth]{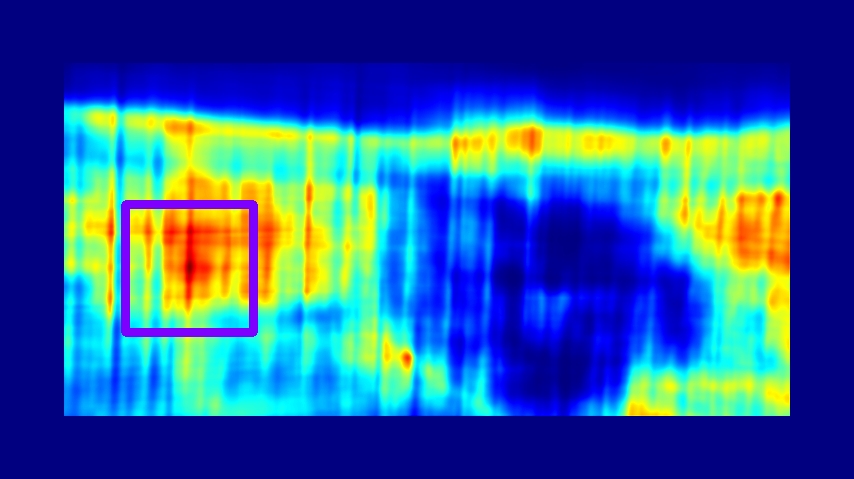}} &
{\includegraphics[width=0.14\linewidth]{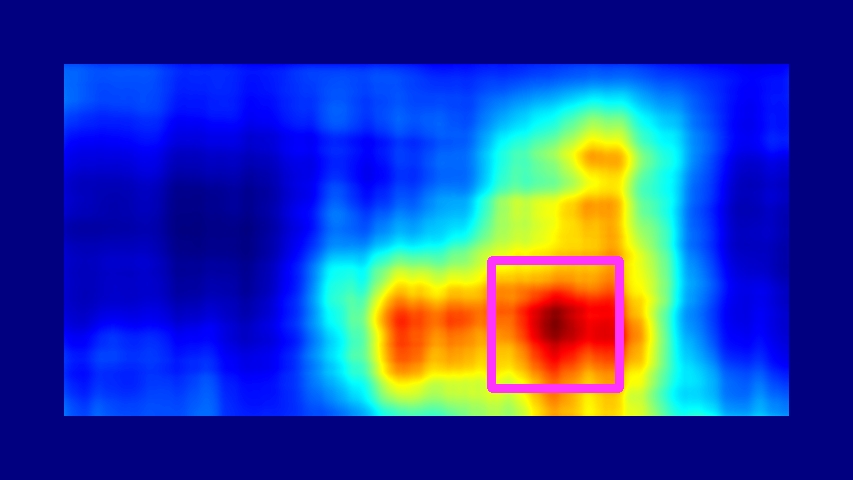}}& 
{\includegraphics[width=0.14\linewidth]{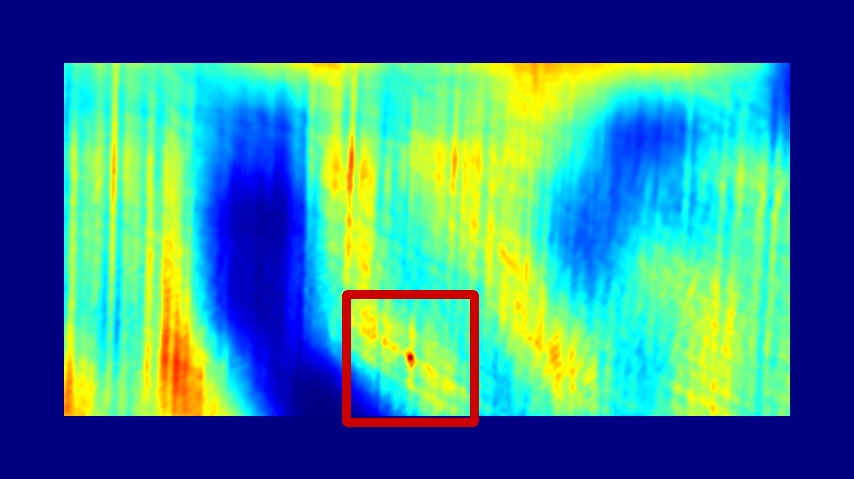}} 
\\
&
{\includegraphics[width=0.14\linewidth]{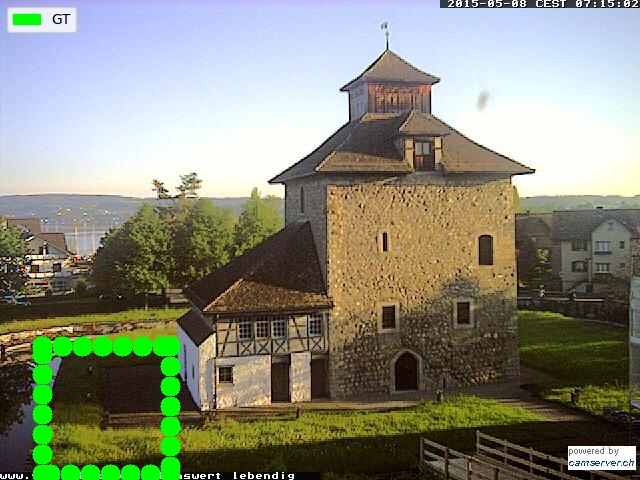}} &
{\includegraphics[width=0.14\linewidth]{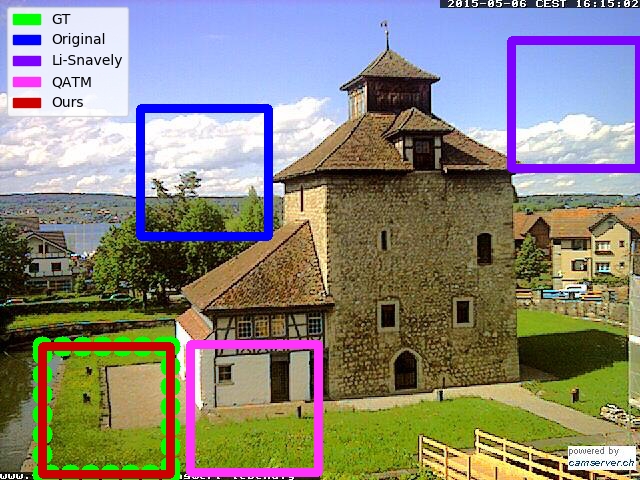}} &
{\includegraphics[width=0.14\linewidth]{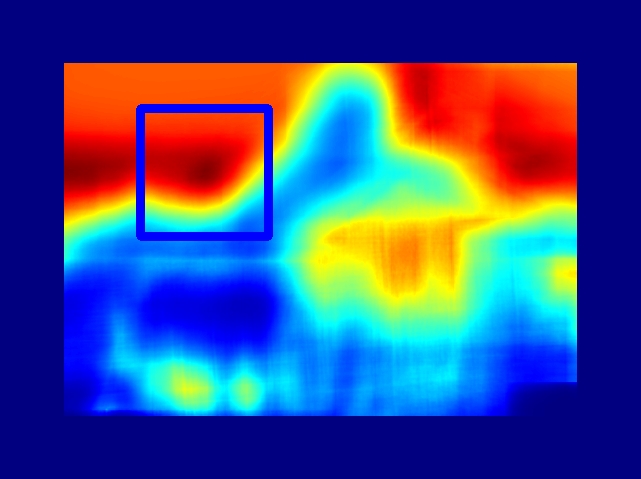}} &
{\includegraphics[width=0.14\linewidth]{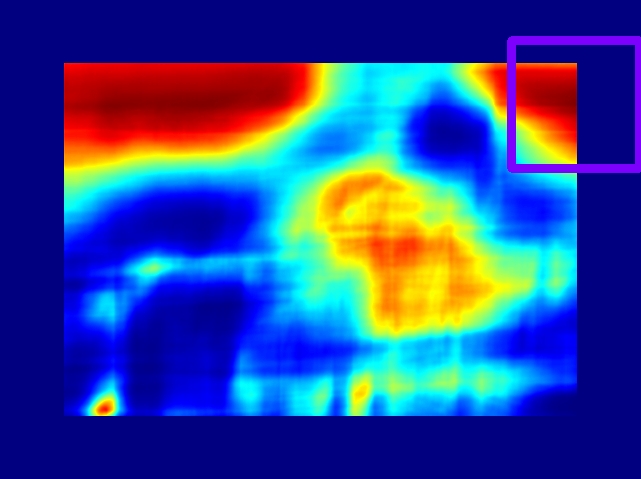}} &
{\includegraphics[width=0.14\linewidth]{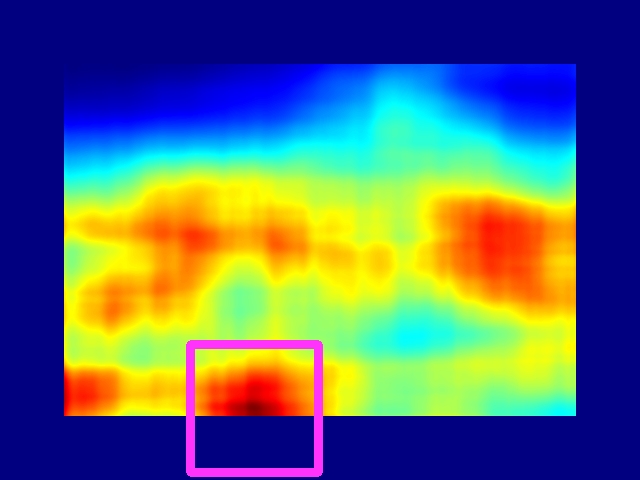}}&
{\includegraphics[width=0.14\linewidth]{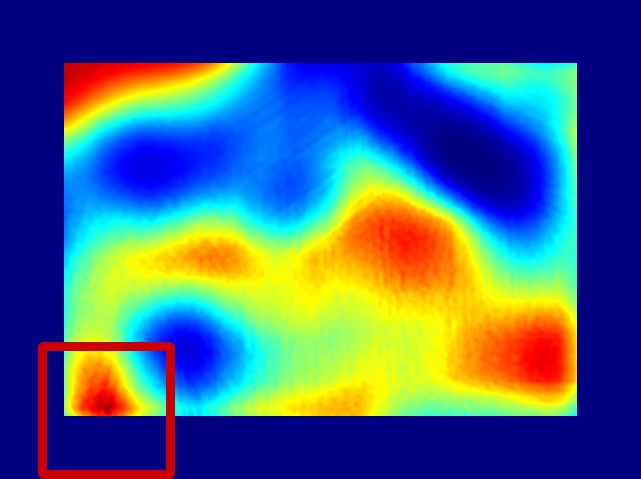}} 
\\
&
\capstyle{Reference Image} &
\capstyle{Target Image} &
\capstyle{Original} &
\capstyle{Li-Snavely\cite{li2018learning}} &
\capstyle{QATM\cite{cheng2019qatm}}&
\capstyle{Ours}
\\
\end{tabular}
\caption{Patch matching of challenging scenes. Reference image on left (green frame). The frames marking the algorithms' results are overlaid on the target image. Heatmaps of each algorithm (right) indicate high (red) to low (blue) matching scores.}
\label{fig:pm_heatmaps}
\end{center}
\end{figure}

\begin{figure}
\centering
\subfloat{%
       \includegraphics[trim={0.4cm 0.2cm 1.6cm 0.8cm},clip,width=0.45\linewidth]{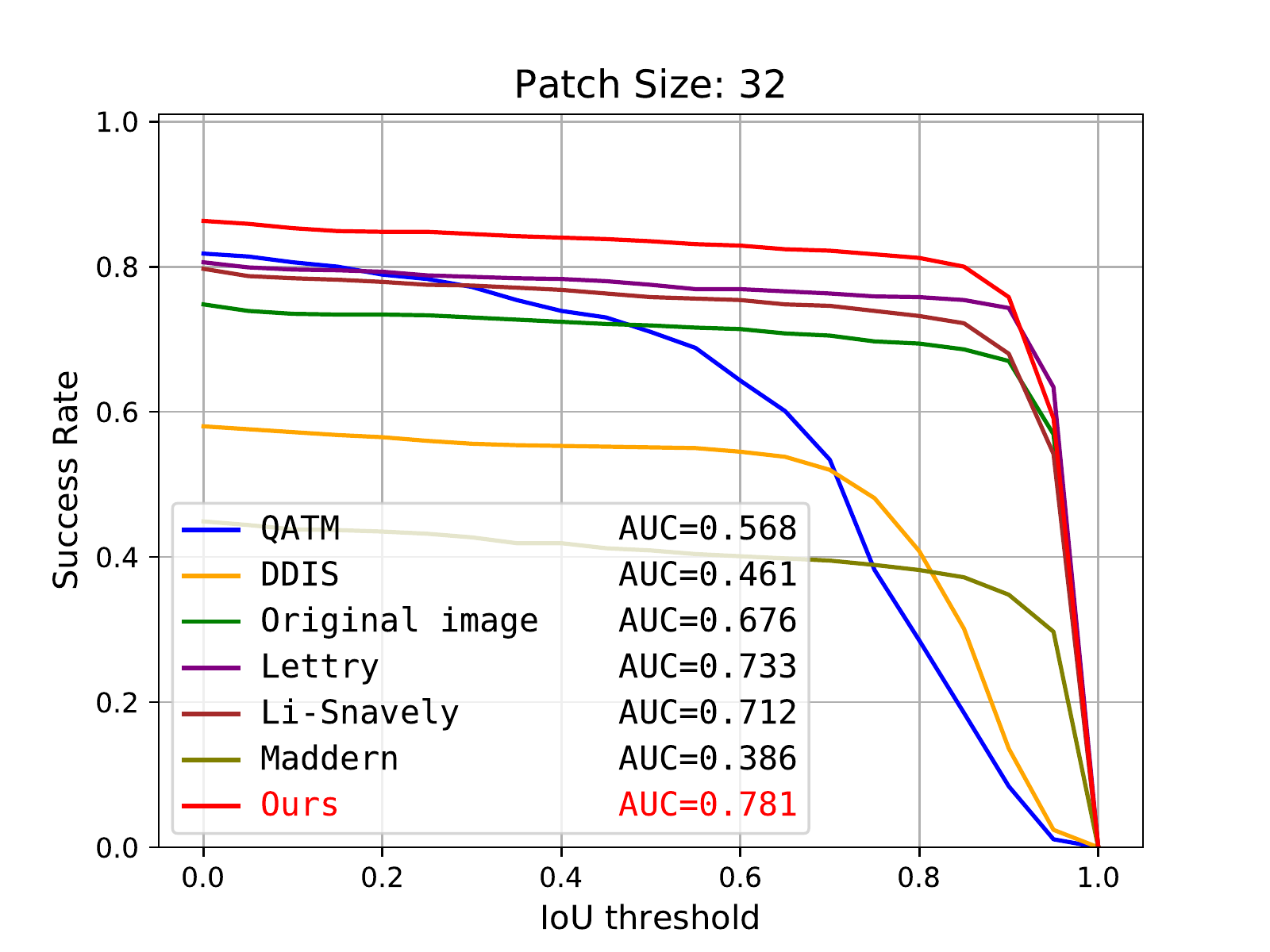}}
\subfloat{%
       \includegraphics[trim={0.4cm 0.2cm 1.6cm 0.8cm},clip,width=0.45\linewidth]{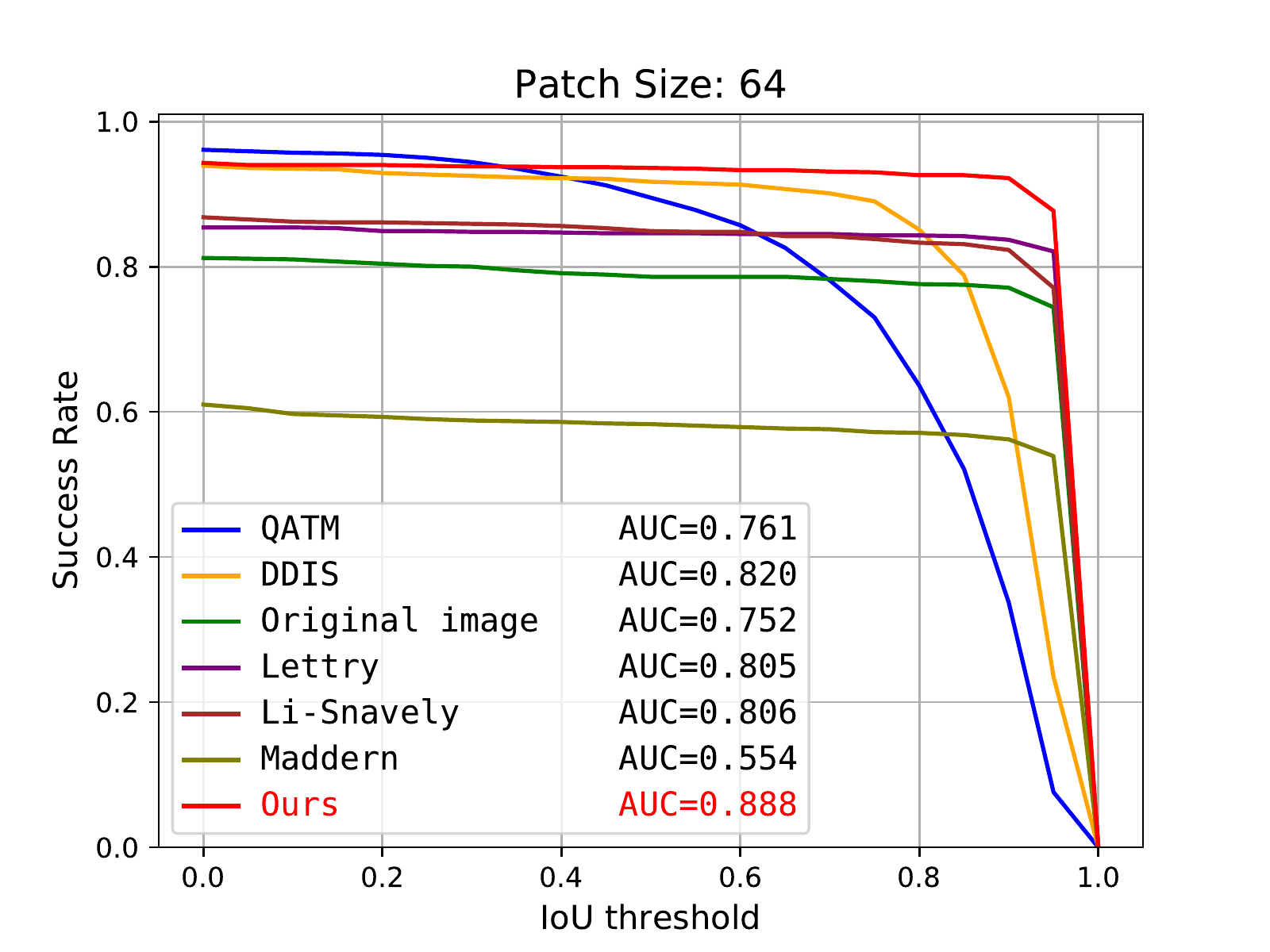}}
       \\
\subfloat{%
       \includegraphics[trim={0.4cm 0.2cm 1.6cm 0.8cm},clip,width=0.45\linewidth]{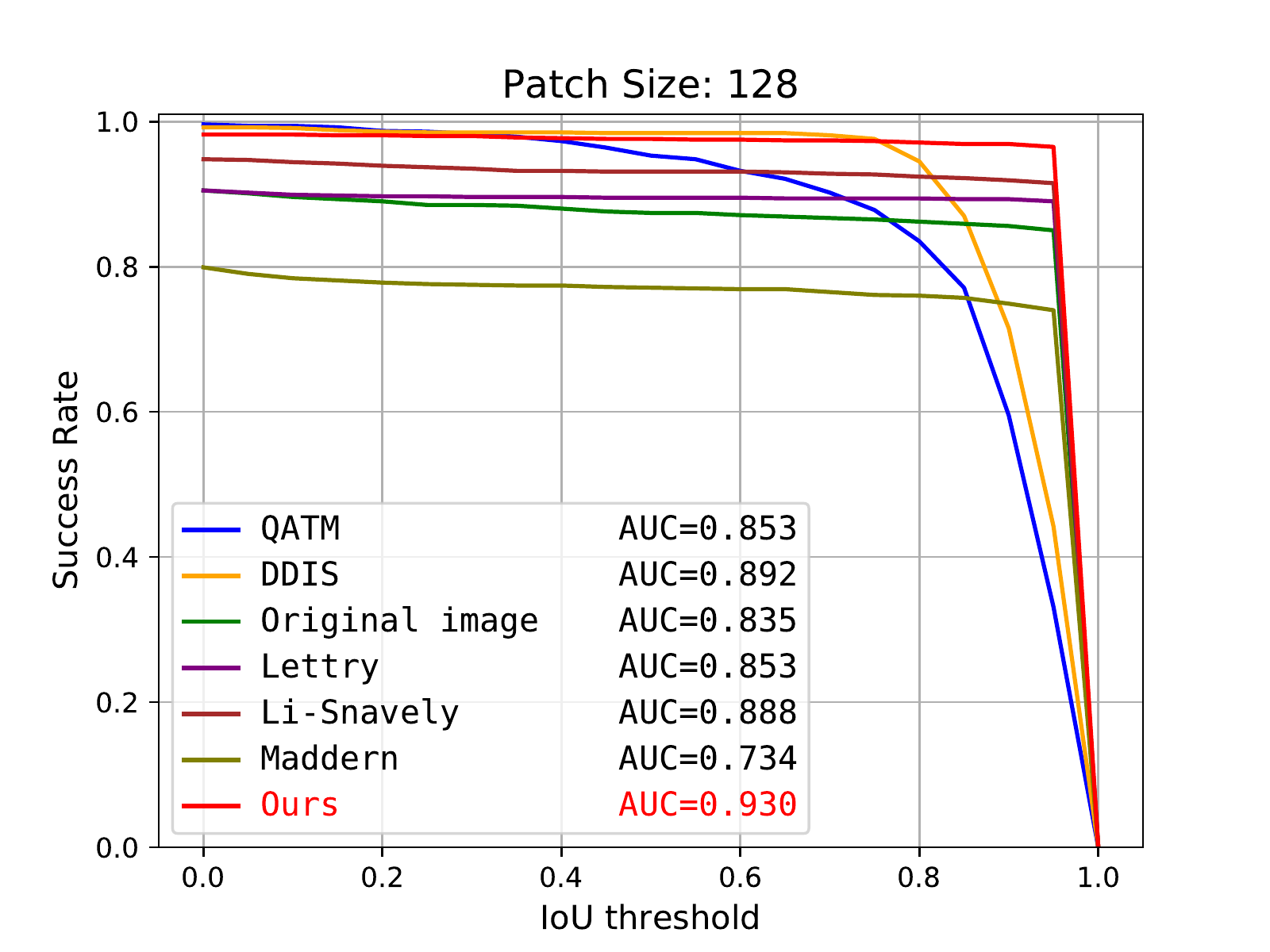}}
\caption{Patchch matching results: IoU-ROC curves and AUC scores.}
\label{fig:pm}
\end{figure}

In Fig. \ref{fig:pm_heatmaps} some matching results are shown, along with correlation-based heatmaps, which correspond to the closest match. Whereas the algorithms are generally robust to  minor illumination changes, in these challenging cases, only our proposed transform succeeds.
In Fig. \ref{fig:pm} and Table \ref{table:pm} results of extensive quantitative experiments are shown. 
Accuracy is measured by intersection over union  (IoU). Plots show the IoU-ROC curves and the area under the curve (AUC) scores for all algorithms. On the right of Fig. \ref{fig:pm} a summary of the AUC score is given for all patch sizes. Our representation consistently achieves the highest score for all patch sizes. We conclude that using classical patch-matching algorithms jointly with our proposed transform (as pre-processing)  surpasses not only other invariant transforms but also state-of-the-art end-to-end dedicated matching methods. 
\vspace{2em}

\begin{table}
    \centering
    \begin{tabular}{|c|c|c|c|c|c|c|}
        \hline
        \backslashbox[32mm]{Method}{Patch size}& 32 & 64 & 128\\
        \hline
        QATM & 0.568 & 0.761 & 0.853\\
        DDIS & 0.461 & 0.820 & 0.892\\
        Original image & 0.676 & 0.752 & 0.835\\
        Lettry & 0.733  & 0.805 & 0.853 \\
        Li-Snavely & 0.712 & 0.806 & 0.888 \\
        Maddern & 0.386 & 0.554 & 0.734  \\
        {\bf Ours} & {\bf0.781} & {\bf0.888} & {\bf0.930}\\
        \hline
        \end{tabular}
    \label{table:pm}
    \caption{
   Patch matching results: scores for all patch sizes. Experiment on 100 scenes, 10 patches in each scene.}
\end{table}

\begin{figure}
\centering
\subfloat[\label{fig:reg_a}]{%
       \includegraphics[width=0.44\linewidth]{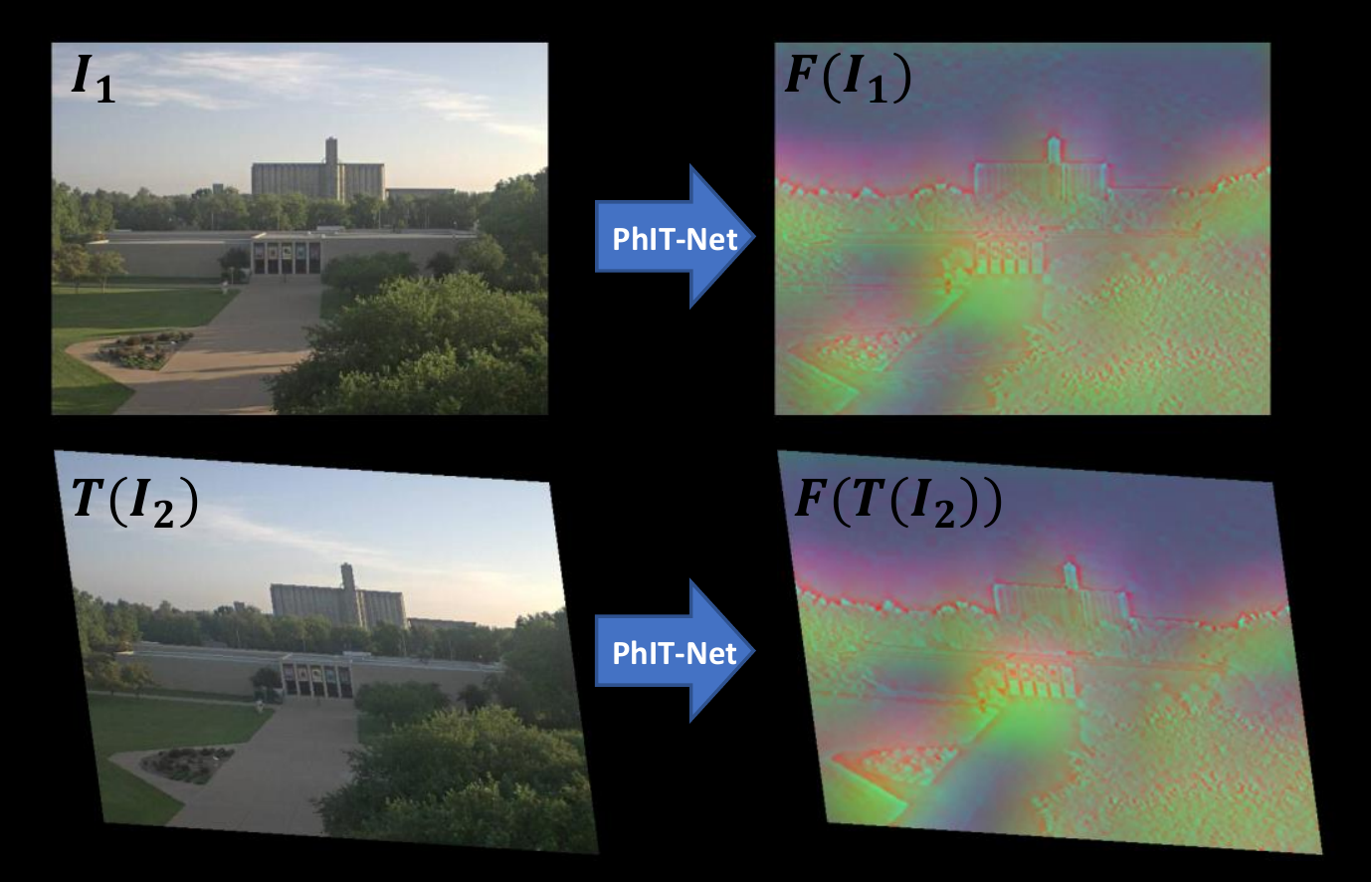}}
\hspace{0.2cm}
\subfloat[\label{fig:reg_b}]{%
        \adjustbox{raise=4em}{
        \renewcommand{\arraystretch}{0.9}
        \begin{tabular}{|c|c|c|c|c|}
        \hline
        \backslashbox[23mm]{Method}{Angle}& 2 & 10 & 18 & 26\\
        \hline
        Ground Truth  & 35.99  & 34.74 & 32.75 & 31.99\\
        \hline
        Original image & 19.85 & 21.22 & 21.79 & 20.97\\
        Lettry & 21.13  & 21.20 & 17.98 & 15.26\\
        Li-Snavely & 26.51 & 28.22 & 27.86 & 27.22\\
        Maddern & 22.80 & 24.21 & 22.56 & 19.49 \\
        {\bf Ours} & {\bf28.71} & {\bf30.10} & {\bf30.04} & {\bf29.53}\\
        \hline
        \end{tabular}}}
\caption{(a) Illustration of the registration test. Top left -- Reference image ($I_1$); bottom left -- Target image ($I_2$) transformed by a rigid transformation $T(\cdot)$; right -- the respective photo-consistent transform. (b) PSNR results of the registration test.}
\label{fig:reg}
\end{figure}

\newpage
\noindent\textbf{Registration.}
The rigid registration test is performed based on two images (reference and target) of the same scene under different illumination. An affine transformation is applied to the target image (see an example in Fig. \ref{fig:reg_a}). The goal of the registration algorithm is to estimate the reverse affine transformation matrix which aligns the transformed target and reference images. It is expected that an illumination invariant representation can improve the algorithm's accuracy. 
Registration is performed by ECC registration \cite{evangelidis2008parametric}, based on cross correlation,
as implemented in the OpenCV library.
In Fig. \ref{fig:reg} (right) the results for various angles are shown. The accuracy is measured by applying the estimated inverse transformation on the (transformed) target and computing the PSNR (with respect to the original target). Note that there are some minor errors also when the inverse transformation is known precisely (referred in the table as ``Ground Truth``), due to numerical errors in applying the affine transformation. Our representation achieves the highest average PSNR.

\section{Conclusion}
A photo-consistent image transform is proposed, based on a data-driven invariant framework. 
The desired invariance property is learnt, while retaining geometrical coherence. We show that general and simple axioms yield state-of-the-art results, without resorting to over-simplified model constraints. 
Excellent matching and registration results are obtained by combining fast  classical algorithms with our representation, also in extreme lighting variations. 
This idea can be generalized to design new representations, that are invariant to other types of nuisance image changes. 

\section*{Acknowledgment}
We thank Eyal Gofer for his help in improving the clarity
and phrasing of the paper. We thank the research group
members, Shachar, Ilya, Reut, Tom and Jonathan for their
feedback on the paper. We acknowledge support by the Israel Science Foundation (Grant No. 534/19), the Ministry of Science and Technology (Grant 3-15621) and by the Ollendorff Minerva Center.

\newpage\clearpage
\appendix{}
\section*{Supplementary Material}

In the supplementary material we provide details about the training of our model and explain in details about the training data. We also present additional evaluation examples of the proposed photo-consistent transform and we explain the ablation study performed during our research which resulted in the final configuration of our model.

\section{Training Details}
The hyperparameters values used in our final model are given below. In addition, we explain the implementation of the scale consistency loss.
\subsection{Hyperparameters}
\begin{itemize}
    \item \textbf{Training steps:}
    Batch size=16, Epoch size=1000, Number of epochs=75.
    \item \textbf{Adam optimizer:}
    Momentum=0.9, Beta=0.999, Learning rate=1e-5. 
\end{itemize}

\subsection{Scale Consistency Loss}
The scale consistency loss was defined in the paper, Eq. (11)
\begin{equation}
    L_{SC}(f_a)=D_{scale}(F(G(f_a,\rho)),G(F(f_a),\rho)),
\end{equation}
where $G$ is "Up-sample and Crop" and represents a bilinear up-sampling by a random factor $\rho \in \left(1,2\right] $ followed by a crop to the original patch size.
It is calculated using an additional instance of PhIT-Net (apart from the A, P, N instances).
The loss is computed only with respect to the anchor patches. The reason is that it is an "internal" loss, relating the patch to itself. It is not based on patch comparison, thus there is no need to duplicate it for the whole triplet allowing a faster training process.  We remind that in the triplet network model the training is based on various instances of the same network with shared weights. Hence, a loss computed for one instance affects all instances.

\section{Training and Test Data}
\label{sec:train_test_data}

\subsection{Outdoors Dataset}
We train and test our main model using outdoor images from the BigTime dataset (See Figure \ref{fig:BTsample}). 
Ideally, for training we would like the scenes to be completely static with changes only in illumination conditions. However, this is not always the case. Since the images are taken from time-lapse videos, there are small camera movements over time. Thus, alignment is not perfect. In addition, there are sky changes over time. Finally, the scenes are actually only semi-static, there are changes which happen over time such as cars or people that appear or disappear, windows that are open or shut, etc. 
Both the changing skies and the object changes are not part of the illumination variations we assume for the learning process. These issues were already raised by \cite{li2018learning} who provide masks for regions which exhibit change not related to illumination. We take into consideration these masks in the patch selection at training. The camera movements are not corrected in the dataset. Thus we chose to manually select the most stable scenes (with minimal camera movement) for training.

The training is based on square patches of $64 \times 64$ pixels. This size is large enough to include most relevant semi-local illumination cues. We found that smaller patches do not contain enough details and the use of larger patches does not improve quality and considerably slows down the training.
The training set is composed of $240K$ triplets, extracted from $600$ image pairs of 10 outdoor scenes.
The evaluation was done using 100 image pairs selected from 17 additional outdoor scenes not used in training.


\subsection{Indoors Dataset}
In addition to our main outdoors dataset, we train and evaluate our model also on a set of indoor images. The dataset is comprised of 23 different scenes from Middlebury 2014 stereo dataset \cite{scharstein2014high}). Each scene contains two images acquired under different lighting conditions. We train our model with patches extracted from 16 scenes (in the same manner done for BigTime) and test it with the remaining 7 scenes.

\begin{figure}[h]
\centering
\includegraphics[width=0.95\linewidth]{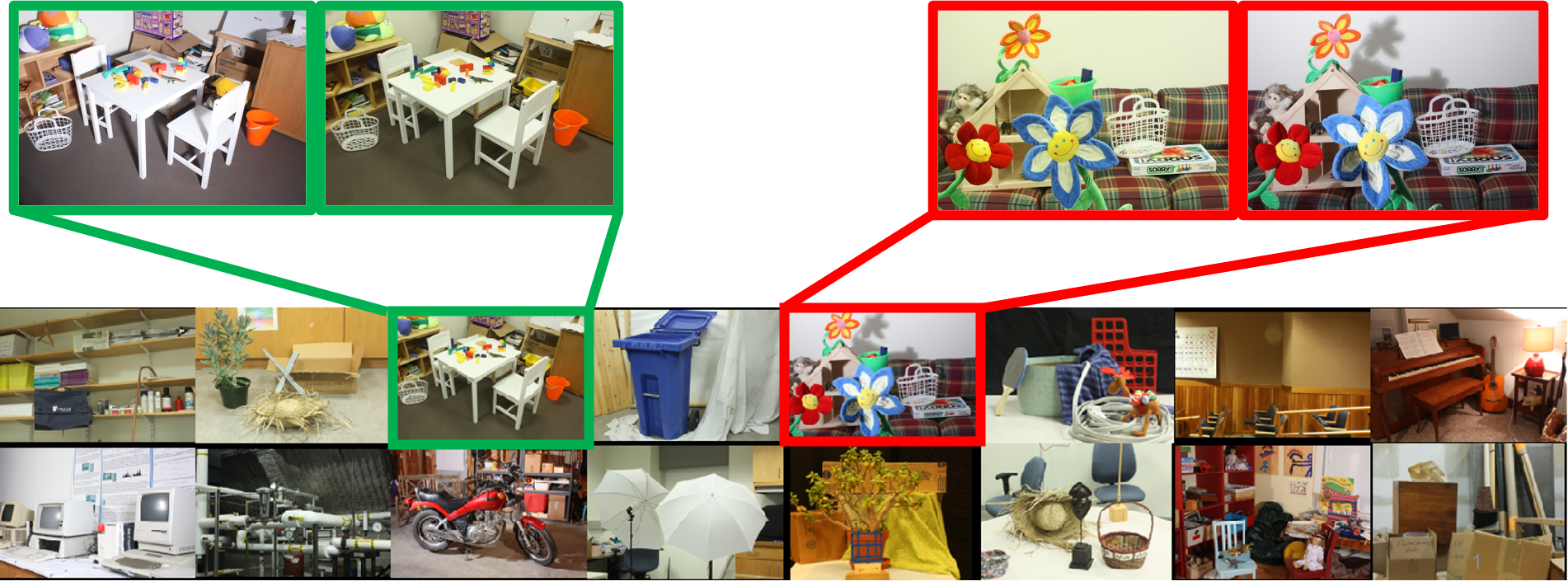}
\caption{We use BigTime \cite{li2018learning} as our outdoors dataset. For each scene there are several images under different lighting conditions. The dataset is composed of diverse scenes.
}
\label{fig:MBsample}
\end{figure}

\section{Additional Examples}
\subsection{Visual Photo-consistency}

\begin{figure}[h]
\begin{center}
\setlength{\tabcolsep}{0.1em} 
\begin{tabular}{ATTT}
\capstyle{Original image} &

{\includegraphics[height=1.35cm]{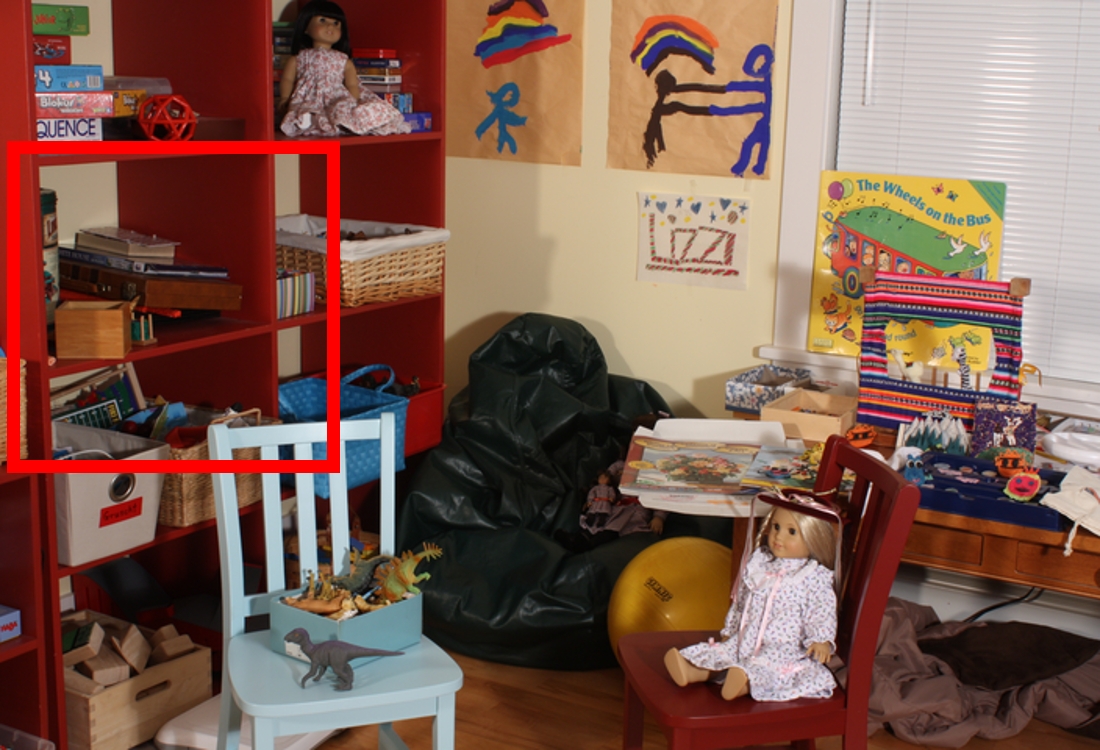}} &
{\includegraphics[height=1.35cm]{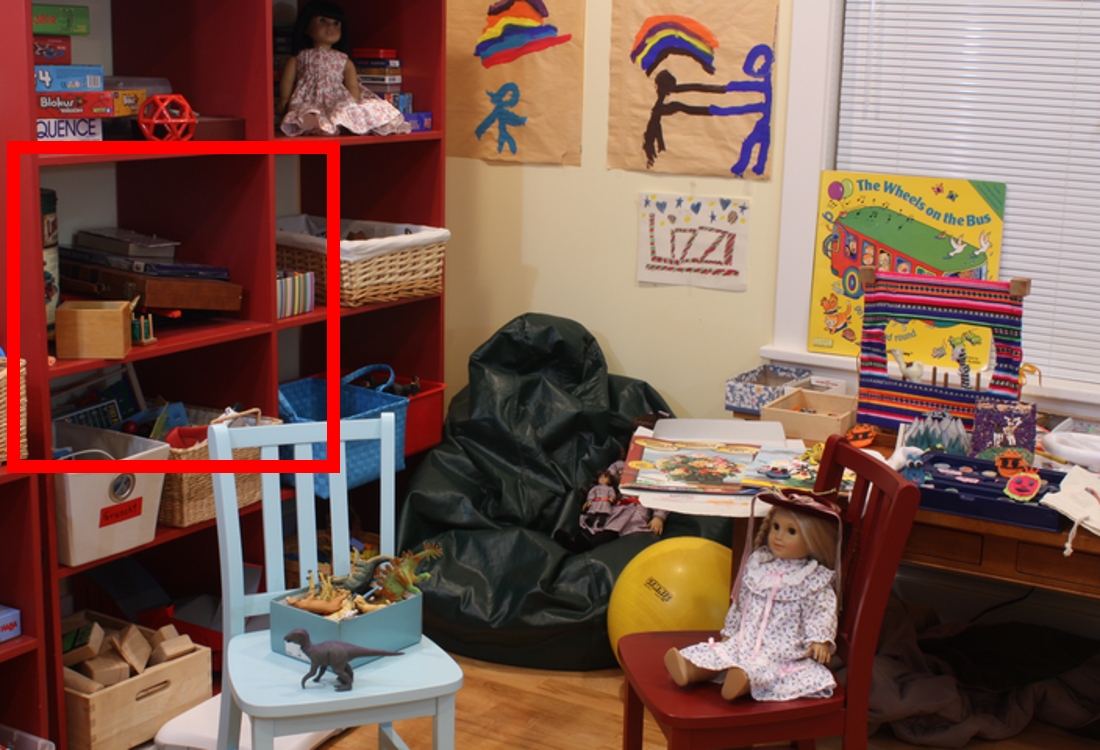}} &
{\includegraphics[height=1.35cm]{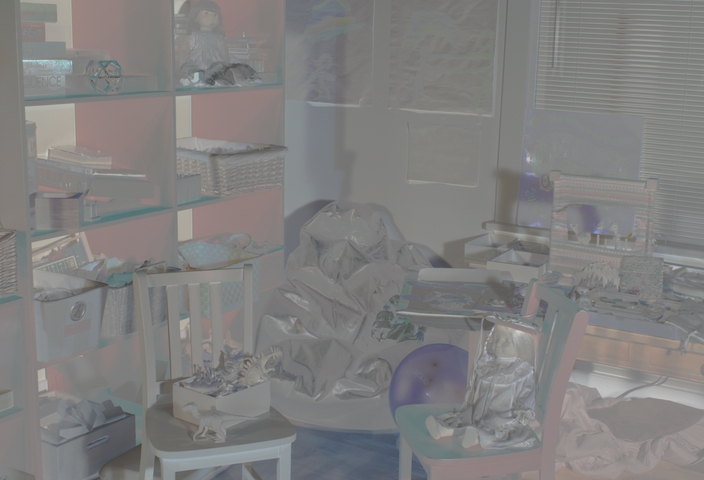}} 
\\
\capstyle{Ours} &

{\includegraphics[height=1.35cm]{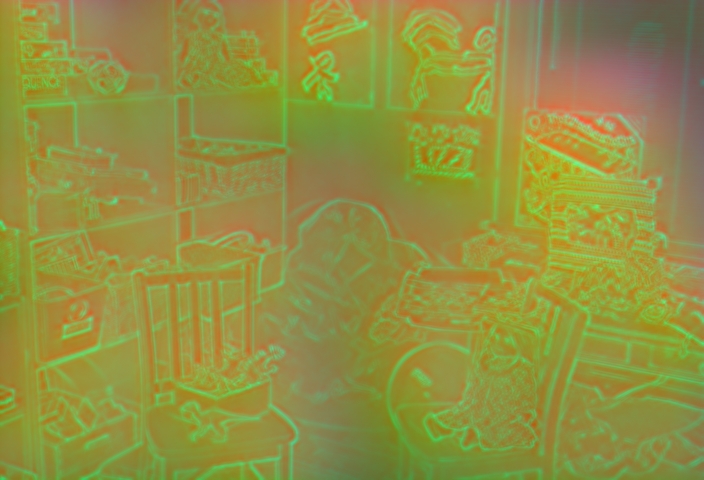}} &
{\includegraphics[height=1.35cm]{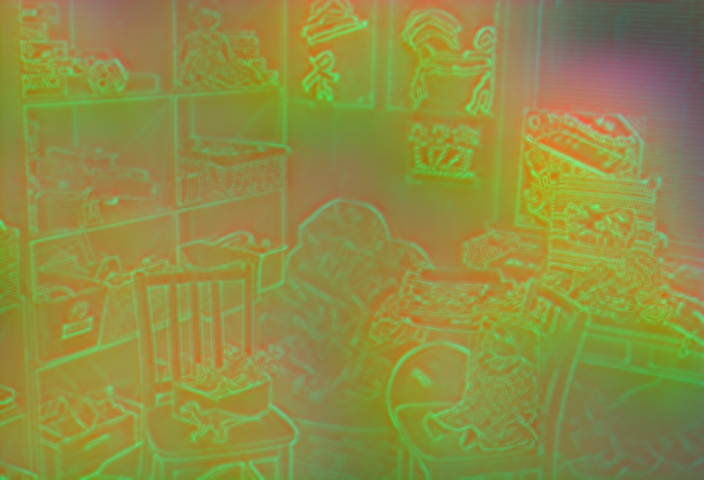}} &
{\includegraphics[height=1.35cm]{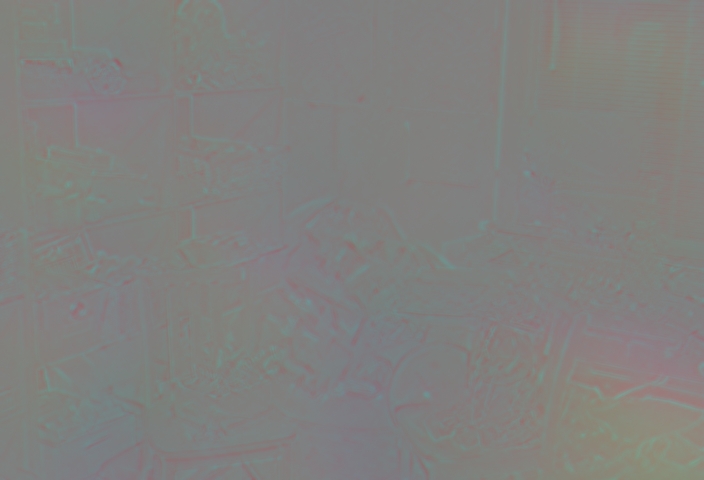}} 
\\
\capstyle{Lettry \cite{lettry2018unsupervised}} &

{\includegraphics[height=1.35cm]{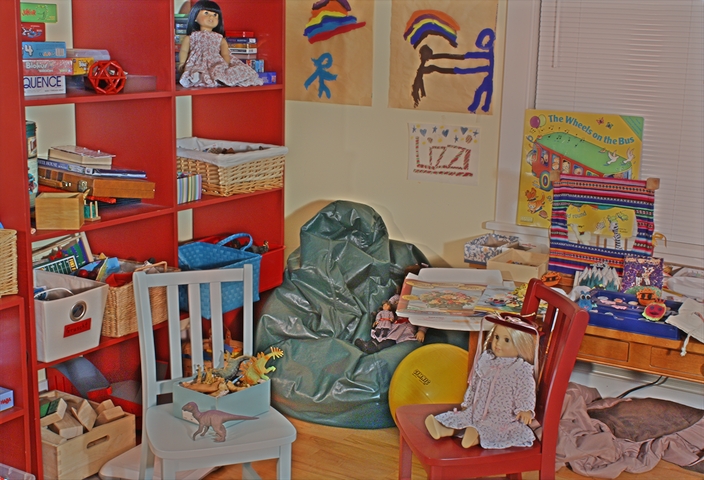}} &
{\includegraphics[height=1.35cm]{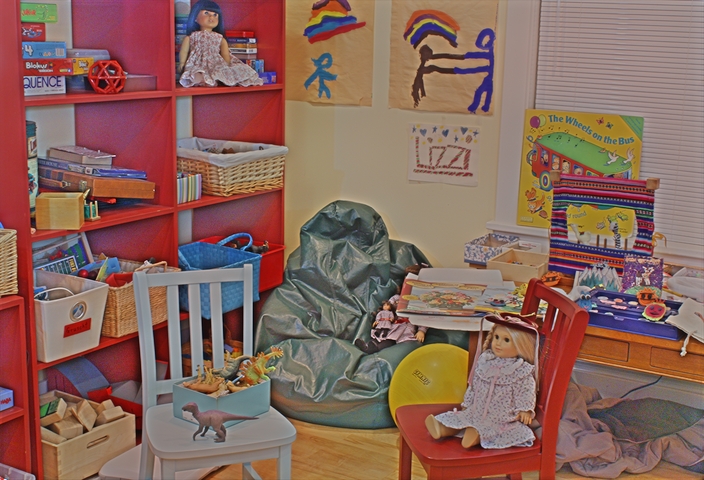}} &
{\includegraphics[height=1.35cm]{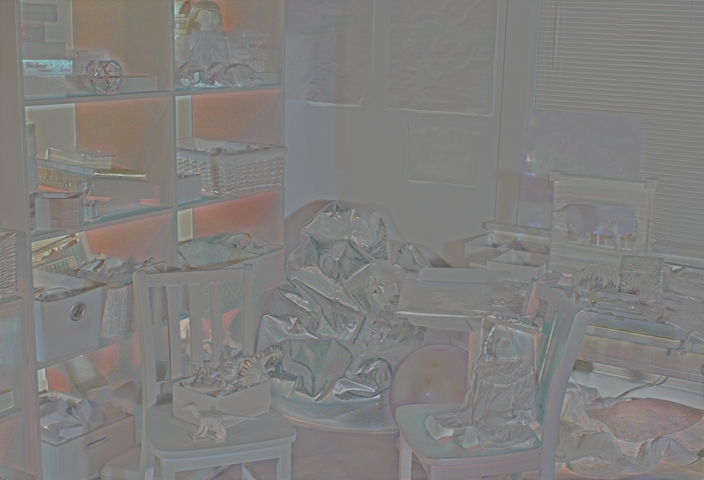}} 
\\
\capstyle{Li-Snavely \cite{li2018learning}} &

{\includegraphics[height=1.35cm]{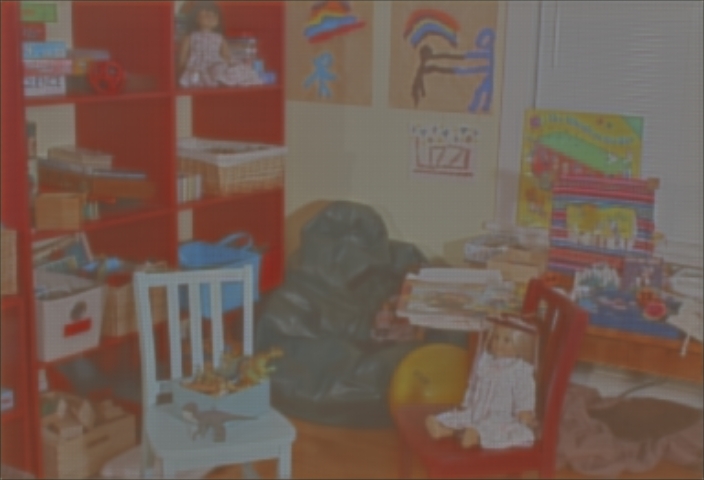}} &
{\includegraphics[height=1.35cm]{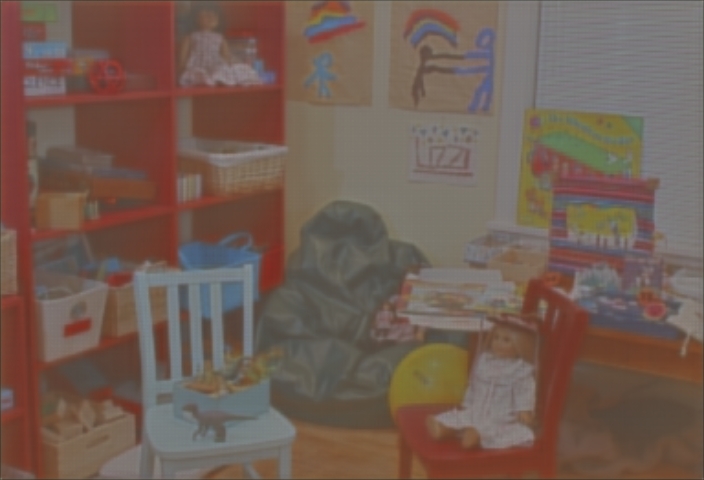}} &
{\includegraphics[height=1.35cm]{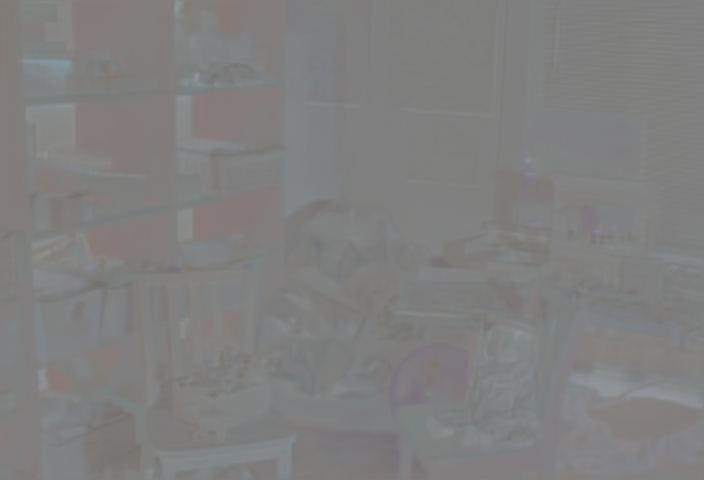}} 
\\
\capstyle{Maddern \cite{maddern2014illumination}} &

{\includegraphics[height=1.35cm]{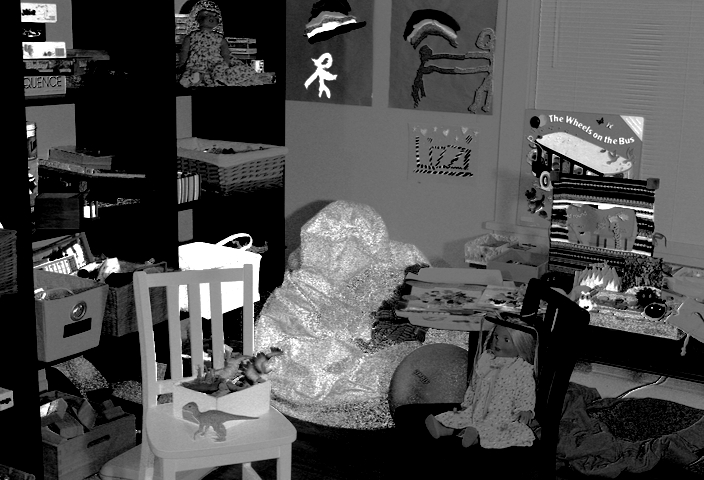}} &
{\includegraphics[height=1.35cm]{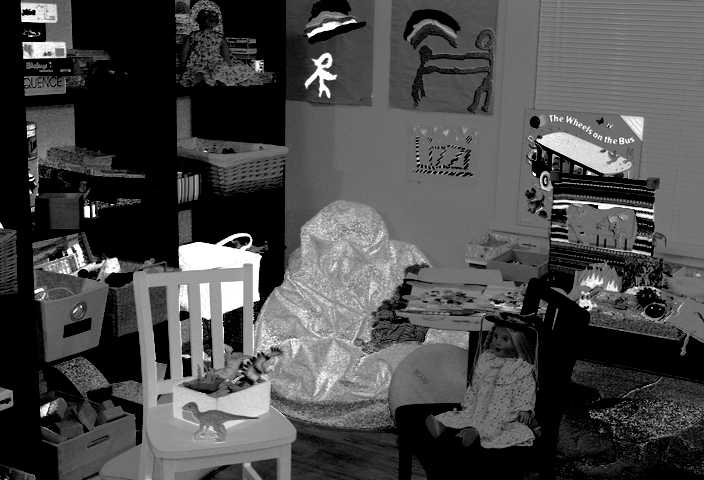}} &
{\includegraphics[height=1.35cm]{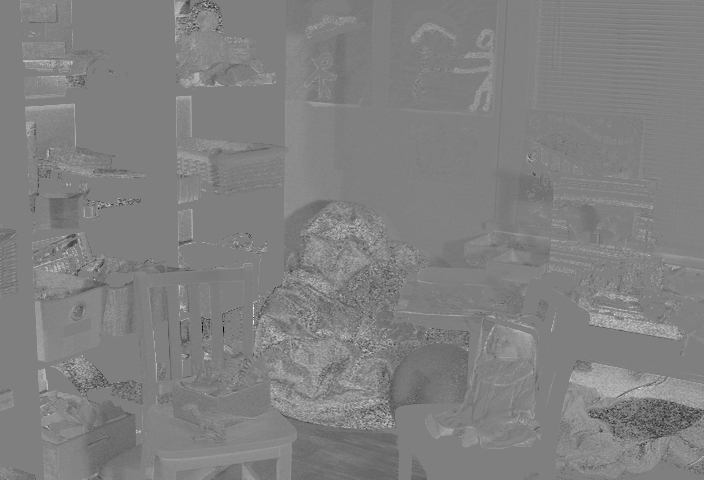}} 
\\
\vspace{0.5cm}
\\
\capstyle{Original image} &

{\includegraphics[height=1.35cm]{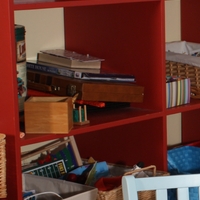}} &
{\includegraphics[height=1.35cm]{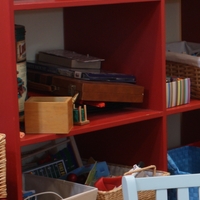}} &
{\includegraphics[height=1.35cm]{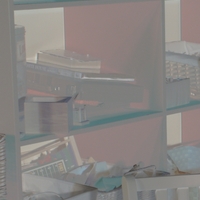}} 
\\
\capstyle{Ours} &

{\includegraphics[height=1.35cm]{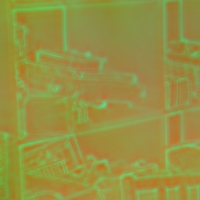}} &
{\includegraphics[height=1.35cm]{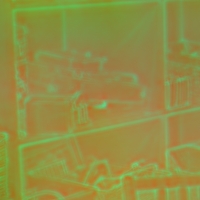}} &
{\includegraphics[height=1.35cm]{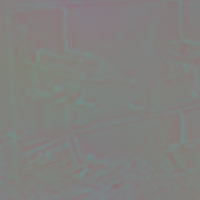}} 
\\
\capstyle{Lettry \cite{lettry2018unsupervised}} &

{\includegraphics[height=1.35cm]{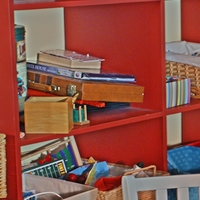}} &
{\includegraphics[height=1.35cm]{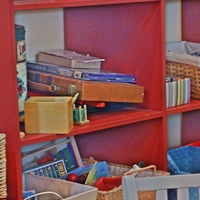}} &
{\includegraphics[height=1.35cm]{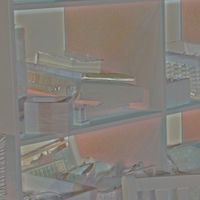}} 
\\ 
\capstyle{Li-Snavely \cite{li2018learning}} &

{\includegraphics[height=1.35cm]{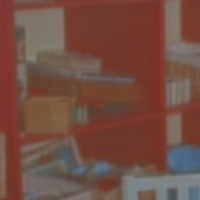}} &
{\includegraphics[height=1.35cm]{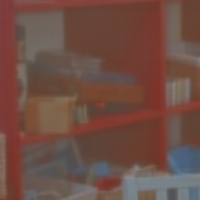}} &
{\includegraphics[height=1.35cm]{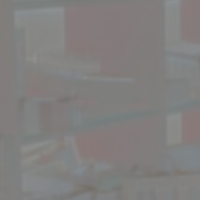}} 
\\
\capstyle{Maddern \cite{maddern2014illumination}} &

{\includegraphics[height=1.35cm]{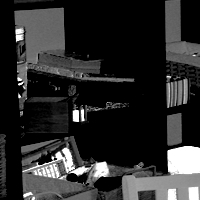}} &
{\includegraphics[height=1.35cm]{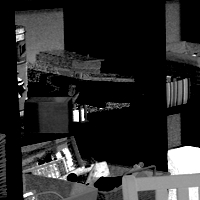}} &
{\includegraphics[height=1.35cm]{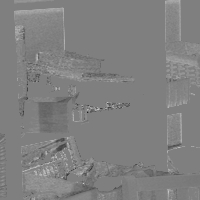}} 
\\
&
\capstyle{Image 1} &
\capstyle{Image 2} &
\capstyle{Difference}
\end{tabular}

\caption{Visual comparison of invariant representation methods. The
representation of two images under different illumination conditions is shown. The difference (ideally zero) affirms that our representation is highly stable under illumination
changes (zero is gray). 
In the top of the figure we see the comparison with the full images and in the bottom we show enlarged crops from the full images. The original crops location is marked with a red frame in the original images (first row).}
\label{fig:visual_diff_MB2}
\end{center}
\end{figure}

\bigskip
\noindent\textbf{Indoors Dataset (Middlebury).}
As explained before, in addition to training a model with the BigTime dataset, we also trained and evaluated our model on an indoors dataset. 
In Fig. \ref{fig:visual_diff_MB2} we present additional examples of scenes from this dataset \cite{scharstein2014high} with our transform, compared to other methods. We also show enlarged crops of interesting regions in the images.
We can see in these examples that our representation, have the lowest differences compared to the others.  
We note that the Maddern representation of the bicycle scene (left) also exhibits a small difference between different illuminations. However, it removes significant structural information (and hence did not perform well in the quantitative experiments, as shown in the paper).

\subsection{Patch Matching}
\begin{figure}[h]
\begin{center}
\setlength{\tabcolsep}{0.1em} 
\begin{tabular}{cccccc}
\centering
{\includegraphics[width=0.15\linewidth]{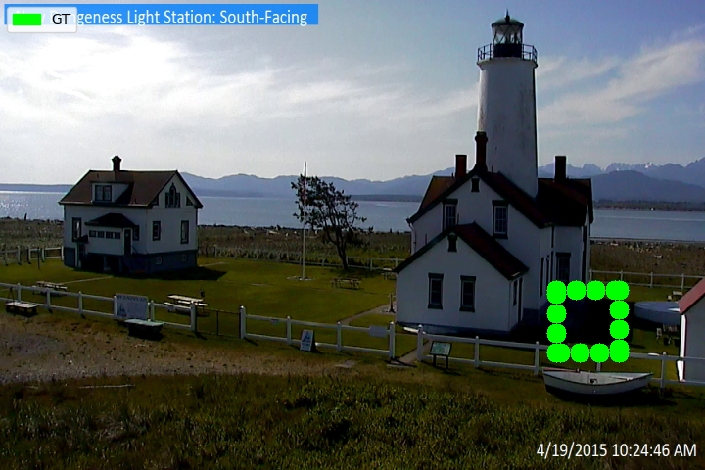}} &
{\includegraphics[width=0.15\linewidth]{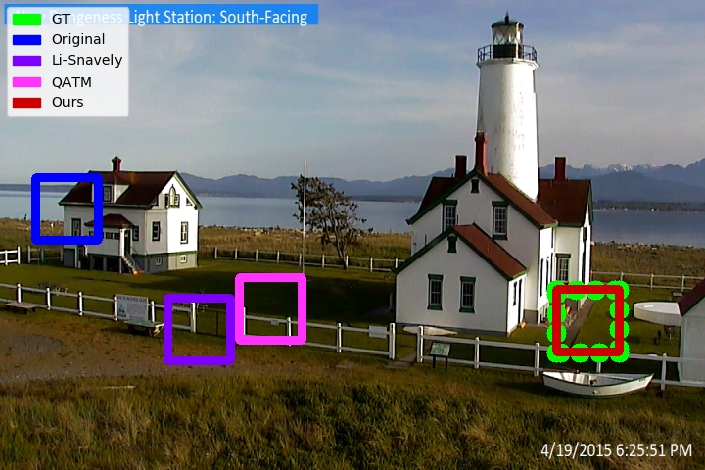}} &
{\includegraphics[width=0.15\linewidth]{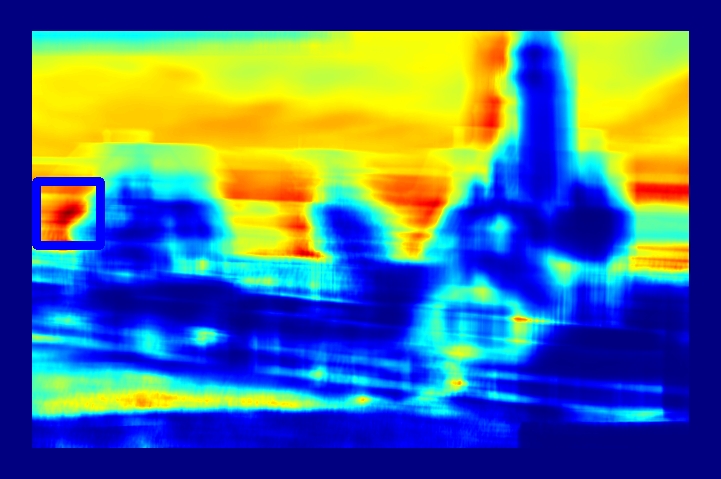}} &
{\includegraphics[width=0.15\linewidth]{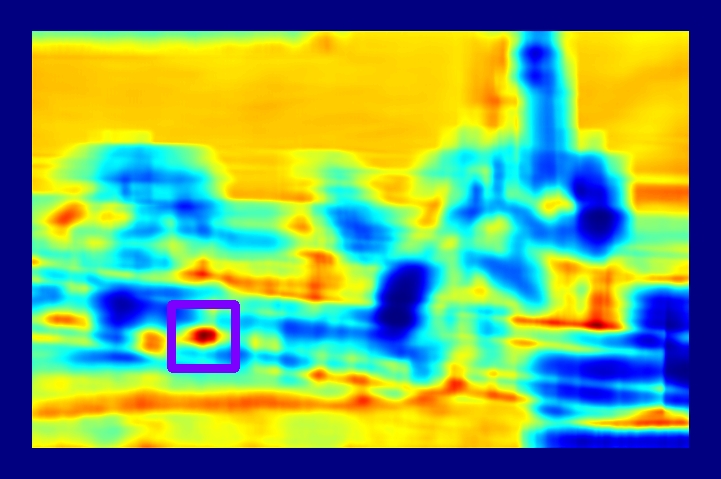}} &
{\includegraphics[width=0.15\linewidth]{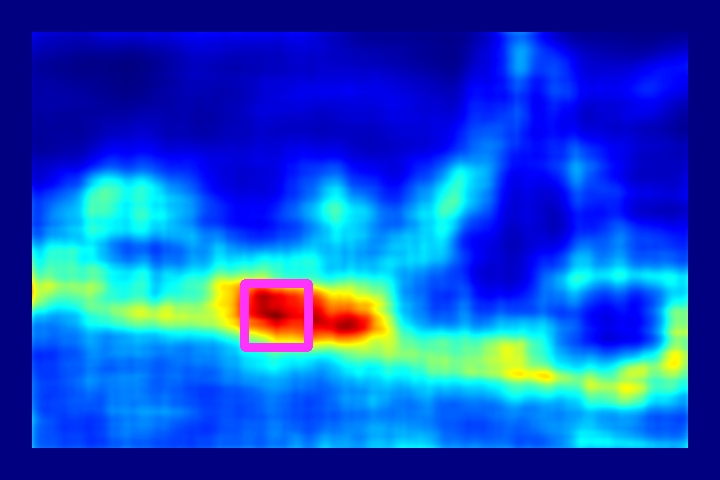}}&
{\includegraphics[width=0.15\linewidth]{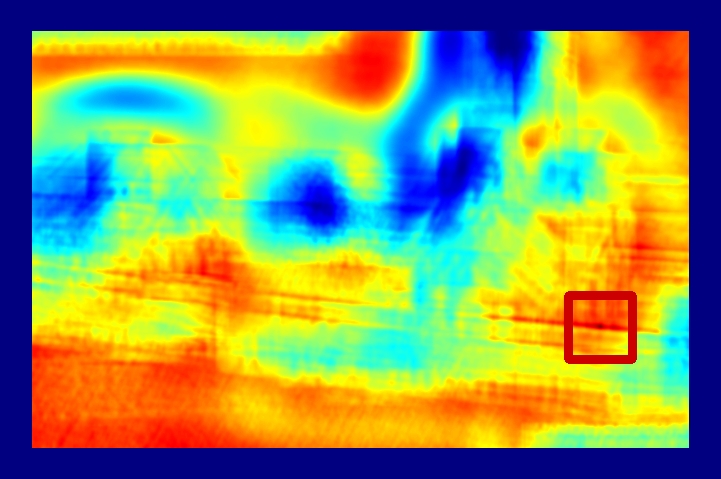}} 
\\
{\includegraphics[width=0.15\linewidth]{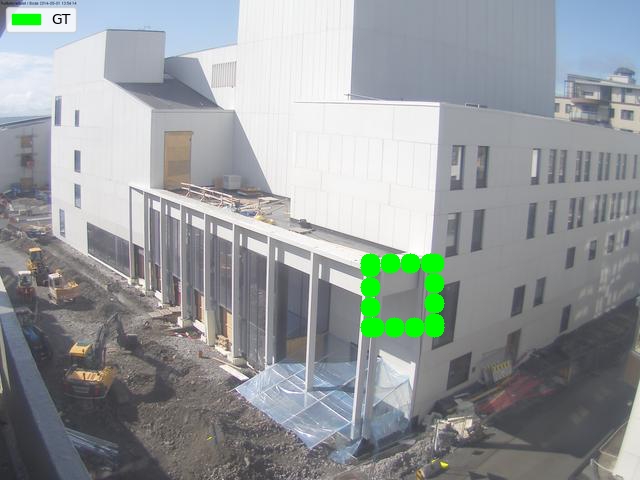}} &
{\includegraphics[width=0.15\linewidth]{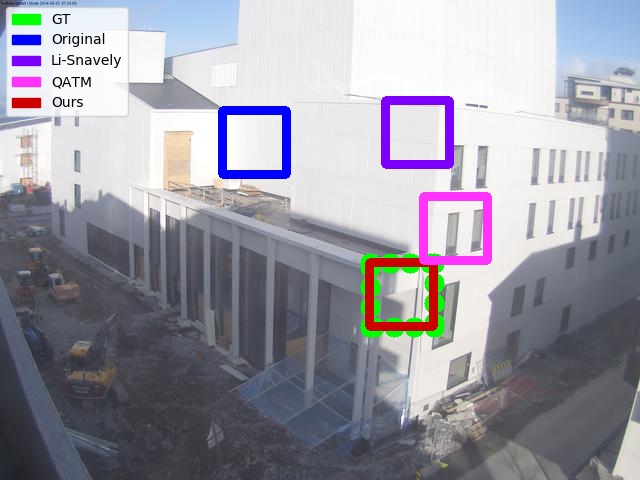}} &
{\includegraphics[width=0.15\linewidth]{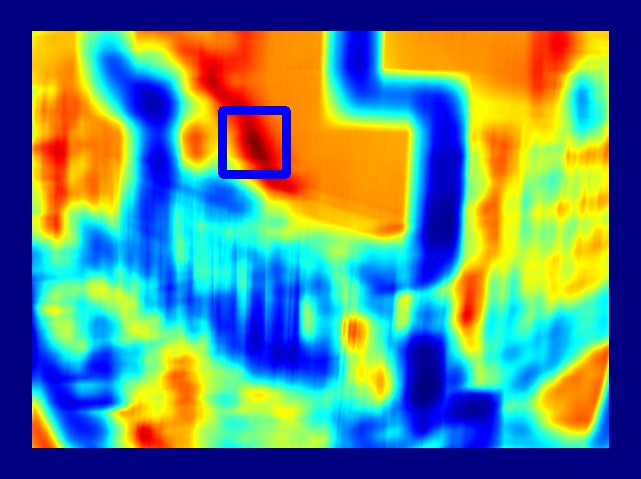}} &
{\includegraphics[width=0.15\linewidth]{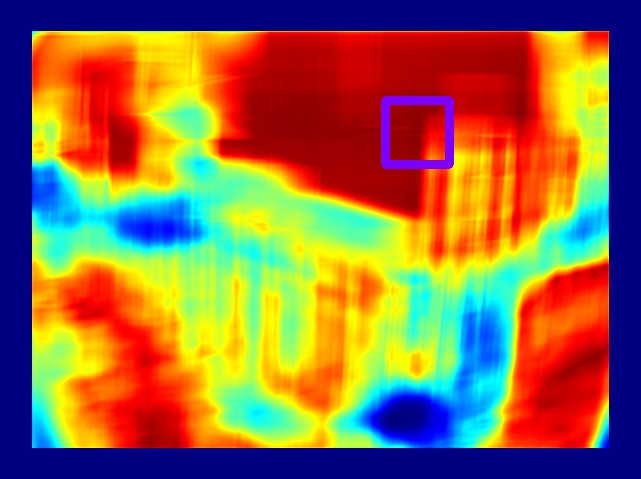}} &
{\includegraphics[width=0.15\linewidth]{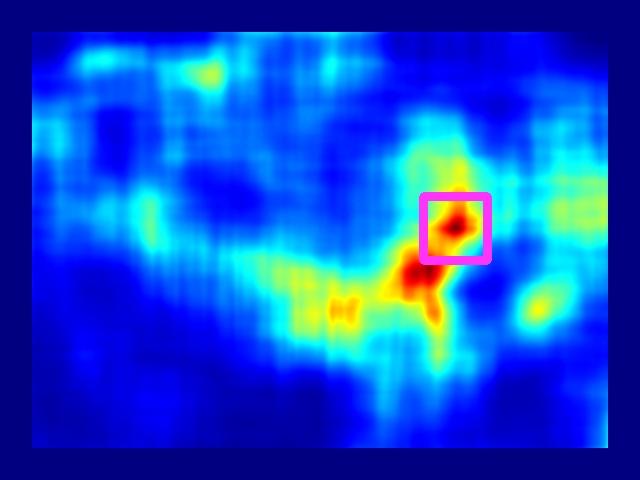}}&
{\includegraphics[width=0.15\linewidth]{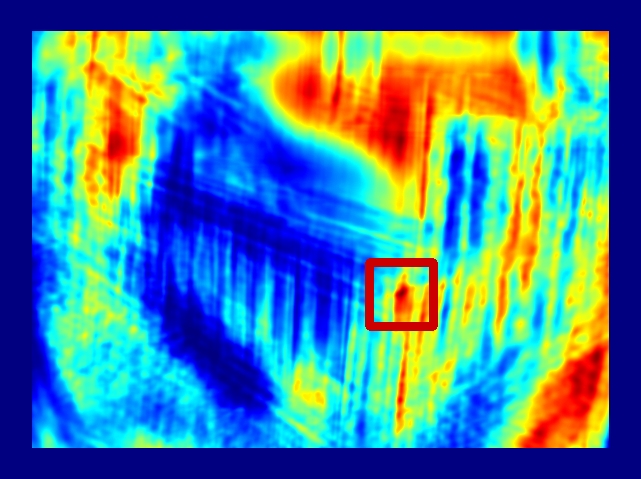}} 
\\
{\includegraphics[width=0.15\linewidth]{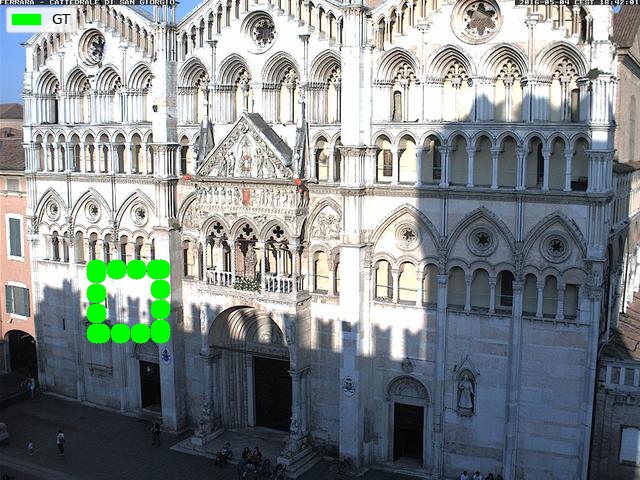}} &
{\includegraphics[width=0.15\linewidth]{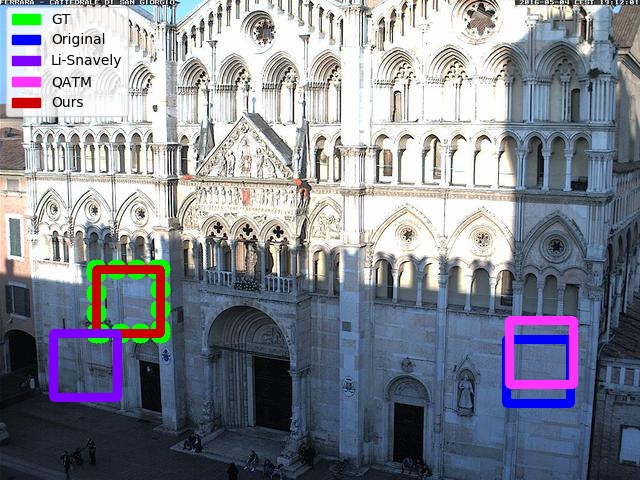}} &
{\includegraphics[width=0.15\linewidth]{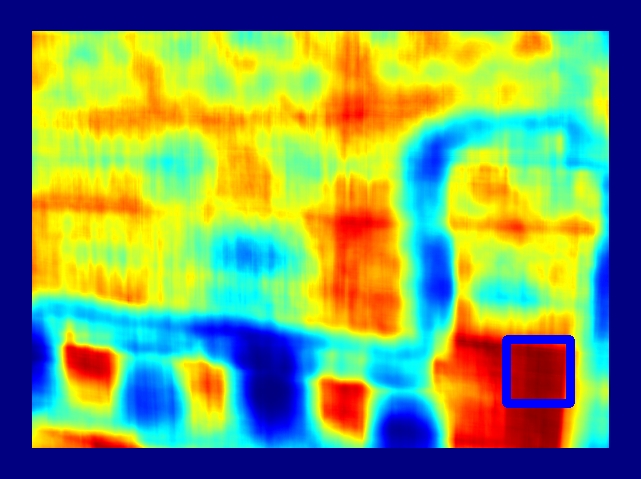}} &
{\includegraphics[width=0.15\linewidth]{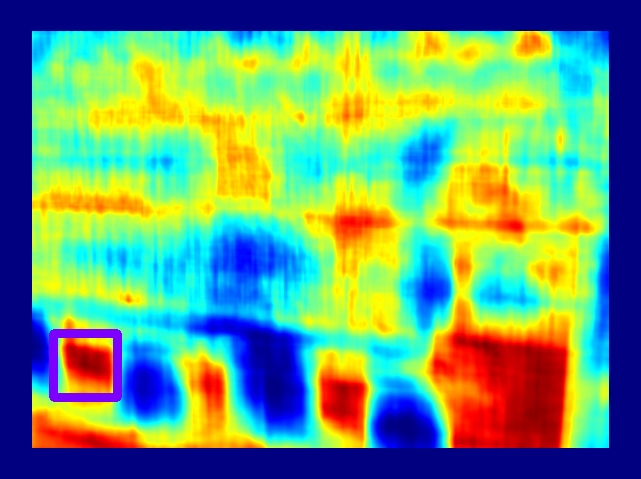}} &
{\includegraphics[width=0.15\linewidth]{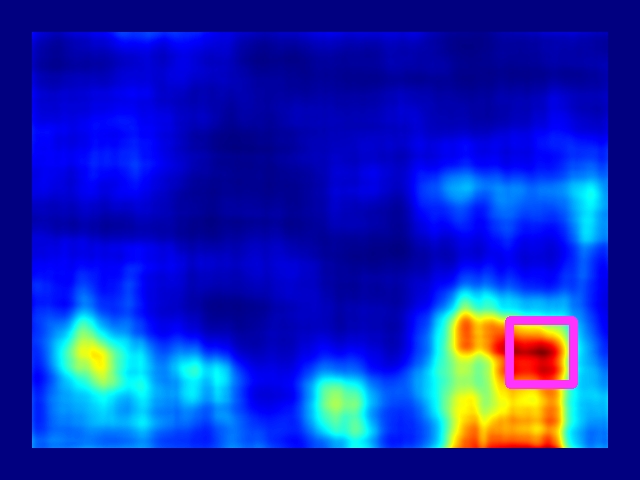}}&
{\includegraphics[width=0.15\linewidth]{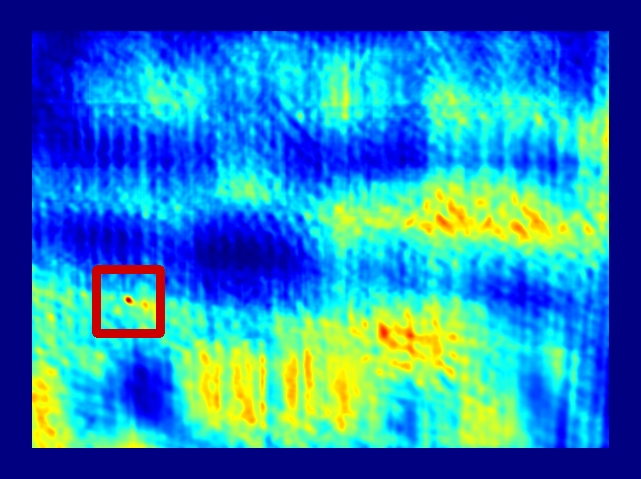}} 
\\
{\includegraphics[width=0.15\linewidth]{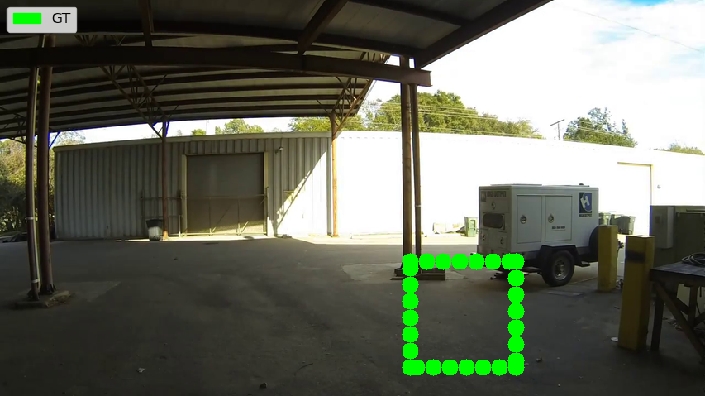}} &
{\includegraphics[width=0.15\linewidth]{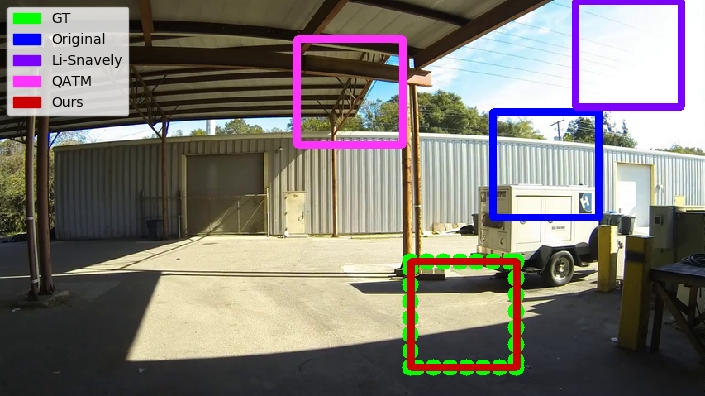}} &
{\includegraphics[width=0.15\linewidth]{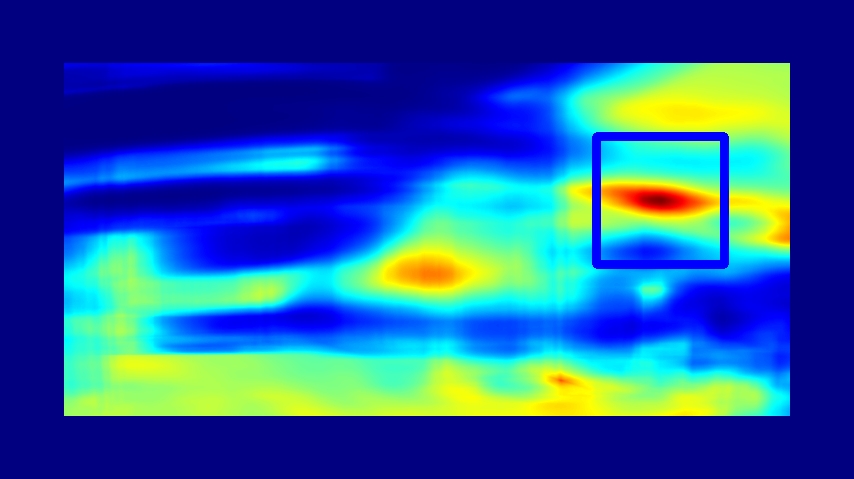}} &
{\includegraphics[width=0.15\linewidth]{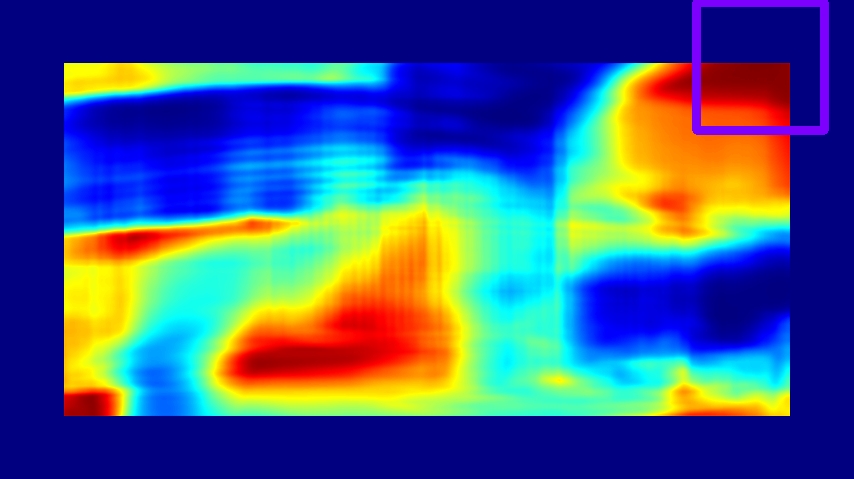}} &
{\includegraphics[width=0.15\linewidth]{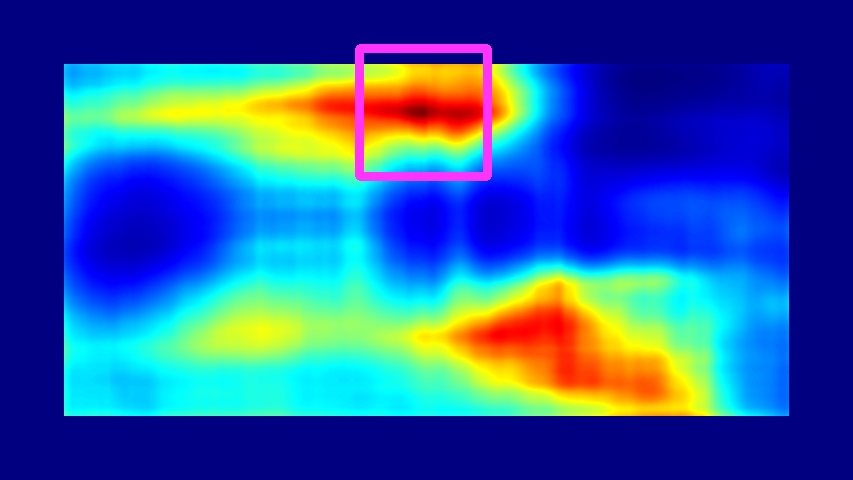}}&
{\includegraphics[width=0.15\linewidth]{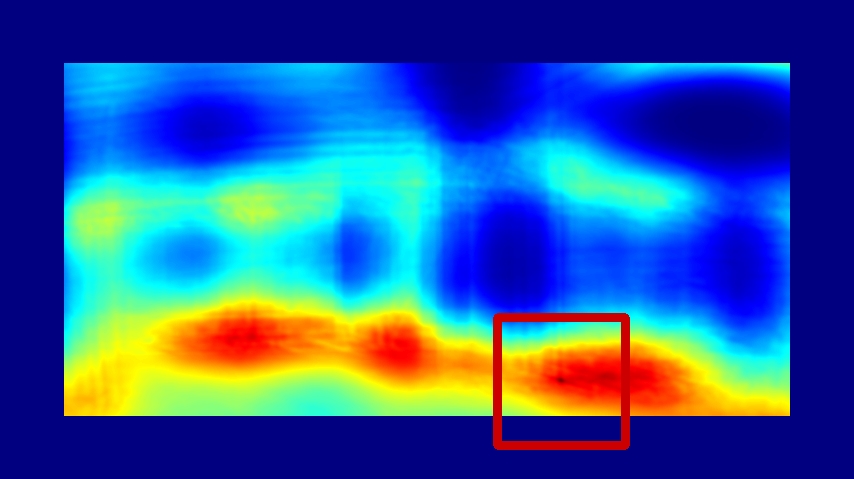}} 
\\
\hline
\\
{\includegraphics[width=0.15\linewidth]{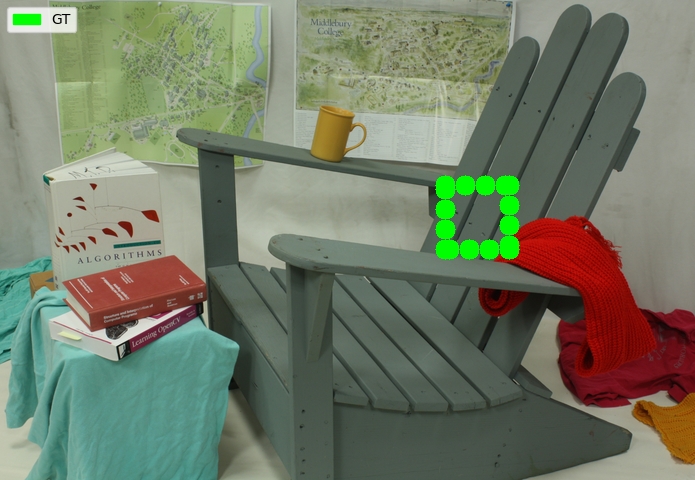}} &
{\includegraphics[width=0.15\linewidth]{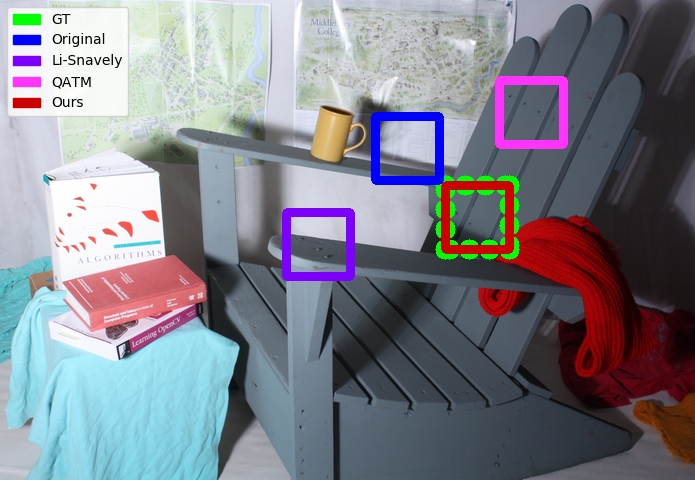}} &
{\includegraphics[width=0.15\linewidth]{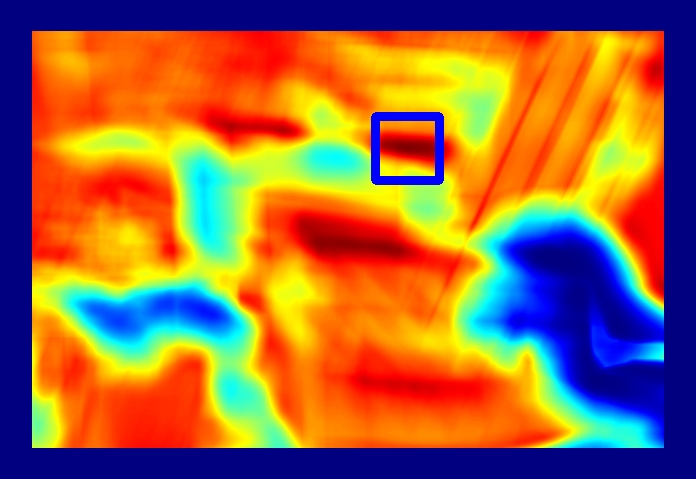}} &
{\includegraphics[width=0.15\linewidth]{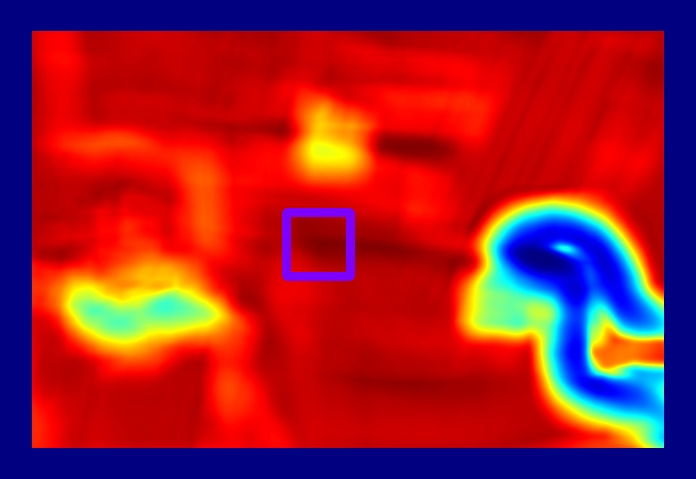}} &
{\includegraphics[width=0.15\linewidth]{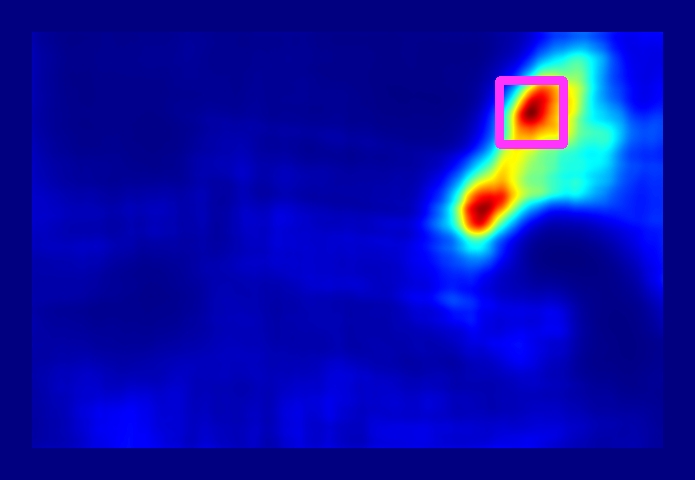}}&
{\includegraphics[width=0.15\linewidth]{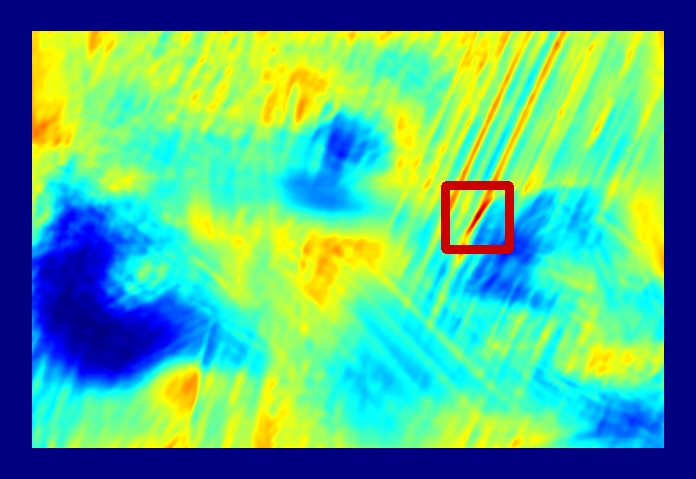}} 
\\
{\includegraphics[width=0.15\linewidth]{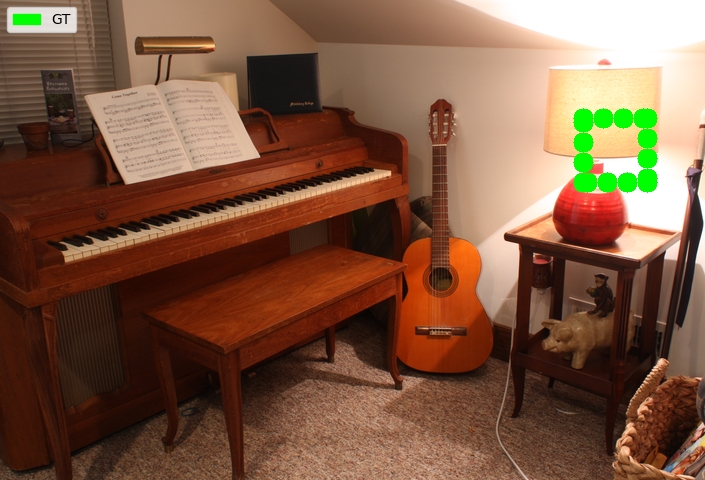}} &
{\includegraphics[width=0.15\linewidth]{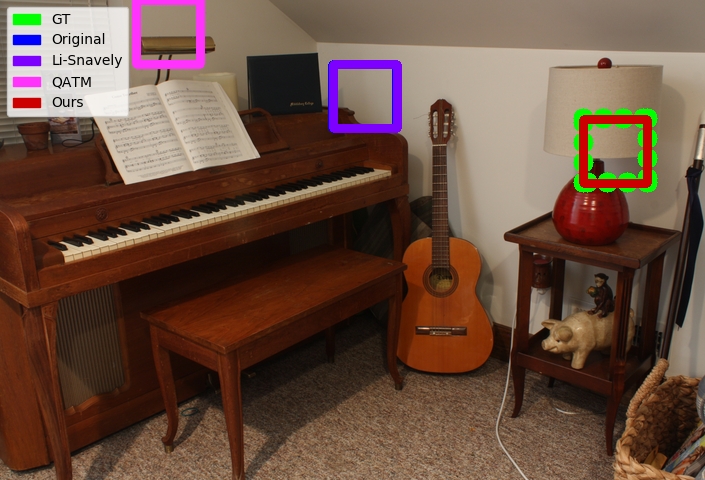}} &
{\includegraphics[width=0.15\linewidth]{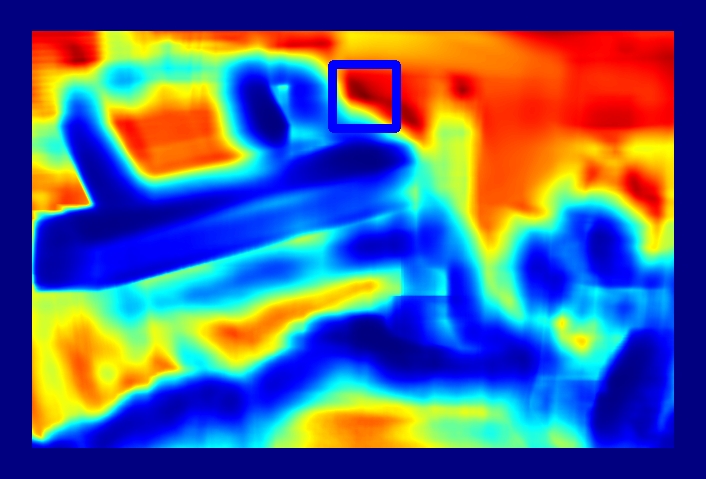}} &
{\includegraphics[width=0.15\linewidth]{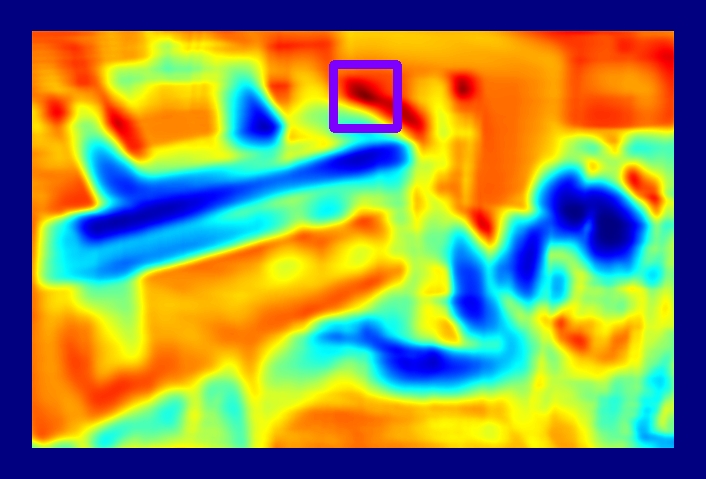}} &
{\includegraphics[width=0.15\linewidth]{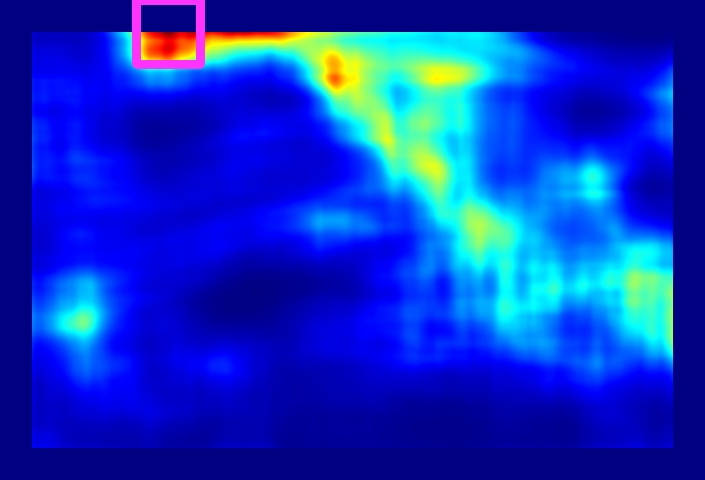}}&
{\includegraphics[width=0.15\linewidth]{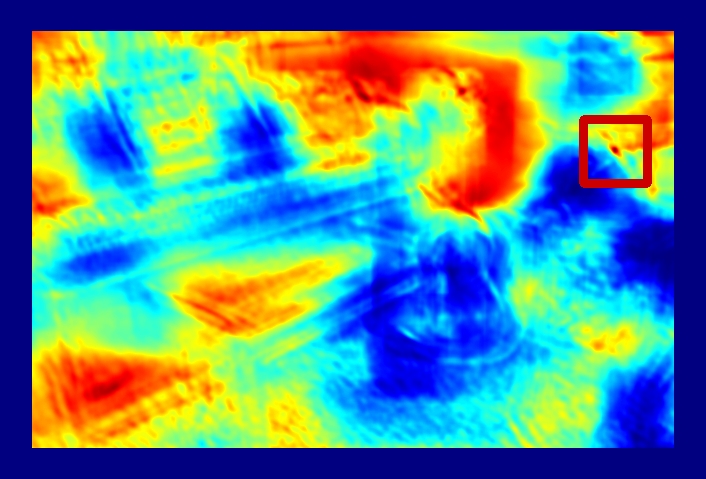}} 
\\
{\includegraphics[width=0.15\linewidth]{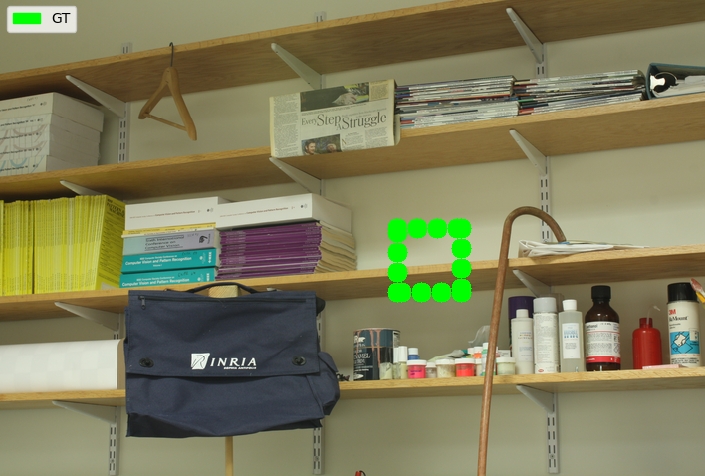}} &
{\includegraphics[width=0.15\linewidth]{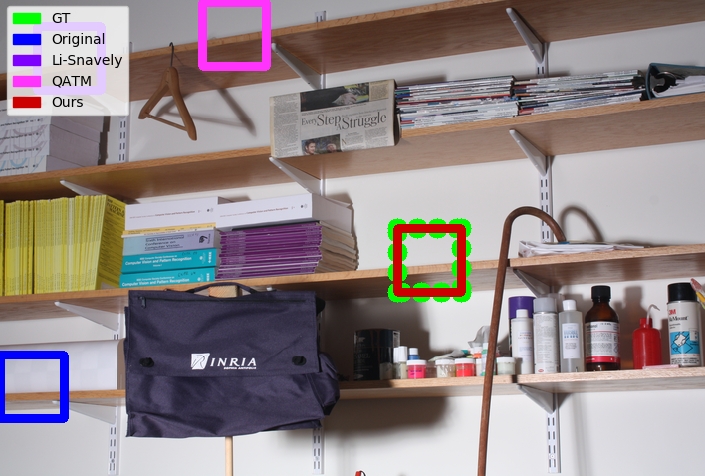}} &
{\includegraphics[width=0.15\linewidth]{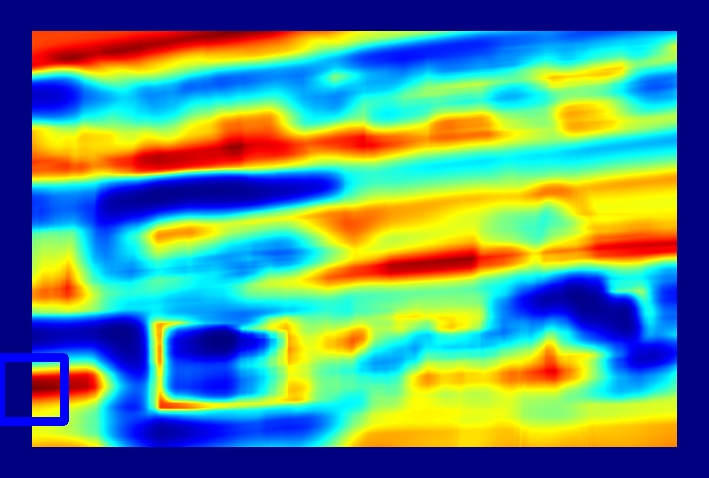}} &
{\includegraphics[width=0.15\linewidth]{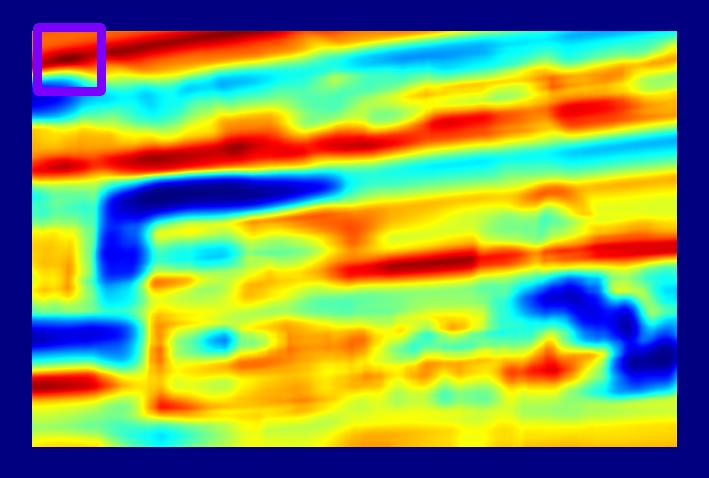}} &
{\includegraphics[width=0.15\linewidth]{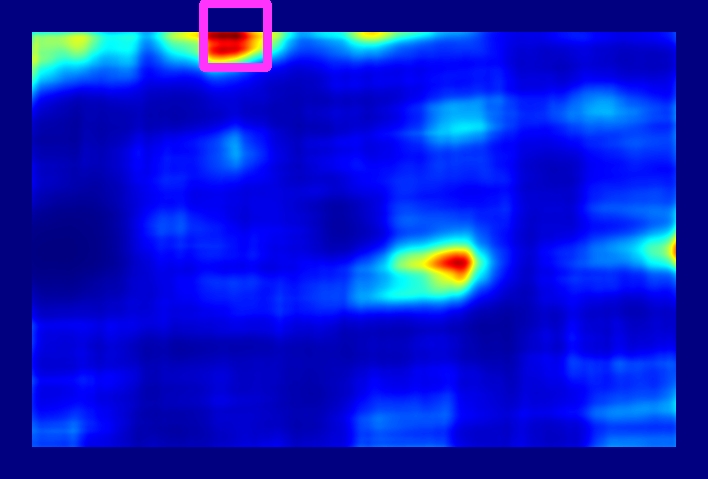}}&
{\includegraphics[width=0.15\linewidth]{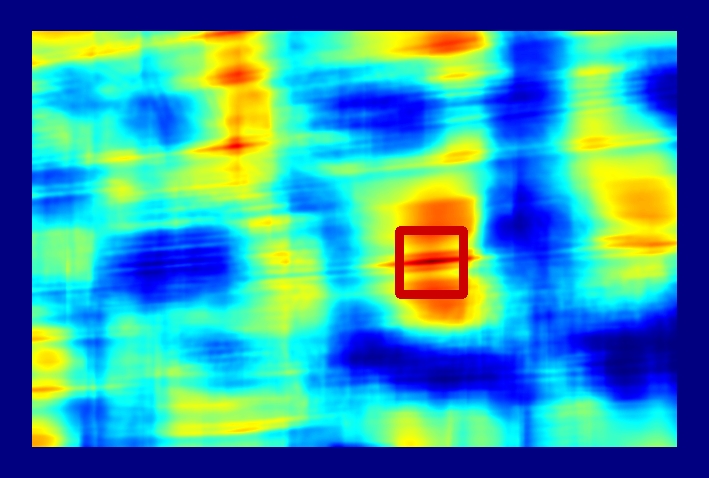}} 
\\
{\includegraphics[width=0.15\linewidth]{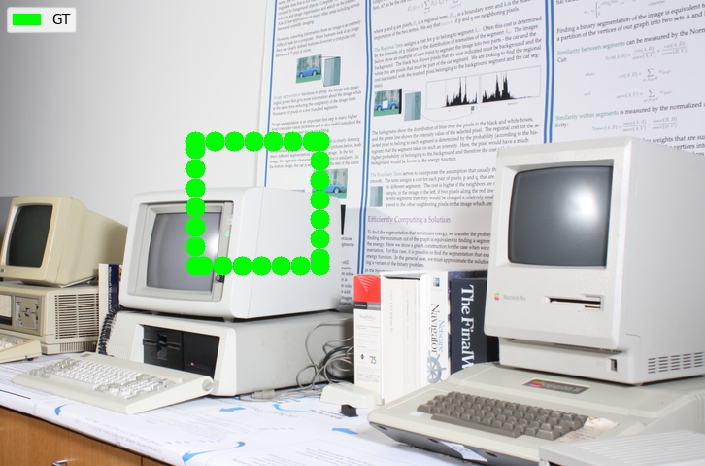}} &
{\includegraphics[width=0.15\linewidth]{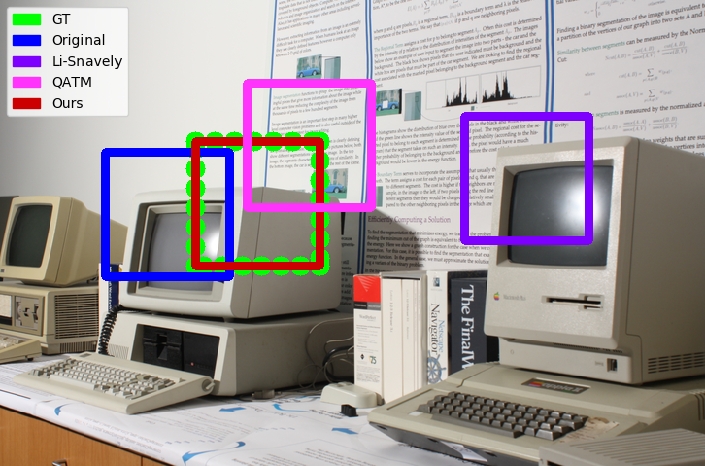}} &
{\includegraphics[width=0.15\linewidth]{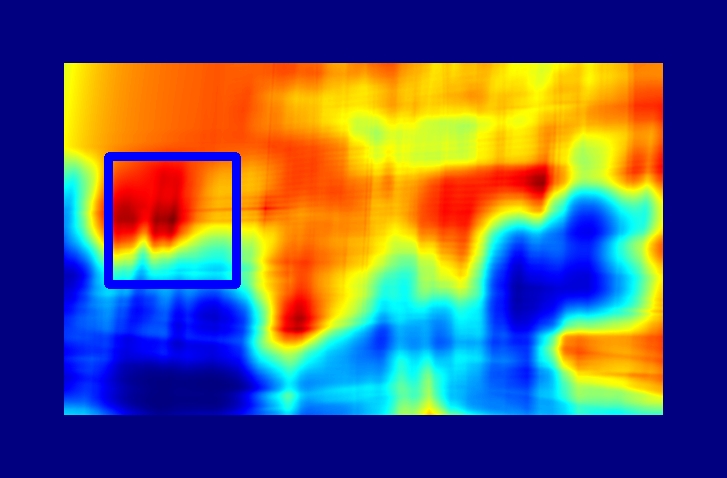}} &
{\includegraphics[width=0.15\linewidth]{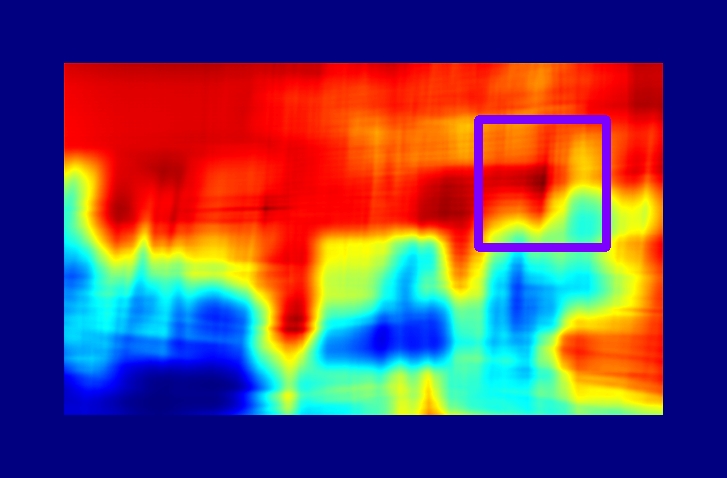}} &
{\includegraphics[width=0.15\linewidth]{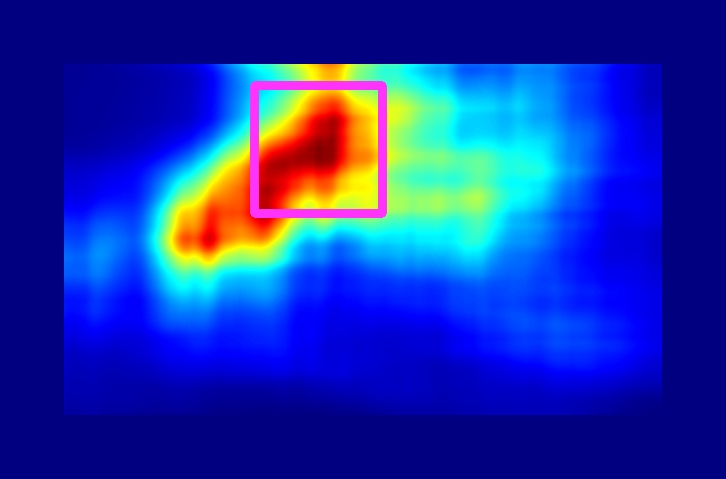}}&
{\includegraphics[width=0.15\linewidth]{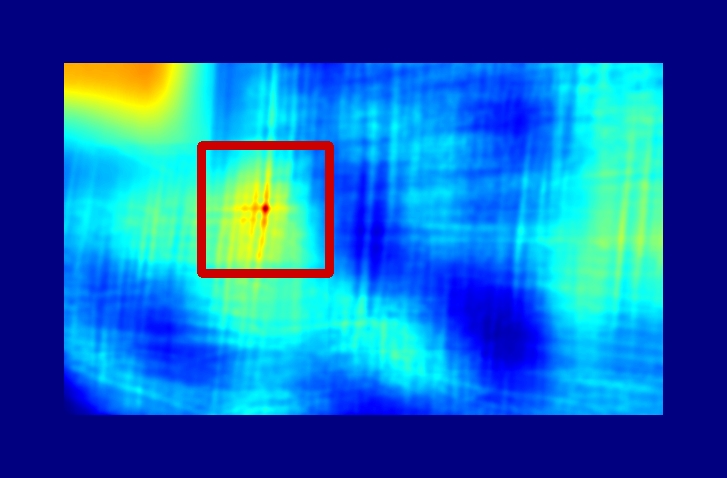}} 
\\
\capstyle{Reference Image} &
\capstyle{Target Image} &
\capstyle{Original} &
\capstyle{Li-Snavely\cite{li2018learning}} &
\capstyle{QATM\cite{cheng2019qatm}}&
\capstyle{Ours}
\\
\end{tabular}
\caption{Examples of patch matching results. The reference and target images belong to the same scene with different illumination. The goal is to correctly locate in the target image a patch which is selected in the reference image (green frame). The frames marking the results of the different algorithms are overlaid on the target image. Heatmaps of each algorithm (right) indicate high (red) to low (blue) matching scores. (Scenes above the line -- BigTime, scenes below the line -- Middlebury.)}
\label{fig:pm_heatmaps2}
\end{center}
\end{figure}

Following Section 5.2 and Fig. 8 in the paper, we present in Fig. \ref{fig:pm_heatmaps2} additional matching results of difficult scenarios on both datasets. We show the matching results and the correlation-based heatmaps, according to which the algorithm selects its match.

\clearpage
\newpage
\section{Ablation Study}
In this section we show different configurations of PhIT-Net. We examine two types of variations: Loss function variations and representations with different number of channels. The effects of these changes are quantified and visualized (Figure \ref{fig:ablation2}). The study is performed on the BigTime dataset by training a new network for each variation and evaluating its performance on the patch matching task, described in Section 5 in the paper.

\subsection{Loss Function Variations}

\noindent\textbf{Scale Consistency Loss.} In this study we set to zero the weight of the scale consistency loss,

\begin{equation}
\label{eq:scale_const2}
    L_{SC}(f_a)=D_{scale}(F(G(f_a,\rho)),G(F(f_a),\rho)),
\end{equation}

\begin{figure}[h]
\begin{center}
\setlength{\tabcolsep}{0.1em} 
\begin{center}
\begin{tabular}{ccccc}
{\includegraphics[width=0.19\linewidth]{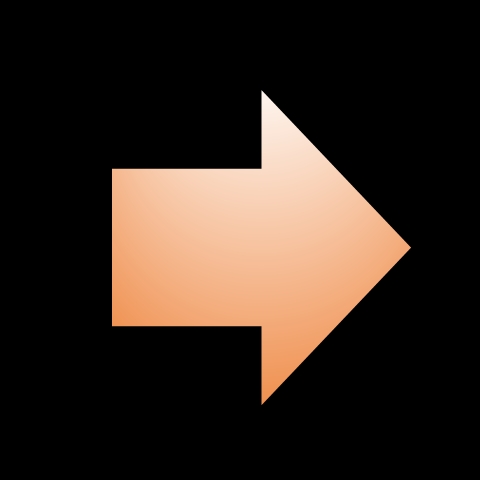}} &
{\includegraphics[width=0.19\linewidth]{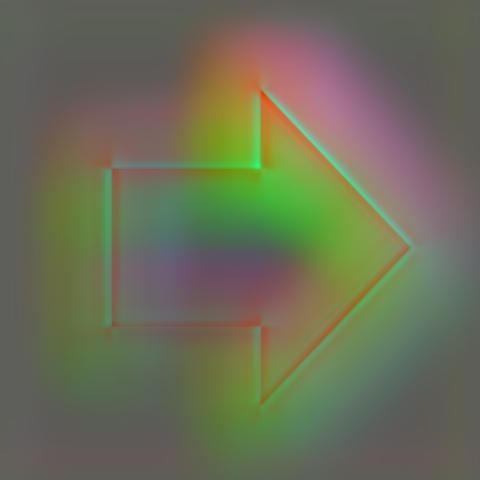}} &
{\includegraphics[width=0.19\linewidth]{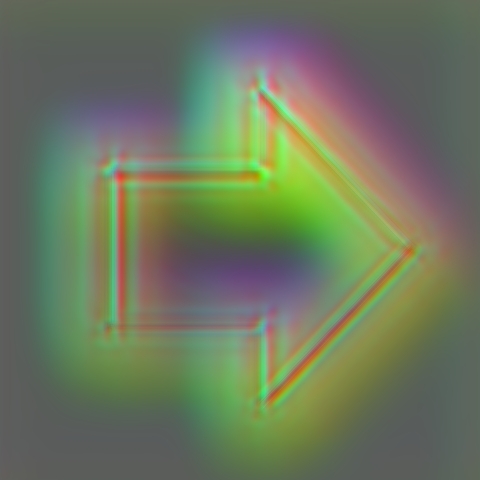}} &
{\includegraphics[width=0.19\linewidth]{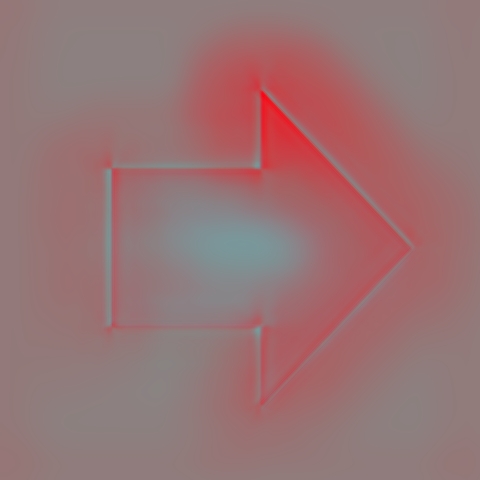}} &
{\includegraphics[width=0.19\linewidth]{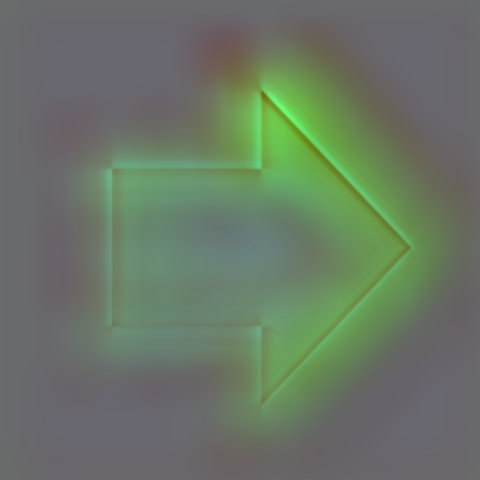}} 
\\
{\includegraphics[width=0.19\linewidth]{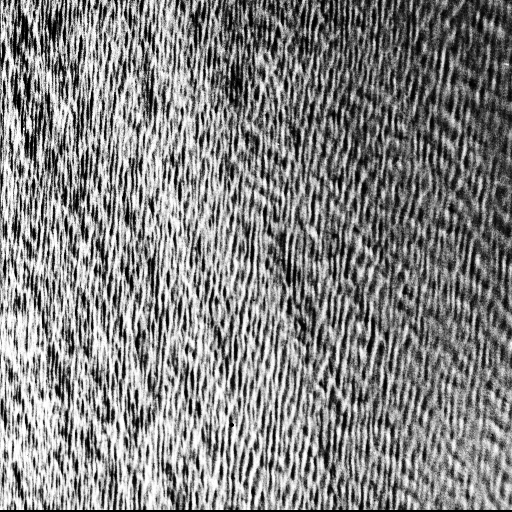}} &
{\includegraphics[width=0.19\linewidth]{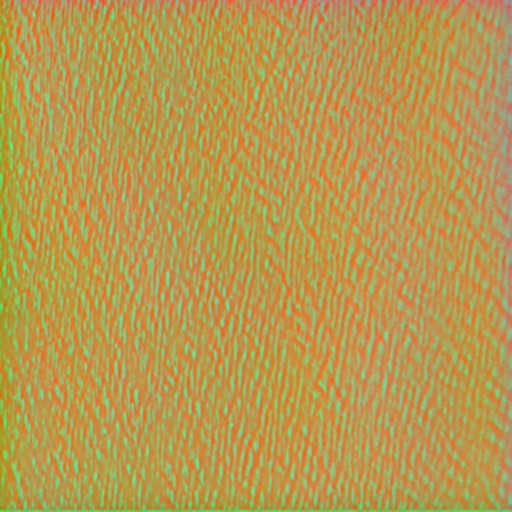}} &
{\includegraphics[width=0.19\linewidth]{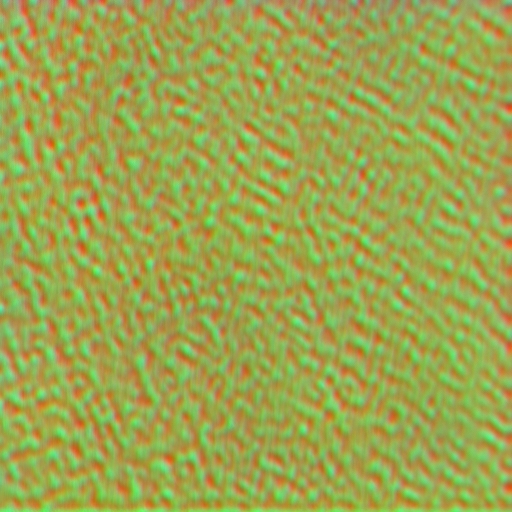}} &
{\includegraphics[width=0.19\linewidth]{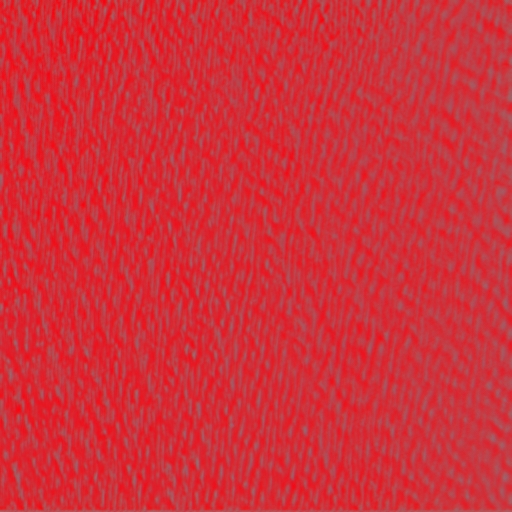}} &
{\includegraphics[width=0.19\linewidth]{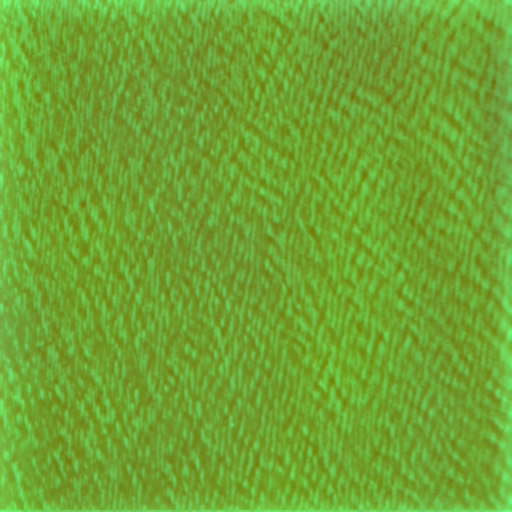}} 
\\
{\includegraphics[width=0.19\linewidth]{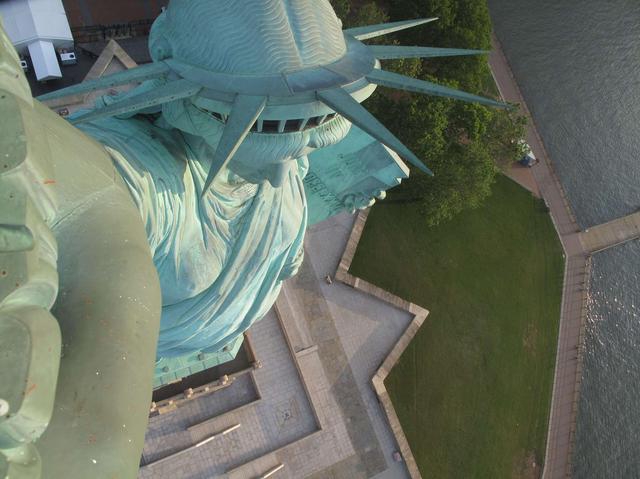}} &
{\includegraphics[width=0.19\linewidth]{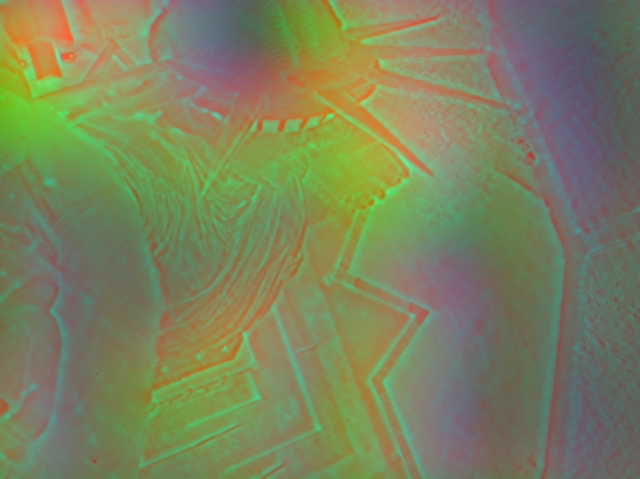}} &
{\includegraphics[width=0.19\linewidth]{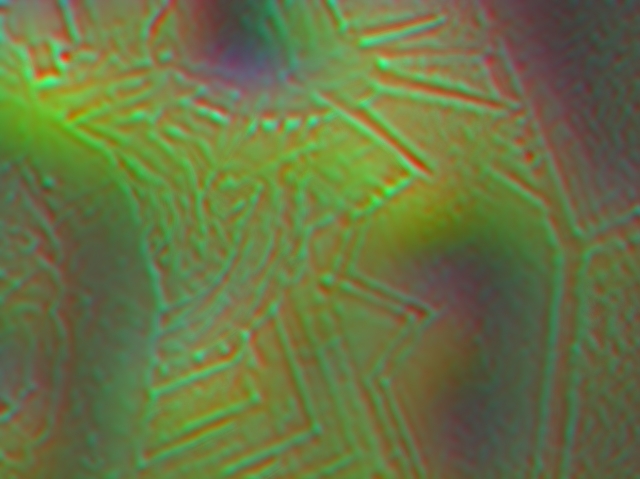}} &
{\includegraphics[width=0.19\linewidth]{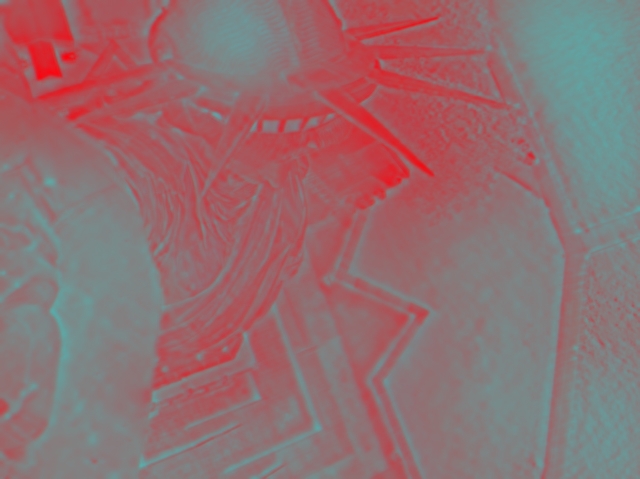}} &
{\includegraphics[width=0.19\linewidth]{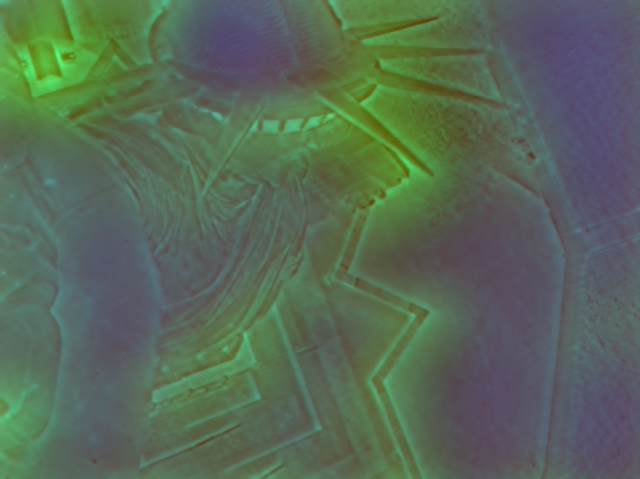}} 
\\
\capstyle{(a)} &
\capstyle{(b)} &
\capstyle{(c)} &
\capstyle{(d)} &
\capstyle{(e)}
\end{tabular}
\end{center}
\caption{Variations of the representation. (a) Original image, (b) Full model representation, (c) No Scale consistency loss, (d) No Multi-channel similarity loss, (e) With Rotation invariance loss.}
\label{fig:ablation2}
\end{center}
\end{figure}

Removing this loss reduces the sharpness of the representation, see Figure \ref{fig:ablation2}, column (c). This also affects the patch matching results. In the full model the performance is better for smaller patches (a matching task which is harder). However, the full model exhibits slightly worse accuracy for larger patches.

\smallskip\noindent\textbf{Multi-Channel Similarity Loss.} In this study we set to zero the weight of the multi-channel similarity loss,

\begin{equation}
\label{eq:multi_channel2}
     L_{MC}(I)=\sum_{i}\sum_{j\neq i} {(1-D_{corr}(I_{i},I_{j}))^2}.
\end{equation}

Without this loss the channels of the representation tend to be similar to each other or the negative of each other (highly correlated or anti-correlated), see Figure \ref{fig:ablation2}, column (d). Adding this loss promotes variability amongst the different channels and reduce information loss.

\smallskip
\noindent\textbf{Rotation invariance. } Following the purpose of introducing the scale consistency loss, Eq. \eqref{eq:scale_const2}, it appears natural to introduce also a rotation consistency loss. This can be formalized as:
\begin{equation}
L_{RI}(f_a)=\|F(H(f_a,\rho))-H(F(f_a),\rho)\|_2^2.
\end{equation} 
Where $H$ rotates the image/representation by a random angle \mbox{$\rho\in\{90, 180, 270\}$} degrees. This forces the representation to be invariant to (90 degree) rotations. Although this sounds highly reasonable (and might be necessary for some applications), we found out that the addition of this loss deteriorates performance. This might be explained by the fact that the color-coding of the dominant edge-direction, produced by the full model (see Section 5 in the paper), is direction dependant and thus it is lost here. (Figure \ref{fig:ablation2}, column (e)).

\subsection{"K-Channel" Representation}
Since our representation is unconstrained, in principle, it can be composed of an arbitrary number of channels. 
We trained and tested our full model with different number of output channels. This was achieved by changing the last convolutional layer of the network. 
We observe that 3 channels yield an optimal representation (in terms of matching). 

\begin{table}
    \centering
    \begin{tabular}{|c|c|c|c|}
    \hline
    \backslashbox[35mm]{Variation}{Patch Size}& 32 & 64 & 128  \\
    \hline\hline
    3 channels (Full model) & \bf{0.781} & 0.888 & 0.930\\
    \hline
    1 channel & 0.700 & 0.847 & 0.907\\
    2 channels & 0.711 & 0.862 & 0.907\\
    4 channels & 0.779 & 0.873 & 0.924\\
    5 channels & 0.769 & 0.862 & 0.912\\
    \hline
    No scale consistency loss & 0.740 &\bf{0.891} & \bf{0.941}\\
    No multi-channel similarity loss & 0.720 & 0.838 & 0.893\\
    With Rotation invariance loss & 0.748 & 0.831 & 0.897\\
    \hline
    \end{tabular}
    \label{table:ablation}
    \caption{
    Ablation study patch matching results: score by AUC of IoU-ROC curves.}
\end{table}

\newpage
\clearpage
\bibliography{egbib}
\end{document}